%% file: main.tex
\documentclass[twoside]{article}

\usepackage[accepted]{aistats2025}
\usepackage{titletoc}
\newcommand\DoToC{%
  \startcontents
  \printcontents{}{1}{\hrulefill\vskip0pt}
  \vskip0pt \noindent\hrulefill
  }
\usepackage{hyperref}
\usepackage{algorithm}
\usepackage{algorithmic}
\usepackage{subcaption}
\usepackage[round]{natbib}

\usepackage[utf8]{inputenc} 
\usepackage[T1]{fontenc}          
\usepackage{url}            
\usepackage{booktabs}      
\usepackage{amsfonts}      
\usepackage{nicefrac}      
\usepackage{microtype}      
\usepackage[table,dvipsnames]{xcolor}
\usepackage{amsmath}
\usepackage{amssymb}
\usepackage{mathtools}
\usepackage{amsthm}
\usepackage{setspace}
\usepackage{ltablex,booktabs,ragged2e,caption}

\usepackage[capitalize,noabbrev]{cleveref}

\usepackage{url}
\usepackage{algorithm, algorithmic}

\usepackage{graphicx}
\usepackage{multirow}
\usepackage{dsfont}
\usepackage{enumitem}
\usepackage{adjustbox}
\hypersetup{
    colorlinks,
    linkcolor={red!90!black},
    citecolor={blue!50!black},
    urlcolor={blue!80!black}
}
\definecolor{Gray}{gray}{0.93}
\definecolor{lightgray}{gray}{0.97}
\usepackage[most]{tcolorbox}
\usepackage{minitoc}
\newtcolorbox{mybox}[1][]{
enhanced jigsaw,
colback=Gray,
boxrule=0pt,
overlay unbroken and first ={},
breakable,
top=1pt,bottom=1pt,
left=1pt,right=1pt,
arc=0pt,outer arc=0pt,
#1}
\newtcolorbox{mybox_1}[1][]{
enhanced jigsaw,
colback=cyan!30,
boxrule=0pt,
overlay unbroken and first ={},
breakable,
top=1pt,bottom=1pt,
left=1pt,right=1pt,
arc=0pt,outer arc=0pt,
#1}
\newtcolorbox{mybox_c}[1][]{
enhanced jigsaw,
colback=blue!20,
boxrule=0pt,
overlay unbroken and first ={},
breakable,
top=1pt,bottom=1pt,
left=1pt,right=1pt,
arc=0pt,outer arc=0pt,
#1}
\newtcolorbox{mybox_y}[1][]{
enhanced jigsaw,
colback=blue!50,
boxrule=0pt,
overlay unbroken and first ={},
breakable,
top=1pt,bottom=1pt,
left=1pt,right=1pt,
arc=0pt,outer arc=0pt,
#1}
\newcommand\scalemath[2]{\scalebox{#1}{\mbox{\ensuremath{\displaystyle #2}}}}

\newtheorem{definition}{Definition}
\newtheorem{assumption}{Assumption}

\newtheorem{theorem}{Theorem}

\begin{document}
\twocolumn[

\aistatstitle{Causal Temporal Regime Structure Learning}

\aistatsauthor{ Abdellah Rahmani \And Pascal Frossard }

\aistatsaddress{ LTS4, EPFL\\ Lausanne, Switzerland \And  LTS4, EPFL\\ Lausanne, Switzerland  } ]

\begin{abstract}
Understanding causal relationships in multivariate time series is essential for predicting and controlling dynamic systems in fields like economics, neuroscience, and climate science. However, existing causal discovery methods often assume stationarity, limiting their effectiveness when time series consist of sequential regimes, consecutive temporal segments with unknown boundaries and changing causal structures.
In this work, we firstly introduce a framework to describe and model such time series. Then, we present CASTOR, a novel method that concurrently learns the Directed Acyclic Graph (DAG) for each regime while determining the number of regimes and their sequential arrangement. CASTOR optimizes the data log-likelihood using an expectation-maximization algorithm, alternating between assigning regime indices (expectation step) and inferring causal relationships in each regime (maximization step). We establish the identifiability of the regimes and DAGs within our framework. Extensive experiments show that CASTOR consistently outperforms existing causal discovery models in detecting different regimes and learning their DAGs across various settings, including linear and nonlinear causal relationships, on both synthetic and real world datasets.
\end{abstract}

\section{INTRODUCTION}
Causal structure learning from multivariate time series (MTS) variables is essential in many fields like disease evolution \citep{shen2020challenges} or climate science \citep{runge2019detecting}, as causal structures reveal complex real-world mechanisms. Recent approaches have focused on extracting causality from observational data, handling both linear and nonlinear relationships, and accommodating instantaneous and time-lagged connections \citep{pamfil2020dynotears,lowe2022amortized,runge2018causal,gong2022rhino}.
However, these methods often assume that time series data come from a single regime governed by one causal graph, which is inadequate for real-world scenarios. For example, causal dependencies vary across different climatic regimes \citep{karmouche2023regime} and financial settings \citep{huang2020causal}, time series are characterized by multiple unknown regimes in neurological studies regarding epilepsy \citep{rahmani2023meta}. In practice, successive regimes may originate from different causal models over the same variables, each described by a distinct temporal causal graph.\looseness=-1

Recent research has addressed causal discovery from multivariate time series (MTS) with various regimes. For example, CD-NOD \citep{huang2020causal} detects change points and produces a summary causal graph where each variable's parents are the union of its parents across all existing regimes. However, CD-NOD cannot infer individual causal graphs for each regime, does not highlight edges that appear or disappear between regimes, and is incapable of identifying recurring regimes, which is a crucial limitation in many scenarios. Another work addressing MTS composed of multiple regimes is RPCMCI \citep{saggioro2020reconstructing}, which learns a temporal graph for each regime. However, it only infers time-lagged relationships and requires prior knowledge of the number of regimes and transitions between them.
Overall, these methods make restrictive assumptions that limit their applicability in practical settings and cannot simultaneously identify the number of regimes and their corresponding indices, i.e., their start and end points.
\looseness=-1

\begin{figure*}[htbp]
    \centering
    \begin{subfigure}[b]{1.5\columnwidth}
        \centering
        \includegraphics[scale=0.6]{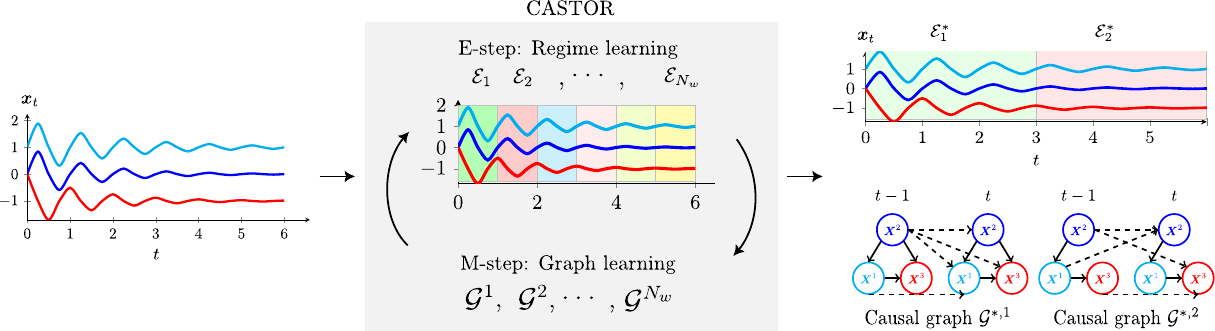} 
        \caption{}
        \label{fig:image_a}
    \end{subfigure}
    \hfill
    \begin{subfigure}[b]{0.5\columnwidth}
        \centering
        \includegraphics[width=0.8\linewidth]{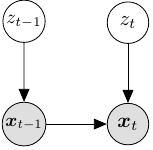} 
        \caption{} 
        \label{fig:image_b}
    \end{subfigure}
    \vspace{-5pt}
    \caption{(a) An illustration of CASTOR processing an input MTS using an EM procedure to infer two regimes determining their partitions ($\mathcal{E}^{*}_1$ and $\mathcal{E}^{*}_2$) and learning the temporal causal graphs. Dashed edges represent time-lagged links; solid arrows indicate instantaneous links. (b) CASTOR's graphical model for lag $L=1$: observed variables are depicted in grey, and the latent variables are uncolored.}
    \vspace{-12pt}
    \label{fig:overall_figure}
\end{figure*}

To address these limitations, we first present a new framework that formulates Structural Equation Models (SEMs) and Causal Graphical Models (CGMs) for MTS composed of multiple regimes. We introduce CASTOR, which, to the best of our knowledge, is the first method designed to learn causal relationships, determine the number of regimes, and uncover their arrangement from MTS with multiple regimes, each corresponding to an MTS block. The method is presented conceptually in Figure \ref{fig:image_a}. Unlike other methods, CASTOR does not require prior knowledge of the number of regimes or their indices. It optimizes the data log-likelihood using an expectation-maximization algorithm (EM), alternating between assigning regime indices (expectation step) and inferring causal relationships in each regime (maximization step). We prove that in our framework, comprising multivariate time series with multiple regimes modeled by Gaussian structural equation models with equal error variances, both the regimes and their corresponding graphs are identifiable up to a permutation of the regime labels.

Our extensive comparative analysis with causal discovery models tailored for MTS with multiple regimes shows that CASTOR consistently outperforms them across various scenarios in both structure learning and regime detection. Furthermore, we compare CASTOR with models that assume stationary MTS by providing them the ground truth regime partition information. Even with this advantage, our approach demonstrates similar or superior performance in inferring DAGs on both synthetic and real-world datasets. We finally apply CASTOR to two real-world datasets, IT monitoring data and Biosphere-Atmosphere data where the results show its ability to detect regimes and also generate explanatory DAGs. The main contributions of this paper can be summarized as follows:
\begin{itemize}
\vspace{-3pt}
\setlength\itemsep{-2pt}
    \item We formulate a new SEMs and CGMs for MTS composed of multiple regimes.
    \item We present, CASTOR, the first method designed to learn the number of regimes, their indices and their corresponding DAG from MTS with multiple regimes.
    \item We show that the exact maximization of the score function identifies the ground truth regimes and graphs up to a permutation in the case of Gaussian noise with equal variance.
    \item We show that CASTOR outperforms state-of-the-art methods in a wide variety of conditions, including linear and non-linear causal relationships and different number of nodes and regimes on both synthetic and real-world datasets.
\end{itemize}
\vspace{-18pt}
\paragraph{Related works. }\label{relatedworks}\citet{assaad2022survey} offer an extensive survey of methods for learning temporal causal relationships. Granger causality is the primary approach used for causal discovery from MTS \citep{lowe2022amortized,bussmann2021neural, xu2019scalable}. However, it is unable to accommodate instantaneous effects. DYNOTEARS \citep{pamfil2020dynotears}, on the other hand, leverages the acyclicity constraint established by \citet{zheng2018dags}  to continuously relax the DAG and differentiably learn instantaneous and time lagged structures. However, DYNOTEARS is still limited to linear functional forms. TiMINo \citep{peters2013causal} provides a general theoretical framework for temporal causal discovery with functional causal models. However, the aforementioned methods assume that MTS are composed of a single regime. \\
Several studies have sought to tackle the challenge of causal discovery in Non-stationary time series data \citep{huang2020causal, gunther2023causal, saggioro2020reconstructing}. Remarkably, \citet{huang2020causal} address the setting of time series composed of different regimes by
modulating causal relationships through a regime index. While it provides a summary graph highlighting behavioral changes across regimes, they cannot infer individual causal graphs neither the exact number of regime. \citet{saggioro2020reconstructing} assume knowledge of the number of regimes and propose the inference of only time-lagged links. Furthermore, they evaluate their algorithm on graphs with a limited number of nodes. Finally, \citet{balsellsidentifiability} addresses first-order regime-dependent causal discovery from MTS with multiple regimes. They proved that first-order Markov switching models with non-linear Gaussian transitions are identifiable up to permutations. Their work offers also a practical algorithms for regime-dependent causal discovery in time series data. However, its primary limitation is the assumption of solely time-lagged relationships, with the theory being restricted to a single time lag. (Detailed related work in Appendix \ref{relatedwork}).
\section{FRAMEWORK}
In this section, we first introduce our notations. Then we define the temporal causal graph, describe the setting of MTS with multiple regimes, and present a new SEMs for their representation.\looseness=-1

\textbf{Notation. } Matrices, vectors, and scalars are denoted by uppercase bold $\boldsymbol{G}_{\tau}$, lowercase bold $\boldsymbol{x_t}$ and lowercase normal letters $x_{t-\tau}^i$, respectively. Ground-truth variables are indicated with an asterisk, such as $\mathcal{G}^*$. We assume all distributions have densities $p\left(\boldsymbol{x}_{t}\right)$ w.r.t. the Lebesgue measure. The notation $[|0: L|]$ represents the set of integers $\{0,...,L\}$ and $|\cdot|$ denotes set cardinality. We denote a temporal causal graph (Definition \ref{definitionn1}) as $\mathcal{G}=(\mathbf{V}, E)$, represented by a collection of adjacency matrices $\boldsymbol{G}_{\tau \in [|0: L|]}=\{\boldsymbol{G}_{0},\dots,\boldsymbol{G}_{L}\}$. Following \citet{gong2022rhino}, $\mathbf{P a}_\mathcal{G}^i(<t)$ refers to the lagged parents of node $i$ in $\mathcal{G}$ at previous time $t-\tau$ with $ 1\leq\tau \leq L$, while $\mathbf{P a}_\mathcal{G}^i(t)$ denotes the instantaneous parents at the current time $t$ (i.e., $\tau=0$). $(\boldsymbol{x}_t)_{t \in \mathcal{T}} = (x_t^i)_{i \in \mathbf{V},t \in \mathcal{T}}$ represent a MTS of $|\mathbf{V}| = d$ components and length $|\mathcal{T}|$. $\mathcal{T}$ is the time index set.

\begin{definition}[Temporal Causal Graph] The temporal causal graph, associated with the MTS $\boldsymbol{(x_t)}_{t \in \mathcal{T}}$, is defined by a DAG $\mathcal{G}=(\mathbf{V}, E)$ and a fixed maximum lag $L$. Its vertices $\mathbf{V}$ consists of the set of components $x_{t'}^1, \ldots, x_{t'}^d$ for each $t' \in [|t-L:t|]$ . The edges $E$ of the graph are defined as follows: $\forall \tau \in [|1:L|]$ variables $x_{t-\tau}^i$ and $x_t^j$ are connected by a lag-specific directed link $x_{t-\tau}^i \rightarrow x_t^j$ in $\mathcal{G}$ pointing forward in time if and only if $x^i$ at time $ t- \tau $ causes $x^j$ at time $t$. Then the coefficient $[G_{\tau}]_{i j}$ associated with the adjacency matrix $\boldsymbol{G}_{\tau} \in \mathcal{M}_{d}(\mathbb{R})$ will be non-null and $x^i \in \mathbf{P a}_\mathcal{G}^j(<t)$ if $\tau \neq 0$. For instantaneous links ($\tau = 0$), we can not have self loops i.e. $i \neq j$. If $\tau=0,\text{ we have an edge } x_{t}^i \rightarrow x_t^j \text{ and } x^i \in \mathbf{P a}_\mathcal{G}^j(t)$ if and only if $x^i$ at time $ t$ causes $x^j$ at time $t$.
\label{definitionn1}
\end{definition} 
\textbf{MTS with multiple regimes assumption. }A MTS can exhibit either a single regime as assumed in prior works like Rhino \citep{gong2022rhino} and DYNOTEARS \citep{pamfil2020dynotears} or $K$ different non-overlapping regimes, as in our approach. For a MTS $(\boldsymbol{x}_t)_{t \in \mathcal{T}}$ composed of $K$ disjoint regimes, each regime $u$ is a stationary MTS block with a minimum duration $\zeta$ and has its own temporal causal graph $\mathcal{G}^u$, as defined in definition \ref{definitionn1}. We denote the set of these temporal causal graphs as $\mathbf{\mathcal{G}} = (\mathcal{G}^u)_{u \in [|1:K|]}$. Regimes occur sequentially, with the constraint that a subsequent regime $v$ (where $v = u+1$) cannot commence until at least $\zeta$ time units have passed since the start of the preceding one $u$, and also persists for a minimum of $\zeta$ samples. Additionally, if regime $u$ reoccurs, its duration in the second appearance is also no less than $\zeta$ samples (Minimum regime duration). The indices corresponding to all occurrences of regime $u$ are stored in a set denoted by  $\mathcal{E}_u$. The collection $\mathcal{E} = (\mathcal{E}_u)_{u \in [|1:K|]}$ represents the unique time partition of the MTS $(\boldsymbol{x}_t)_{t \in \mathcal{T}}$ composed of $K$ different regimes. In many application areas \citep{karmouche2023regime, rahmani2023meta}, non-stationarity can be modeled not through abrupt or continuous changes but rather as piecewise constant regimes. Importantly, the graphs $\mathcal{G}^u$ are regime-dependent, meaning that they vary across different regimes (i.e., $\mathcal{G}^u \neq \mathcal{G}^v$).

\textbf{SEMs for MTS with multiple regimes. }We now propose a novel functional form for SEMs that incorporates linear or non-linear relations, instantaneous links and multiple regimes. We have $\forall u \in [|1:K|], \forall t \in \mathcal{E}_u:$
\vspace{-3pt}
\begin{equation}
  x_t^i= g^u_i\left(\mathbf{P a}_{\mathcal{G}^u}^i(<t), \mathbf{P a}_{\mathcal{G}^u}^i(t)\right)+\epsilon_t^i,
\label{6}
\end{equation}
where $g^u_i$ is a general differentiable linear or non-linear function and $\epsilon_t^i \sim \mathcal{N}(0, 1)$, follows to a normal distribution. By assuming Causal Markov property, we can define the associated Causal Graphical Model (CGM), with $\boldsymbol{x}_{<t}$ refers $\{\boldsymbol{x}_{t-L},...,\boldsymbol{x}_{t-1}\}$, $\forall u \in [|1:K|], \forall t \in \mathcal{E}_u:$
\looseness=-1
\vspace{-3pt}
\begin{equation}
\small
p\left(\boldsymbol{x}_{t} \mid \textcolor{black}{\boldsymbol{x}_{<t}},\mathcal{G}^u\right)=\prod_{i=1}^d  p\left(x_t^{i} \mid \mathbf{P a}_{\mathcal{G}^u}^i(<t), \mathbf{P a}_{\mathcal{G}^u}^i(t)\right).
\label{4}
\vspace{-3pt}
\end{equation}
When the MTS consists of $K$ unknown regimes, it cannot be represented by a single DAG. A new formulation describing the CGM in such scenarios is as follows:
\vspace{-3pt}
\begin{equation}
p\left(\boldsymbol{x}_{t} \mid \textcolor{black}{\boldsymbol{x}_{<t}} \right)=\sum_{u=1}^{K}p(z_{t, u}) \cdot p\left(\boldsymbol{x}_{t} \mid \textcolor{black}{\boldsymbol{x}_{<t}}, \mathcal{G}^u\right), 
\label{55}
\vspace{-2pt}
\end{equation}
where $p\left(\boldsymbol{x}_{t} \mid \textcolor{black}{\boldsymbol{x}_{<t}},\mathcal{G}^u\right)$ is specified in Eq (\ref{4}), while $p(z_{t, u})$ models the probability of $\boldsymbol{x}_{t}$ belonging to regime $u$ ($\boldsymbol{x}_t$ belongs to regime $u$ if $z_{t,u} = 1$ Figure \ref{fig:image_b}). As we explained above, the regimes are non-overlapping hence the  $p(z_{t, u}) = \mathds{1}_{\mathcal{E}_u}(t)$ is an indicator function, defined as $\mathds{1}_{\mathcal{E}_u}(t)=1$ if $t \in \mathcal{E}_u$ and 0 otherwise. Previous works assume prior knowledge of time partition $\mathcal{E}$ or report a summary causal graph \citep{huang2020causal}, falling short of elucidating the full temporal causal graph. In the next section we present CASTOR, a causal discovery method tailored for MTS with multiple regimes.
\begin{figure*}[t]
    \centering
    \begin{subfigure}[b]{0.75\columnwidth}
       \centering

\includegraphics[width=0.95\linewidth]{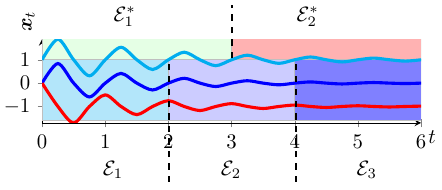} 
\caption{}
\label{fig:pure_imp}

    \end{subfigure}
    \hfill
    \begin{subfigure}[b]{0.65\columnwidth}
    \centering
    \includegraphics[width=0.99\linewidth]{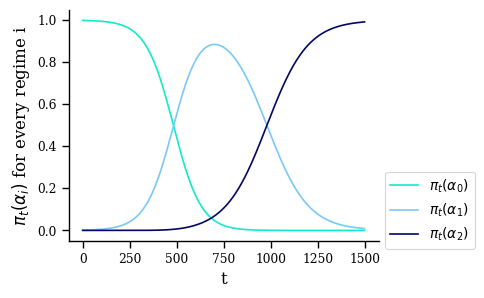} 
    \caption{}
\label{fig:pi_alignement}
    \end{subfigure}
    \hfill
    \begin{subfigure}[b]{0.55\columnwidth}
        \centering
        \includegraphics[width=0.99\linewidth]{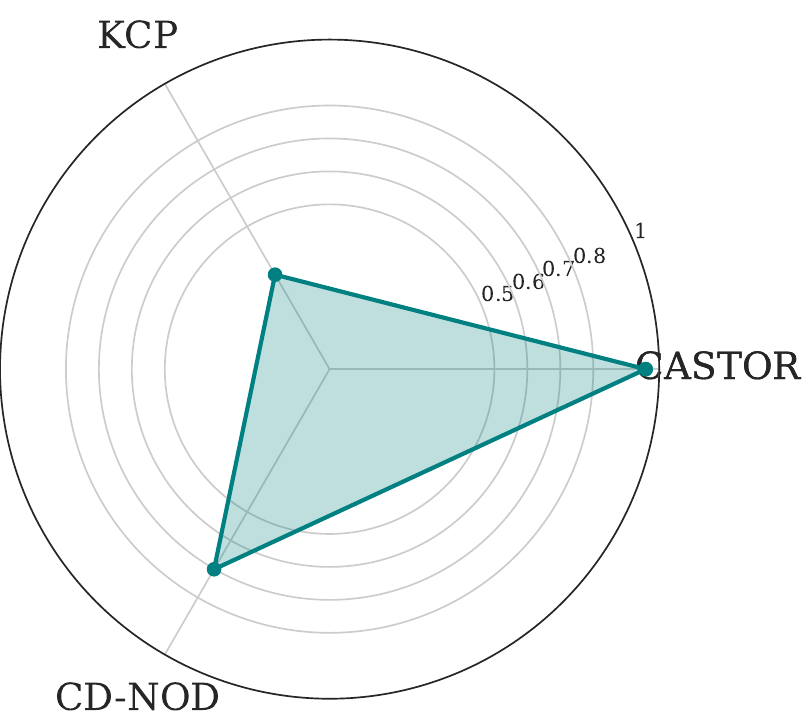} 
        \caption{}
        \label{fig:re_acc_exp}
    \end{subfigure}
    \vspace{-5pt}
    \caption{(a) Initialization with $N_w = 3$ windows: $\mathcal{E}_1$ and $\mathcal{E}_3$ are pure regimes; $\mathcal{E}_2$ is impure, containing samples from ground-truth regimes $\mathcal{E}^*_1$ and $\mathcal{E}^*_2$ ($K = 2$). (b) Illustration of $\pi\left(\boldsymbol{\alpha}^u, t\right)$ after CASTOR's first iteration with equal windows of 500 samples for an MTS of 1500 samples with two ground-truth regimes: $\mathcal{E}^*_1 = [|0:799|]$ and $\mathcal{E}^*_2 = [|800:1500|]$. (c) Comparison between CASTOR, CD-NOD and KCP on regime detection for a MTS of 10 nodes and 4 regimes using accuracy metric.\looseness=-1
    }
    \vspace{-12pt}
    \label{fig:overall_figure}
\end{figure*}
    \section{CASTOR: CAUSAL TEMPORAL REGIME STRUCTURE LEARNING}
\vspace{-5pt}
\label{sec3}
We propose a method to jointly learn the number of regimes $K$, their indices $\mathcal{E} = (\mathcal{E}_u)_{u \in [|1:K|]}$, and the corresponding DAGs $\mathbf{\mathcal{G}} = (\mathcal{G}^u)_{u \in [|1:K|]}$ from a MTS $(\boldsymbol{x}_t)_{t \in \mathcal{T}}$ with unknown regimes by maximizing the log-likelihood:\looseness=-1
\vspace{-3pt}
\begin{equation}
\small
  \log p\left( \boldsymbol{x}_{0:|\mathcal{T}|} \right) = \sum_{t=0}^{|\mathcal{T}|} \log \left( \sum_{u=1}^{K} \mathds{1}_{\mathcal{E}_u}(t) \, p\left( \boldsymbol{x}_{t} \mid \boldsymbol{x}_{<t}, \mathcal{G}^u \right) \right).
  \label{logp}
  \vspace{-3pt}
\end{equation}
Learning the DAGs $\mathbf{\mathcal{G}} = (\mathcal{G}^u)_{u \in [|1:K|]}$, concurrently entails the estimation of the regime distribution $p\left(\boldsymbol{x}_{t} \mid \textcolor{black}{\boldsymbol{x}_{<t}},\mathcal{G}^u\right)$. We model CASTOR's estimation of the joint density of the $u^{th}$ regime by:
\vspace{-3pt}
\begin{equation}
    f^{u}\left(\boldsymbol{x}_t\right):=\prod_{i=1}^d f^u_i\left(\mathbf{P a}_{\mathcal{G}^u}^i(<t), \mathbf{P a}_{\mathcal{G}^u}^i(t)\right),
    \label{6f}
    \vspace{-3pt}
\end{equation}
where $f^{u}$ is a distribution family. It is important to highlight that while $f^u$ can in theory be any distribution, in this particular study, we assume normal noise, used by many works (\cite{pamfil2020dynotears,huang2020causal}) and for which they showed the identifiability of causal graphs for one regime \citep{peters2014identifiability}. As a result from SEM Eq (\ref{6}), our distribution $f^{u}$ will be a Gaussian distribution.\\ Section~\ref{sec3.1} presents the challenges of the learning problem, while Sections \ref{3.1} and \ref{3.2} detail the procedures for regime learning and graph learning.
\vspace{-5pt}
\subsection{Challenges of the learning problem and EM choice justification}
\vspace{-5pt}
\label{sec3.1}
Since the regime indices are unknown, the learning problem is challenging. The sum inside the logarithm in Eq (\ref{logp}) renders the log-likelihood intractable. To address this, we employ the EM algorithm \citep{dempster1977maximum}, which introduces variables \( \gamma_{t,u} \), that the posterior probability $p \left(\boldsymbol{z}_t|\boldsymbol{x}_t, \boldsymbol{x}_{<t} \right)$, to model regime participation and alternates between regime learning (E-step) and graph learning (M-step). Hence we have the expected log likelihood:
\vspace{-3pt}
\begin{equation}
\scalebox{0.85}{$
\mathbb{E}_{\boldsymbol{z} \mid \boldsymbol{x}}\left[\log p\left(\boldsymbol{x}_{0: T}, \boldsymbol{z}_{0: T}\right)\right]=\sum_{t=0}^{|\mathcal{T}|}\sum_{u=1}^{K} \gamma_{t, u} \log \Big( \mathds{1}_{\mathcal{E}_u}(t) \cdot f^{u}\left(\boldsymbol{x}_t\right)\Big),$}
    \label{logp_em_}
    \vspace{-3pt}
\end{equation}
However, the logarithm term may be zero, causing the log-likelihood to diverge. Additionally, applying EM requires prior knowledge of the number of regimes, which we assume is unavailable.

We aim to learn continuous functions that solve the divergence problem, but also robust against random regime switching for a given sample $\boldsymbol{x}_{t} $. We want a sample $\boldsymbol{x}_t$, if belonging to regime $u$ in the current iteration, to be assigned to the neighboring regimes ($u-1, u+1$) or the same regime $u$ in the next iteration. We employ the soft-max function $\pi\left(\boldsymbol{\alpha}^u, t\right)=\frac{\exp \left(\alpha_1^u t+\alpha_0^u\right)}{\sum_{j=1}^{N_w} \exp \left(\alpha_1^j t+\alpha_0^j\right)}$ (function of $\boldsymbol{\alpha}^u \in \mathbb{R}^2$ and time index $t$).

Regarding the second challenge, the unavailability of the number of regimes, CASTOR initially divides the MTS into $N_w > K$ equal time windows in the initialisation step (the length of the initialized windows is greater than $\zeta$ minimum regime duration), where each window represents one initial regime estimate. Our initialization scheme builds some initial \textcolor{RoyalBlue}{pure} regimes (regimes composed of samples from the same ground truth regime) and other \textcolor{RoyalBlue}{impure} ones (regimes composed of samples from two neighboring ground truth regimes), idea illustrated in Figure \ref{fig:pure_imp}.
The expected log-likelihood of Eq (\ref{logp_em_}) becomes:
\begin{equation*}
\vspace{-5pt}
\scalebox{0.85}{$
\mathbb{E}_{\boldsymbol{z} \mid \boldsymbol{x}}\!\bigl[\log p(\boldsymbol{x}_{0:T}, \boldsymbol{z}_{0:T})\bigr]
= \sum_{t=0}^{|\mathcal{T}|}\sum_{u=1}^{N_w} \gamma_{t,u}\;\log \Bigl(\pi(\boldsymbol{\alpha}^u, t)\,f^{u}(\boldsymbol{x}_t)\Bigr).$}
\vspace{-2pt}
\end{equation*}

After our initialization scheme, we start with some \textcolor{RoyalBlue}{pure} regimes and \textcolor{RoyalBlue}{impure} regimes due to the fact that our initial windows are sufficiently small. CASTOR estimates the graphs (by learning $f^{u}\left(\boldsymbol{x}_t\right)$) for the different initial regimes and also learns the parameters $\boldsymbol{\alpha}^u$ that maximizes the alignment between $\gamma_{t, u}$ (rectangular function) and $\pi\left(\boldsymbol{\alpha}^u, t\right)$ (example on Figure \ref{fig:pi_alignement}) using the M-step (subsection \ref{3.2}). This alignment is desirable to ensure stability in the assignment of the samples. Then, our method alternates between updating the regime indices $\mathcal{E} = (\mathcal{E}_u)_{u \in [|1:N_w|]}$ during the E-step (subsection \ref{3.1}) and learning the temporal causal graphs $(\mathcal{G}^u)_{u \in [|1:N_w|]}$ along with new $\pi\left(\boldsymbol{\alpha}^u, t\right)$ during the M-step. This process repeats until a predefined maximum number of iterations is reached.

\textbf{Justification of EM choice. } We argue that inferring the regimes and learning the associated DAGs are intertwined problems, which makes the EM algorithm a suitable choice for this learning task. We investigate the possibility of solving the problem of causal structure learning from MTS with multiple regimes by first using change point detection method to identify regime indices, then applying existing causal discovery methods for each regime. However, as shown in Figure \ref{fig:re_acc_exp}, standard change point detection methods failed to detect regimes resulting from changes in causal mechanisms. Specifically, when comparing CASTOR and CD-NOD \citep{huang2020causal} with KCP \citep{arlot2019kernel}, a state-of-the-art change point detection method, we observed that KCP was unable to detect the regimes. This is likely because changes in causal mechanisms involve shifts in conditional distributions, which are harder for KCP to detect. Further details are provided in the Appendix \ref{app_reg_detect}.
\vspace{-5pt}
\subsection{Expectation step: Regime learning}
\vspace{-5pt}
\label{3.1}
During the E-step, CASTOR updates $\gamma_{t, u}$ (as shown in Eq (\ref{10}), derivation details in Appendix \ref{appA}), the probability of $\boldsymbol{x}_t$ belonging to regime $u$ depends on two factors, the position of $\boldsymbol{x}_t$ within the current regime and whether the current regime is \textcolor{RoyalBlue}{pure or impure}.\\
\begin{equation}
\begin{aligned}
\textcolor{black}{\gamma_{t, u}} & \textcolor{black}{=\frac{p\left(z_{t, u}=1\right) p\left(\boldsymbol{X}_t \mid \boldsymbol{X}_{<t}, z_{t, u}=1, \mathcal{G}^u\right)}{\sum_{j=1}^{N_w} p\left(z_{t, j}=1\right) p\left(\boldsymbol{X}_t \mid \boldsymbol{X}_{<t}, z_{t, j}=1, \mathcal{G}^j\right)}} \\
& \textcolor{black}{=\frac{\pi\left(\boldsymbol{\alpha}^u, t\right) f^{u}\left(\boldsymbol{X}_t\right)}{\sum_{j=1}^{N_w} \pi\left(\boldsymbol{\alpha}^j, t\right) f^{j}\left(\boldsymbol{X}_t\right)} \propto \pi\left(\boldsymbol{\alpha}^u, t\right) f^{u}\left(\boldsymbol{X}_t\right)}
\end{aligned}
\label{10}
\end{equation}
\textbf{Case 1:} When $\boldsymbol{x}_t$ belongs to a pure regime $u$ and is far from the border in the current iteration, $\pi\left(\boldsymbol{\alpha}^u, t\right)$ is high (e.g., $\pi(\boldsymbol{\alpha}^0, t \in [0,300])$ in Figure \ref{fig:pi_alignement}). The graph for $u$ is meaningful because it was learned on pure data, so $f^{u}(\boldsymbol{x}_t)$ is also high. Consequently, $\gamma_{t,u} \propto \pi\left(\boldsymbol{\alpha}^u, t\right) f^{u}(\boldsymbol{x}_t)$ (Eq (\ref{10})) remains maximal, keeping $\boldsymbol{x}_t$ in regime $u$ in the next iteration.
\looseness=-1

\textbf{Case 2:}
$\boldsymbol{x}_t$ belongs to pure regime $u$ and is near the border in the current iteration. In this situation, $\pi\left(\boldsymbol{\alpha}^u, t\right)$ and $\pi\left(\boldsymbol{\alpha}^{u+1}, t\right)$ are approximately equal (e.g., $\pi(\boldsymbol{\alpha}^0, t \in [350,500])$ and $\pi(\boldsymbol{\alpha}^1, t \in [350,500])$ in Figure~\ref{fig:pi_alignement}). Since the graph for regime $u$ was learned from pure data, $f^{u}(\boldsymbol{x}_t)$ remains high. Therefore, $\gamma_{t,u}$ stays maximal, keeping $\boldsymbol{x}_t$ in regime $u$ in the next iteration.

\textbf{Case 3:}
$\boldsymbol{x}_t$ belongs to impure regime $u+1$ and is near the border in the current iteration. Here, $\pi\left(\boldsymbol{\alpha}^u, t\right)$ and $\pi\left(\boldsymbol{\alpha}^{u+1}, t\right)$ are roughly equal (e.g., $\pi(\boldsymbol{\alpha}^0, t \in [501,650])$ and $\pi(\boldsymbol{\alpha}^1, t \in [501,650])$ in Figure~\ref{fig:pi_alignement}). However, because the graph of regime $u$ is more meaningful (learned from pure data), $f^{u}(\boldsymbol{x}_t) > f^{u+1}(\boldsymbol{x}_t)$. Thus, $\gamma_{t,u} > \gamma_{t,u+1}$, causing $\boldsymbol{x}_t$ to switch from regime $u+1$ to $u$ in the next iteration.
\looseness=-1

\textbf{Case 4:}
$\boldsymbol{x}_t$ belongs to impure regime $u+1$ and is far from the border (e.g., $t \in [650,850]$ in Figure \ref{fig:pi_alignement}). In this case, it's uncertain whether $\boldsymbol{x}_t$ will switch regimes in the next iteration. However, as the pure regime $u$ expands with each iteration, $\boldsymbol{x}_t$ will eventually be near the border of regime $u+1$, bringing us back to Case~3.

For simplicity reasons, we explicit these cases from one border but the same thing happens in the other border which accelerates convergence.
If a sample $\boldsymbol{x}_t$ belongs to regime $u$, it will never be allocated to a non neighboring regime $v$ due to the fact that $\pi\left(\boldsymbol{\alpha}^u, t\right)>>\pi\left(\boldsymbol{\alpha}^v, t\right)$ and $f^{u}\left(\boldsymbol{x}_t\right)>>f^{v}\left(\boldsymbol{x}_t\right)$.
\vspace{-0.55em}
\begin{mybox}
\textbf{Example. } In Figure \ref{fig:pi_alignement}, after learning the graphs and parameters $\boldsymbol{\alpha}^u$ for the first iteration where the regimes are equal windows of 500 data points. The samples $\boldsymbol{x}_t$, where $t \in [|350,500|]$ belonging to regime 0 in the previous iteration, are more likely to stay in regime 0 or transition to the neighboring regime 1 (\textcolor{cyan}{$\pi(\boldsymbol{\alpha}^1, t \in [350,500])$} same range as \textcolor{Aquamarine}{$\pi(\boldsymbol{\alpha}^0, t \in [350,500])$}) than the non neighboring regime 2 (\textcolor{BlueViolet}{$\pi(\boldsymbol{\alpha}^2, t \in [350,500])$} almost 0).   
\end{mybox}
After updating $\gamma_{t, u}$, for each sample $\boldsymbol{x}_t$, CASTOR assigns a value of 1 to the most probable regime $u$ (with the highest $\gamma_{t, u}$), and 0 to others. Additionally, CASTOR filters out regimes with insufficient samples (fewer than $\zeta$, the minimum regime duration, defined as a hyper-parameter). Discarded regime samples are then reassigned to the nearest regime in terms of probability $\gamma_{t, u}$ in the subsequent iteration which is in general a neighboring regime ensured by the way we set up the probability $\gamma_{t, u}\propto \pi\left(\boldsymbol{\alpha}^u, t\right)f^{u}\left(\boldsymbol{x}_t\right)$.
\subsection{Maximization step: Graph learning}
\label{3.2}
CASTOR utilizes the binary regime indices $\gamma_{t,u}$ learned in the E-step to estimate the DAGs for each regime and learn the parameters $\boldsymbol{\alpha} = \{\boldsymbol{\alpha}^u, u \in [|1,N_w|]\}$ that align $\pi\left(\boldsymbol{\alpha}^u, t\right)$ with $\gamma_{t,u}$ by maximizing the following equation:
\vspace{-7pt}
\begin{equation*}
\begin{split}
\sup_{\mathbf{\mathcal{G}},\boldsymbol{\alpha}} \frac{1}{|\mathcal{T}|} \sum_{u=1}^{N_w} \sum_{t=0}^{|\mathcal{T}|} \gamma_{t,u} \log \pi\left(\boldsymbol{\alpha}^u, t\right) f^u(\boldsymbol{x}_t) - \lambda |\mathcal{G}^u|, \\
\text{s.t.} \quad \boldsymbol{G}^u_0 \text{ is a DAG.} \hfill
\end{split}
\label{Mstep}
\vspace{-8pt}
\end{equation*}
The maximization of the aforementioned equation can be decomposed into two distinct maximization problems. The first problem, \textcolor{cyan}{regime alignment}, focuses on aligning $\pi\left(\boldsymbol{\alpha}^u, t\right)$ with $\gamma_{t,u}$:
\vspace{-5pt}
\begin{equation}
\sup_{\boldsymbol{\alpha}}\frac{1}{|\mathcal{T}|} \sum_{u=1}^{N_w} \sum_{t=1}^{|\mathcal{T}|} \gamma_{t,u} \log \pi\left(\boldsymbol{\alpha}^u, t\right), 
\label{Mstep}
\vspace{-5pt}
\end{equation}
while the second one, \textcolor{YellowGreen}{graph learning}, involves estimating DAGs for every regime:\\
\vspace{-5pt}
\begin{equation}
\begin{split}
\mathcal{S}(\mathbf{\mathcal{G}}, \mathcal{E}) := \sup_{\mathbf{\mathcal{G}}} \frac{1}{|\mathcal{T}|} \sum_{u=1}^{N_w} \sum_{t \in \mathcal{E}_u} \log f^u(\boldsymbol{x}_t) - \lambda |\mathcal{G}^u|,\\ \text{ s.t., } \boldsymbol{G}^u_0 \text{ is a DAG.}
\end{split}
\label{Mstep}
\vspace{-5pt}
\end{equation}
where $\mathbf{\mathcal{G}}$ stands for $\mathbf{\mathcal{G}} = (\mathcal{G}^u)_{u \in [|1:N_w|]}$, $\boldsymbol{\alpha} = \{ \boldsymbol{\alpha}^u, \forall u \in [|1,N_w|]\}$, $|\mathcal{G}^u|$ is the number of edges in the temporal causal graph of regime $u$ and we note $\mathcal{S}(\mathcal{G},\mathcal{E})$ the score function of CASTOR. The first term in CASTOR's score function is the averaged log-likelihood over data while the second term is a penalty term with positive small coefficient $\lambda$ that controls the sparsity constraint. We further impose an acyclicity constraint on the adjacency matrix $\boldsymbol{G}^u_0$ of instantaneous links. The other adjacency matrices $\boldsymbol{G}^u_{\tau \in [|1: L|]}$ are inherently acyclic by definition, because these matrices establish links between variables at time $t$ and their time-lagged parents at time $t-\tau$. It is worth noting that the optimization for \textcolor{cyan}{regime alignment} remains the same for both linear and nonlinear causal relationships. However, the \textcolor{YellowGreen}{graph learning} problem differs between these two settings.\looseness=-1
\subsection{Linear case}
For linear causal relationships, the SEM (Eq (\ref{6})) is:
\vspace{-7pt}
\begin{equation*}
\small
\forall u \in [|1:,K|], \forall t \in \mathcal{E}_u: \boldsymbol{x}_t = \boldsymbol{x}_t \boldsymbol{G}^u_0 + \sum_{\tau=1}^L \boldsymbol{x}_{t-\tau} \boldsymbol{G}^u_\tau + \boldsymbol{\epsilon}_t, \quad 
\vspace{-5pt}
\end{equation*}
where $\epsilon_t \sim \mathcal{N}(0, I).$ Thus, $\forall u \in [|1:,K|], \forall t \in \mathcal{E}_u:$
\vspace{-8pt}
\begin{equation*}
\small
 \boldsymbol{x}_t \mid \boldsymbol{x}_{<t} \sim \mathcal{N}(\boldsymbol{x}_t \boldsymbol{G}^u_0 + \sum_{\tau=1}^L \boldsymbol{x}_{t-\tau} \boldsymbol{G}^u_\tau, I).
 \vspace{-5pt}
\end{equation*}
Using CASTOR's score function results in the following minimization problem (details in Appendix \ref{appA}):
\\
\vspace{-5pt}
\begin{equation}
\begin{split}
\scalemath{0.9}{\min_{\mathbf{\mathcal{G}}} \frac{1}{|\mathcal{T}|} \sum_{u=1}^{N_w} \sum_{t=1}^{|\mathcal{T}|} \gamma_{t,u} \left\| \boldsymbol{x}_t - \left( \boldsymbol{x}_t \boldsymbol{G}^u_0 + \sum_{\tau=1}^L \boldsymbol{x}_{t-\tau} \boldsymbol{G}^u_\tau \right) \right\|_F^2 }\\
\scalemath{0.9}{+\lambda|\mathcal{G}^u| + \frac{\rho}{2} h\left(\boldsymbol{G}^u_0\right)^2 
+\alpha h\left(\boldsymbol{G}^u_0\right)},
\end{split}
\label{1313}
\vspace{-5pt}
\end{equation}
where $\mathbf{\mathcal{G}}$ stands for $\mathbf{\mathcal{G}} = (\mathcal{G}^u)_{u \in [|1:N_w|]}$ and $\alpha, \rho$ characterize the strength of the DAG penalty. The function $h(\boldsymbol{G})=\operatorname{tr}\left(e^{\boldsymbol{G} \odot \boldsymbol{G}}\right)-d$ corresponds to the acyclicity constraints proposed in \cite{zheng2018dags} ($\odot$ is the Hadamard product). For example, let $\boldsymbol{G}_0$ be the adjacency graph for instantaneous relation. The constrain condition requires $h(\boldsymbol{G}_0) = 0$. We employ an augmented Lagrangian method \citep{zheng2018dags, pamfil2020dynotears, brouillard2020differentiable, liu2023causal} to address the optimization challenge that incorporates the acyclicity constraints. 
\begin{algorithm}[tb]
   \caption{CASTOR algorithm}
   \label{alg:CASTOR}
\begin{algorithmic}
   \STATE {\bfseries Input:} MTS $\boldsymbol{X}$, window size $W$, lag $L$, maximum number of iteration $N_\text{iter}$, minimum regime duration $\zeta$
   \FOR{$i=1$ {\bfseries to} $N_\text{iter}$}
   \begin{mybox_1}
      \STATE $\gamma_{t, u} \gets \frac{\pi_{t}(\alpha_u) f^{u}\left(\boldsymbol{x}_t\right)}{\sum_{j=1}^{N_w} \pi_{t}(\alpha_j) f^{j}\left(\boldsymbol{x}_t\right)}$ \texttt{Expectation step} 
   \end{mybox_1}
\begin{mybox_c}
   \STATE $\alpha \gets \text{argmin}_\alpha \sum_{u=1}^{N_w} \sum_{t=1}^{|\mathcal{T}|} \gamma_{t, u} \log \left(\pi_{t}(\alpha_u)\right)$ \texttt{Regime alignment} 
   \end{mybox_c}
   \begin{mybox_y}
\STATE $\mathcal{G} \gets \text{argmin of Eq (\ref{1313}) or (\ref{eq1010})}$ \texttt{Graph learning} 
\end{mybox_y}
\IF{$\sum_{t}^{|\mathcal{T}|}\gamma_{t, u}\leq \zeta$}
\STATE $\forall t: \gamma_{t, u} \gets 0$ 
\ENDIF
\ENDFOR
\STATE {\bfseries Output:} $\gamma$, $\mathcal{G}$
\end{algorithmic}
\end{algorithm}
\subsection{Non-linear case}
\label{3.2.1}
For non-linear causal relationships in the MTS $\boldsymbol{(x_t)}_{t \in \mathcal{T}}$, we estimate the distribution parameters $f^u$ (Eq (\ref{Mstep})) by modeling the non-linear SEM (Eq (\ref{6})). According to Eq (\ref{6}), each component $x^i_t$ follows a Gaussian distribution:
\vspace{-5pt}
\begin{equation*}
\small
x^i_t \mid \boldsymbol{x}_{<t} \sim \mathcal{N}(g^u_i(\mathbf{Pa}_{\mathcal{G}^u}^i(<t), \mathbf{Pa}_{\mathcal{G}^u}^i(t)), 1).
\end{equation*}
We employ Neural Networks (NN) to capture the non-linearity, as in \citet{zheng2020learning, brouillard2020differentiable, liu2023causal}. For each regime $u$ and component $i$, a separate and small $\text{NN}^u_i$ models the distribution parameters. We aggregate the lagged variables into $\boldsymbol{x}_t^{\text{lag}} = [\boldsymbol{x}_{t-1} | \cdots | \boldsymbol{x}_{t-L}]$ and input both $\boldsymbol{x}_t$ and $\boldsymbol{x}_t^{\text{lag}}$ into $\text{NN}^u_i$ to predict $\hat{x}^i_t$, thereby estimating $f^u_i$.
Our neural networks are defined as,
\vspace{-7pt}
\begin{equation*}
\small
\quad  \forall i \in [|1:d|]: \text{NN}^u_i(\boldsymbol{x}_t, \boldsymbol{x}_t^{\text{lag}}) = \psi^u_i\left(\phi^u_i(\boldsymbol{x}_t), \phi^{u,\text{lag}}_i(\boldsymbol{x}_t^{\text{lag}})\right),
\vspace{-7pt}
\end{equation*}
where $\psi^u_i$ consists of locally connected layers \citep{zheng2020learning} and activation functions, and $\phi^u_i$, $\phi^{u,\text{lag}}_i$ are composed of linear layers and sigmoid functions. The locally connected layers help to capture variable dependencies in the initial layer.

For each node $i$, the instantaneous and time-lagged interactions with node $j$ are captured by the norms of the corresponding columns in the first layer weight matrices:
\begin{equation*}
\centering
\small
\begin{aligned}
& [\boldsymbol{G}^u_0]_{ij} = \left\| \Theta^u_i(\text{column } j) \right\|_2, \\& [\boldsymbol{G}^u_\tau]_{ij} = \left\| \Theta_i^{u,\text{lag}}(\text{column } ((\tau-1) \cdot d + j)) \right\|_2,
\end{aligned}
\vspace{-5pt}
\end{equation*}
where $\Theta^u_i$ and $\Theta_i^{u,\text{lag}}$ are parameters of the first layers of the NNs $\phi^u_i$ and $\phi^{u,\text{lag}}_i$, respectively. The matrix $\Theta_i^{\text{lag}} \in \mathcal{M}_{d, dL}(\mathbb{R})$ has $dL$ columns, with $L$ as the maximum lag. Incorporating NNs into the maximization step (Eq.~\ref{Mstep}) results in the following minimization problem (details in Appendix~\ref{appA}):
\vspace{-5pt}
\begin{equation}
\small
\begin{aligned}
\min_{\theta,\mathcal{G}} \quad  \frac{1}{|\mathcal{T}|} \sum_{u=1}^{N_w} \sum_{t=1}^{|\mathcal{T}|} \sum_{i=1}^d & \gamma_{t,u} \mathcal{L}\left(x^i_t, \psi^u_i\left(\phi^u_i(\boldsymbol{x}_t), \phi^{u,\text{lag}}_i(\boldsymbol{x}_t^{\text{lag}})\right)\right)\\
& + \lambda |\mathcal{G}^u| + \frac{\rho}{2} h(\boldsymbol{G}^u_0)^2 + \alpha h(\boldsymbol{G}^u_0),
\end{aligned}
\label{eq1010}
\end{equation}
where $\mathcal{L}$ is a least squares loss, $\theta$ includes all network parameters, and $\mathcal{G} = (\mathcal{G}^u)_{u \in [|1:N_w|]}$. We enforce the acyclicity constraint using an augmented Lagrangian term. Algorithm~\ref{alg:CASTOR} summarizes our CASTOR model for both linear and nonlinear causal relationships.
\section{IDENTIFIABILITY RESULTS}
\label{identi_res}
In this section, we present the identifiability results of our framework, highlighting that the causal structure can be recovered from observational data only. Following \citet{brouillard2020differentiable, gong2022rhino, liu2023causal}, we outline our assumptions below:
\looseness=-1
\begin{definition}[Causal Stationarity, \citet{runge2018causal}]
    A stationary time series process $(\boldsymbol{x}_t)_{t \in \mathcal{T}}$ with graph $\mathcal{G}$ is called causally stationary over a time index set $\mathcal{T}$ if and only if for all links $\boldsymbol{x}_{t-\tau}^i \rightarrow \boldsymbol{x}_t^j$ in the graph
\begin{equation*}
    x_{t-\tau}^i \not\!\perp\!\!\!\perp x_t^j \mid \boldsymbol{x}_{<t} \setminus \{x_{t-\tau}^i\}.
 \vspace{-10pt}   
\end{equation*}
\label{def_caus_sta}
\end{definition}
\begin{assumption}[Causal Stationarity for MTS with multiple regime]
A MTS $(\boldsymbol{x}_t)_{t \in \mathcal{T}}$ with $K$ regimes, graph set $(\mathcal{G}^u)_{u \in [|1:K|]}$, and regime partition $\mathcal{E} = (\mathcal{E}_u)_{u \in [|1:K|]}$ is \textbf{causally stationary} over the time index set $\mathcal{T}$ if, for each regime $u \in [|1:K|]$, the sub-series $(\boldsymbol{x}_t)_{t \in \mathcal{E}_u}$ is causally stationary with graph $\mathcal{G}^u$ as defined in definition \ref{def_caus_sta}.
\label{mainass1}
\end{assumption}
\vspace{-5pt}
\begin{assumption}[Causal Markov Property (CPM)]
A set of joint distributions $(p(\cdot|\mathcal{G}^u))_{u \in [|1:K|]}$ satisfies the \textbf{CPM} with respect to the DAGs $(\mathcal{G}^u)_{u \in [|1:K|]}$ if, for each $u \in [|1:K|]$, the distribution $p(\cdot|\mathcal{G}^u)$ satisfies the CPM relative to the DAG $\mathcal{G}^u$. Specifically, in every regime $u$, each variable is independent of its non-descendants given its parents.
\label{mainass2}
\end{assumption}
\vspace{-2pt}
\begin{assumption}[Causal Minimality]
 Given a set of DAGs $(\mathcal{G}^u)_{u \in [|1:K|]}$ and a set of joint distribution $(p(\cdot|\mathcal{G}^u))_{u \in [|1:K|]}$, we say that this set of distributions satisfies causal minimality w.r.t. the set of DAGs $(\mathcal{G}^u)_{u \in [|1:K|]}$  if for every $u$: $p(\cdot|\mathcal{G}^u)$ is Markovian w.r.t the DAG $\mathcal{G}^u$ but not to any proper subgraph of $\mathcal{G}^u$.
\label{mainass3}
\end{assumption}
\vspace{-2pt}
\begin{assumption}[Causal Sufficiency]
A set of observed variables $\boldsymbol{V}$ is causally sufficient for a process $\boldsymbol{x}_t$ if and only if in the process every common cause of any two or more variables in $\boldsymbol{V}$ is in $\boldsymbol{V}$ or has the same value for all units in the population.  \label{mainass4}
\end{assumption}
Using the above assumptions and operating in the settings outlined in \citet{brouillard2020differentiable, liu2023causal}, Theorem \ref{theo:1} states that the ground truth solution $\left(\mathcal{G}^*, \mathcal{E}^*\right)$ uniquely maximizes the score defined in Eq (\ref{Mstep}), up to a permutation.
\vspace{-2pt}
\begin{theorem}
Assume SEMs with Gaussian noise, presented in Eq(\ref{6}), that  satisfy the causal Markov property, stationarity, minimality and sufficiency. If each regime has enough data and the penalty coefficients in Eq (\ref{1313}-\ref{eq1010}) are sufficiently small, it holds asymptotically that for any estimation 
\vspace{-5pt}
\begin{equation*}
   \mathcal{S}\left(\mathcal{G}^*, \mathcal{E}^*\right) > \mathcal{S}(\hat{\mathcal{G}}, \hat{\mathcal{E}})  
   \vspace{-5pt}
\end{equation*}
If any of the estimated graphs $\hat{\mathcal{G}}^u$ represents an edge disagreement with all the ground truth graphs $\mathcal{G}^*$ or any of the estimated regimes in $\hat{\mathcal{E}}$ is close to none of the ground truth regimes in the sense of Kullback–Leibler. \label{theo:1}
\end{theorem}
\vspace{-5pt}
Full proof is provided in Appendix \ref{appB}. 
When this score is maximized CASTOR can identify true regimes and causal graphs up to a permutation. However, the convergence is not always guaranteed due to EM instability and the non-convexity of acyclicity constraints. 
Moreover, Theorem \ref{theo:1} does not provide additional information on the ranking of various solutions.
Given two sub optimums, one  closer to the ground truth solution  w.r.t. KL divergence, Theorem \ref{theo:2} shows  the closest one has the higher score. 
\looseness=-1
\begin{theorem}
Assume the same conditions as in Theorem \ref{theo:1}, for any estimations $(\mathcal{G}, \mathcal{E})$ and $(\mathcal{G'}, \mathcal{E'})$, such that $(\mathcal{G}, \mathcal{E})$ is closer to the optimal solution $\left(\mathcal{G}^*, \mathcal{E}^*\right)$ than $(\mathcal{G'}, \mathcal{E'})$ in terms of Kullback–Leibler, it holds asymptotically: $\mathcal{S}(\mathcal{G}, \mathcal{E})>\mathcal{S}(\mathcal{G'}, \mathcal{E'}).$
\label{theo:2}
\vspace{-0.8em}
\end{theorem}
\section{EXPERIMENTS}
\subsection{Synthetic data}
\label{sec5.1} \label{others}
\textbf{Data generation. }We conduct extensive experiments to evaluate CASTOR's performance on synthetic datasets (details in Appendix \ref{appsyntheticdata}). For ground truth graph generation, we use the Barabási-Albert model (degree 4) for instantaneous links and the Erdős–Rényi model (degree 1–2) for time-lagged relationships. For non-linear cases, the functions $g^u_i$ in Eq (\ref{6}) use random weights from a uniform distribution over ]0,2] and activation functions randomly chosen from \{\texttt{Tanh, LeakyReLU, ReLU}\}. We consider $L = 1$, while additional experiments with multiple lags are provided in the Appendix. Regime durations are randomly selected from \{300, 400, 500, 600\}. We test different numbers of nodes (\{5, 10, 20, 40\} for linear cases, \{10, 20\} for non-linear) and varying regime counts ($K \in \{2, 3, 4, 5\}$). Each combination of $K$ and $d$ nodes is repeated three times, resulting in over 60 distinct datasets.

\textbf{Benchmarks. }  We benchmark our model against several baselines, including causal discovery methods for MTS with multiple regimes, such as CD-NOD \citep{huang2020causal} and RPCMCI \citep{saggioro2020reconstructing}. Since CD-NOD returns a summary graph (see Appendix \ref{app_reg_detect}), we compute a comparable summary graph from CASTOR's output for fair evaluation. CASTOR is also compared with models for single-regime MTS, including Rhino, with and without historically dependent noise, \citep{gong2022rhino}, PCMCI+ , using Partial correlation for linear relationships and GPDC for non-linear ones, \citep{runge2020discovering}, DYNOTEARS \citep{pamfil2020dynotears}, and VARLINGAM \citep{hyvarinen2010estimation}.
\textit{Given that these models cannot deal with multiple regimes, to make the evaluation fair, we put them in a more favorable position and provide these models with the true regime partition information. This is done by training the aforementioned models on each pure regime separately (regime governed by the same graph)}.

\begin{figure}[t]
\centering
\includegraphics[scale=0.27]{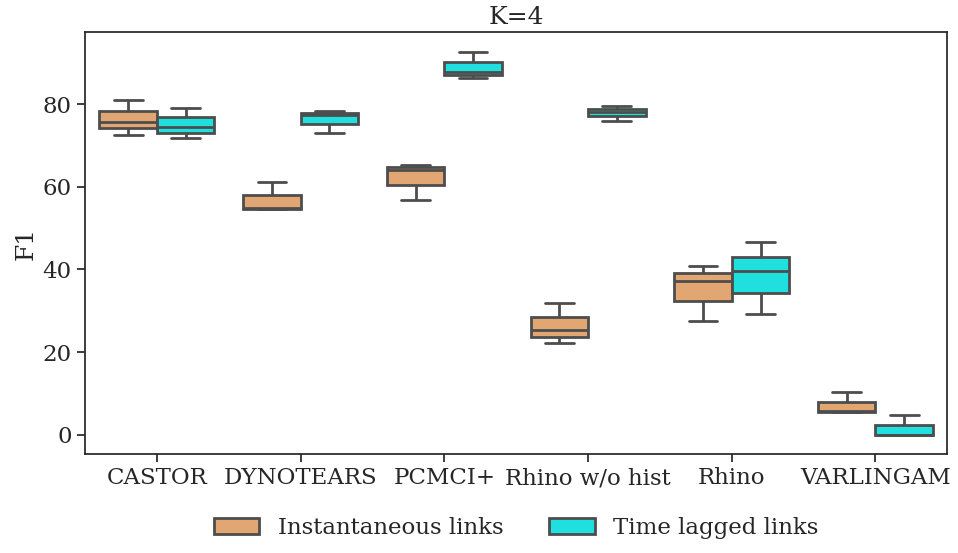}
\caption{F1 scores by Models for 20 nodes and 4 regimes for \textcolor{teal}{non linear causal relationships}. Orange indicates performance on instantaneous links, and sky-blue signifies performance on time-lagged relationships.}
\label{fig:1}
\vspace{-12pt}
\end{figure}

\textbf{Results and discussion. }Table \ref{lineartable} presents the results for linear case, while Figure \ref{fig:1} summarizes the results for the non-linear setting. In the linear case, we notice that CASTOR and DYNOTEARS outperform all the baselines, either those designed for MTS with multiple regimes, such as RPCMCI and CD-NOD or the other methods that assumes stationarity (Table \ref{lineartable}).  We emphasize that CASTOR performs similarly to DYNOTEARS, even though the latter benefits from prior access to ground truth regime partitions by being trained separately on each pure regime. \\
In the non-linear case, CASTOR outperforms all the baseline on instantaneous link, Figure \ref{fig:1}. PCMCI+ and Rhino w/o hist performs better than CASTOR in inferring time lagged links, however it is worth to note that these algorithms has access to the ground truth regime partition while CASTOR learns the number of regimes, their indices and the corresponding DAGs. 
\begin{table}[t]
\centering
\caption{Average F1 scores for different models on \textcolor{teal}{linear SEMs} with $d=40$ nodes. $K$ indicates the number of regimes, \textit{Split} denotes whether regime separation is automatic (A) or manual (M), and \textit{Type} classifies the graph as window (W) or summary (S). \textit{Inst.} refers to instantaneous links, and \textit{Lag} to time-lagged edges.}
    \resizebox{\columnwidth}{!}{
    \begin{tabular}{  *{3}{l}*{4}{l}}
    \toprule
      &  & &  \multicolumn{2}{c}{$K=3$} & \multicolumn{2}{c}{$K=4$}  \\ \midrule
    Model &  Split & Type &  Inst.&Lag & Inst.&Lag  \\ \midrule
     VARLINGAM  &  M&W&   \multicolumn{1}{c}{$9.83$}  & \multicolumn{1}{c}{$1.13$}          & $10.9$                          & $1.43$  \\
     Rhino     & M & W   & \multicolumn{1}{c}{$0.00$ }       & \multicolumn{1}{c}{$20.8$}                   & $0.00$ &$22.8$  \\
     Rhino w/o hist    & M & W &  \multicolumn{1}{c}{$0.00$ }         & \multicolumn{1}{c}{$38.4$}                   & $0.00$                                                                    & $39.1$ \\
     PCMCI+   & M& W & \multicolumn{1}{c}{$54.1$} & \multicolumn{1}{c}{$84.6$}        & $53.7$                                                                   & $86.1$         \\
      DYNOTEARS  & M & W & \multicolumn{1}{c}{\underline{97.4}}& \multicolumn{1}{c}{\underline{98.8}} & \underline{97.3}  & \underline{97.9} \\
      \midrule
     RPCMCI  & A & W & -     & \multicolumn{1}{c}{$18.4$}                             & \multicolumn{1}{c}{-}                                            & -  \\
     \rowcolor{gray!20} CASTOR  & A & W & \multicolumn{1}{c}{\textbf{98.2}}& \multicolumn{1}{c}{\textbf{99.8}} & \textbf{98.3}  & \textbf{98.9} \\
    \midrule
     CD-NOD  & A& S & \multicolumn{2}{c}{         11.3            }  & \multicolumn{2}{c}{ 5.57   }  \\
     \rowcolor{gray!20} CASTOR  &A&S& \multicolumn{2}{c}{\textbf{99.8}}   & \multicolumn{2}{c}{\textbf{99.2}}  \\
     \bottomrule
    \end{tabular}
    }
\label{lineartable}
\end{table}
\begin{table}
\caption{F1 Scores across IT data.}
\begin{adjustbox}{center}
{\fontsize{8pt}{6pt}\selectfont 
\begin{tabular}{@{}lllll@{}}
\toprule
Model     & Graph type & F1 Reg1       & F1 Reg 2         \\ \midrule
PCMCI+      &W& 12.1         & \textbf{29.6}              \\
DYNOTEARS &W& 18.2    & \textbf{28.5}                      \\ 
Rhino &W& 28.6    & 25.8                      \\ 
CASTOR    &W& 18.2          & \textbf{28.5}  \\ 
CASTOR non-lin    &W& \textbf{40.0}          & 24.5  \\ \midrule
CD-NOD & S& \multicolumn{2}{c}{23.5}\\
CASTOR non-lin & S& \multicolumn{2}{c}{\textbf{36.8}}\\
\bottomrule
\end{tabular}}
\end{adjustbox}
\vspace{-0.em}
\vspace{-0.5em}
\label{table2}
\end{table}
In both scenarios (linear and non-linear) RPCMCI struggles to achieve convergence, particularly in  settings with more than 3 different regimes due to its assumption of only inferring time-lagged relations. The comparison with CD-NOD on graph learning and also regime detection show that CASTOR outperforms CD-NOD, which is understandable because, CD-NOD learns one summary graph for the whole MTS and also expects only a few variables of the graph to be affected by the regime change (This assumption may not hold true in real scenarios such as epileptic seizures or climate science). However, CASTOR does not have this assumption and also learns one graph per regime (more details on Appendices \ref{appcomp_cd_nod} and \ref{app_reg_detect}.). Although our settings are identifiable, PCMCI+ infers a Markov equivalent class for the instantaneous links, which explains its performance deterioration in instantaneous relations and with a higher number of nodes. Rhino faces challenges in the absence of historical dependent noises (as confirmed by Figure 4 on page 24 of the Rhino paper). 
Moreover, Rhino utilizes ConvertibleGNN with Normalizing flows to learn the causal graphs. To train this model, a minimum of 50 time series of length 200 (10000 samples), all sharing the same causal graph is needed. Our ablation studies in Appendix \ref{ablation} highlight that neither the size of the window nor the minimum regime duration impact CASTOR performance on both regime detection and causal graph inferring. Additional results and evaluations using other metrics, such as SHD, that confirm the above findings are available in Appendix \ref{applinexp} and \ref{appnonlin}. The comparison with CD-NOD \citep{huang2020causal} and KCP \citep{arlot2019kernel} on the regime detection task is presented in Appendix \ref{app_reg_detect}.
\begin{figure*}[t]
    \centering
    \begin{subfigure}[b]{0.95\columnwidth}
       \centering

\includegraphics[width=0.95\linewidth]{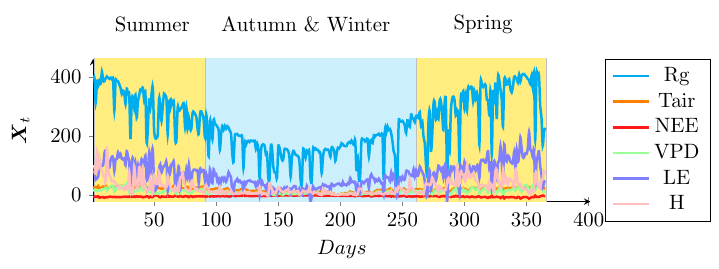}
\caption{}
\label{fig:pure_imp}

    \end{subfigure}
    \hfill
    \begin{subfigure}[b]{0.9\columnwidth}
    \centering
    \includegraphics[width=0.9\linewidth]{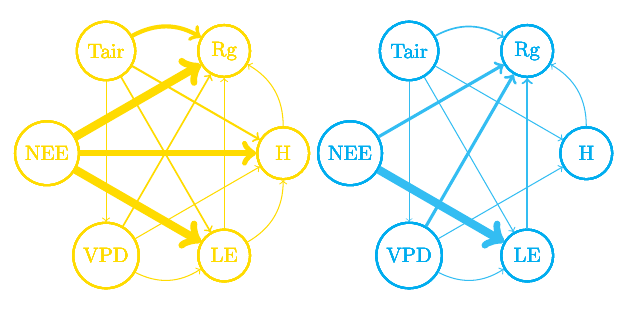}
    \caption{}
\label{fig:pi_alignement}
    \end{subfigure}
    \vspace{-5pt}
    \caption{CASTOR's results on Biosphere-Atmosphere data.(a) CASTOR identifies two regimes, months with hot weather colored in yellow and other with cold weather colored in blue. (b) The instantaneous links are for the two regimes, with the blue graph corresponding to the blue regime, cold weather, and the yellow one to the yellow regime, hot weather.\looseness=-1
    }
    \vspace{-12pt}
    \label{fig:overall_figure}
\end{figure*}
\subsection{Web activity dataset}
We now evaluate CASTOR on two stacked IT monitoring time series datasets, firstly introduced by \citet{bystrova2023causal} (details and analysis of the challenges of these datasets are in Appendix \ref{webdata}), each comprising 1106 timestamps and 7 nodes, sourced from EasyVista.  The web activity data is challenging, because it could present missing values, misaligned time series and partially sleeping time series due to inactivity of certain servers. IT experts are not sure that this data satisfies causal sufficiency assumption. \citet{ait2023case} present a study on the performance of causal discovery method on this data and shows the same range of performance. We compared our method to a subset of the best models mentioned in the benchmark subsection. As it is evident from Table \ref{table2}, CASTOR proficiently identifies the exact number of regimes and their indices. On regime 2, CASTOR (with linear relationships), PCMCI+ and DYNOTEARS outperform Rhino and CASTOR non-lin (uses NN for non linear relationships). However, in regime 1, CASTOR non-linear and Rhino outperform all other models. This superiority can be attributed to the fact that CASTOR non-lin and Rhino employs NNs to learn causal relationships, and the non-linearity in regime 1 complicates graph learning for the other models. CASTOR outperforms CD-NOD in this real world settings due to the fact that CD-NOD assumes that only few variables change from one regime to the other.\looseness=-1
\subsection{Biosphere–Atmosphere data}
\label{biosphere}
We apply CASTOR to biosphere‐atmosphere data from the FLUXNET dataset (San Luis site, Argentina \citep{sanluis}) to demonstrate its real‐world utility. \citet{krich2021functional} used these data to learn causal relations among six variables, global radiation ($R_g$), air temperature ($T_{air}$), net ecosystem exchange (NEE), vapor pressure deficit (VPD), sensible heat (H), and latent heat flux (LE), by running PCMCI+ on three‐month windows (with one overlapping month). Our goal is to automate causal graph learning and regime partitioning. Rather than assume a fixed three month interval, we provide one year of MTS to CASTOR and let it determine when the causal structure changes and which months are similar. Starting with non-overlapping three‐month windows (four initial regimes) and a minimum regime duration of two months, CASTOR converged to two regimes after a few iterations: one grouping Autumn and Winter (April to September) and another for Spring and Summer. The described partition is achieved with an accuracy of 85.4\%.\\
An initial observation is all variables appear as parents of global radiation, a counterintuitive result due to causal sufficiency assumption. CASTOR interpret global radiation as net radiation. Mathematically, we have
\[
R_n = R_g - SW_{\uparrow} + LW_{\downarrow} - LW_{\uparrow},
\]
where \(R_n\) is net radiation, \(R_g\) is global (shortwave) radiation, \(SW_{\uparrow} = \alpha R_g\) is reflected shortwave radiation, \(LW_{\uparrow}\) is longwave radiation emitted from the surface, and \(LW_{\downarrow}\) is incoming longwave radiation. Additionally, net radiation satisfies
\[
R_n = H + LE + G,
\]
with \(H\) as sensible heat, \(LE\) as latent heat flux, and \(G\) as ground heat flux. Causal sufficiency, where we assume that all causal variables are observed, causes CASTOR to infer incorrect link directions. For additional analysis, see Appendix~\ref{app_biosphere}.

\section{CONCLUSION}

We present CASTOR, a method for learning causal relationships from MTS with multiple regimes. CASTOR learns the number of regimes, their indices, and infers causal graphs for each regime simultaneously. It outperforms existing causal discovery models in handling both linear and non-linear relationships across multiple regimes on synthetic and real datasets.
\newpage
\section*{ACKNOWLEDGMENT}
We thank Ali Mourtada, Nikolaos Dimitriadis, Alessandro Favero, Guillermo Ortiz-Jimenez, Anas Essounaini, Thibault
Séjourné for helpful feedbacks and comments. This work was supported by the SNSF Sinergia
project ‘PEDESITE: Personalized Detection of Epileptic Seizure in the Internet of Things (IoT) Era’
\bibliography{ref}

\newpage
\section*{Checklist}

The checklist follows the references. For each question, choose your answer from the three possible options: Yes, No, Not Applicable.  You are encouraged to include a justification to your answer, either by referencing the appropriate section of your paper or providing a brief inline description (1-2 sentences). 
Please do not modify the questions.  Note that the Checklist section does not count towards the page limit. Not including the checklist in the first submission won't result in desk rejection, although in such case we will ask you to upload it during the author response period and include it in camera ready (if accepted).

\textbf{In your paper, please delete this instructions block and only keep the Checklist section heading above along with the questions/answers below.}

 \begin{enumerate}

 \item For all models and algorithms presented, check if you include:
 \begin{enumerate}
   \item A clear description of the mathematical setting, assumptions, algorithm, and/or model. [Yes] 
   \item An analysis of the properties and complexity (time, space, sample size) of any algorithm. [Yes, Appendix \ref{ablation} and \ref{runtime}]
   \item (Optional) Anonymized source code, with specification of all dependencies, including external libraries. [Yes, our model takes few minutes, we encourage the reviewers to run different experiments (to reproduce the results) using the notebooks provided in supplementary material]
 \end{enumerate}

 \item For any theoretical claim, check if you include:
 \begin{enumerate}
   \item Statements of the full set of assumptions of all theoretical results. [Yes]
   \item Complete proofs of all theoretical results. [Yes, Appendix \ref{theoproof} and \ref{app_proof2}]
   \item Clear explanations of any assumptions. [Yes, main text and Appendix \ref{theoproof} and \ref{app_proof2}]     
 \end{enumerate}

 \item For all figures and tables that present empirical results, check if you include:
 \begin{enumerate}
   \item The code, data, and instructions needed to reproduce the main experimental results (either in the supplemental material or as a URL). [Yes]
   \item All the training details (e.g., data splits, hyperparameters, how they were chosen). [Yes]
         \item A clear definition of the specific measure or statistics and error bars (e.g., with respect to the random seed after running experiments multiple times). [Yes]
         \item A description of the computing infrastructure used. (e.g., type of GPUs, internal cluster, or cloud provider). [Yes]
 \end{enumerate}

 \item If you are using existing assets (e.g., code, data, models) or curating/releasing new assets, check if you include:
 \begin{enumerate}
   \item Citations of the creator If your work uses existing assets. [Yes]
   \item The license information of the assets, if applicable. [Not Applicable]
   \item New assets either in the supplemental material or as a URL, if applicable. [Not Applicable]
   \item Information about consent from data providers/curators. [Not Applicable]
   \item Discussion of sensible content if applicable, e.g., personally identifiable information or offensive content. [Not Applicable]
 \end{enumerate}

 \item If you used crowdsourcing or conducted research with human subjects, check if you include:
 \begin{enumerate}
   \item The full text of instructions given to participants and screenshots. [Not Applicable]
   \item Descriptions of potential participant risks, with links to Institutional Review Board (IRB) approvals if applicable. [Not Applicable]
   \item The estimated hourly wage paid to participants and the total amount spent on participant compensation. [Not Applicable]
 \end{enumerate}

 \end{enumerate}
\newpage

\input{supplement}

\end{document}

%% file: supplement.tex
%
%




%

%

\newpage
\onecolumn
\appendix
\section*{Table of content}
\DoToC 
\clearpage
\section{DETAILED RELATED WORK}
\label{relatedwork}
Causal structure learning has been a hot research topics, \citet{hasan2023survey} propose a survey on causal discovery from IID data and time series. For IID data, Some approaches rely on conditional independence to infer causal relationships from observational data. A classic example of such approach is the PC algorithm \citep{spirtes2000causation}. In addition to approaches that work with observational data, there are also methods that support interventional data (COmbINE \citep{triantafillou2015constraint} and HEJ \citep{hyttinen2014constraint}). These methods offer insights into causal relationships based on data acquired through controlled interventions.\\
A novel line of research introduced by \citet{zheng2018dags} has sought to address the combinatorial problem of structure learning by formulating it as a continuous constrained optimization problem. By adopting this approach, they successfully circumvent the need for computationally intensive combinatorial search methods. Similarly, \citet{zhu2019causal} leverage the acyclicity constraint in their work but employ reinforcement learning techniques as a search strategy to estimate the DAG. In contrast, \citet{ke2019learning} focus on learning a DAG from interventional data through the optimization of an unconstrained objective function. \citet{brouillard2020differentiable} have undertaken a comprehensive investigation into the application of continuous-constrained approaches in the context of interventions, providing a general framework for their utilization. Another notable approach, DiBS by \citet{lorch2021dibs}, aims to infer a full posterior distribution over Bayesian networks given limited available observations. This approach enables the quantification of the uncertainty and the estimation of confidence levels of the structure learning procedure.\\

The aforementioned state-of-the-art methods have primarily been applied in the context of independent observations over time. \citet{assaad2022survey} offer an extensive survey for learning temporal causal relationships. However, when it comes to modeling time-dependent causal relationships, researchers have introduced and utilized Dynamic Bayesian Networks (DBNs). DBNs allow for the modeling of discrete-time temporal dynamics within directed graphical models. In certain approaches, contemporaneous dependencies are disregarded, and the focus is solely on recovering time-lagged relationships. Examples of such approaches include the works of \citet{haufe2010sparse}, \citet{song2009time}, and the algorithm tsFCI \citep{entner2010causal} adapts the Fast Causal Inference \citep{spirtes2000causation} algorithm (developed for the causal analysis of non-temporal variables) to infer causal relationships from time series data. \citet{runge2019detecting} proposed a two-stage algorithm PCMCI that can scale to large time series. However, these methods primarily emphasize the identification of relationships between variables at different time points, without explicitly considering contemporaneous relationships. \citet{runge2020discovering} present an extension of PCMCI called PCMCI+ that learns contemporary or instantaneous causal links. Another line of research targets the model with non-Gaussian instantaneous models, \citet{hyvarinen2010estimation} propose, VARLINGAM, a model that combines the non-Gaussian instantaneous models with autoregressive models and shows that a non-Gaussian model is identifiable without prior knowledge of network structure. Another approach, called Time-series Models with Independent Noise (TiMINo) \citep{peters2013causal} studies a class of restricted structural equation models (SEMs) for time-series data that include nonlinear and instantaneous effects. Recently, a novel study conducted by \citet{pamfil2020dynotears} has emerged, utilizing the algebraic characterization of acyclicity in directed graphs established by \citet{zheng2018dags}. Their work focuses on the learning of instantaneous and time-lagged graphs within time series data. To achieve this, they have developed a score-based approach for learning DBNs and employed an augmented lagrangien to optimize the resulting program. The resultant method, known as DYNOTEARS, offers the ability to learn causal graph of time dependent variable, without making implicit assumptions about the underlying graph topologies. By leveraging the algebraic characterization of acyclicity, DYNOTEARS enables the estimation of both instantaneous and time-lagged relationships in time series data. Instead of learning a full temporal causal graph, some methods like NBCB \citep{assaad2021mixed} or Noise-based/Constraint-based approach learns a summary causal graph from observational time series data without being restricted to the Markov equivalent class even in the case of instantaneous relations.\\

While DYNOTEARS, Rhino \citep{gong2022rhino}, PCMCI+, VARLINGAM successfully learn both instantaneous and time-lagged relationships from time series data, it is important to note that the method assumes stationarity and a single regime for the data. However, in numerous real-world scenarios, time series data may exhibit non-stationarity or be composed of multiple regimes, where the causal relationships are different in each regime. This presents a significant challenge for causal discovery.\\
Some research have aimed to address this challenge by developing methods for causal discovery in heterogeneous data. An example of such a method is CD-NOD developed by \citet{huang2020causal}, tackles time series with various regimes. By using the time stamp IDs as a surrogate variable, CD-NOD output one summary causal graph where the parents of each variable are identified as the union of all its parents in graphs from different regimes. Then it detects the change points by using a non stationary driving force that estimates the variability of the conditional distribution $p(x_i|\text{union parents of } x_i)$ over the time index surrogate. While CD-NOD provides a summary graph capturing behavioral changes across regimes, it falls short in inferring individual causal graphs. The overall summary graph does not effectively highlight changes between regimes. Additionally, CD-NOD detects the change points but fails to determine the regime indices, rendering it incapable of inferring the precise number of regimes. In scenarios involving recurring regimes, CD-NOD is unable to detect this crucial information. Another relevant work dealing with MTS composed of multiple regimes is RPCMCI \citep{saggioro2020reconstructing}. In this approach, \citet{saggioro2020reconstructing} learn a temporal graph for each regime. However, they focus initially on inferring only time-lagged relationships and require prior knowledge of the number of regimes and transitions between them. \citet{balsellsidentifiability} addresses first-order regime-dependent causal discovery from MTS with multiple regimes. They proved that first-order Markov switching models with non-linear Gaussian transitions are identifiable up to permutations. Their work offers also a practical algorithms for regime-dependent causal discovery in time series data. However, its primary limitation is the assumption of solely time-lagged relationships, with the theory being restricted to a single time lag. 
\newpage
\section{CASTOR FRAMEWORK: INTUITION}
CASTOR represents a causal discovery framework tailored for Multivariate Time Series (MTS), composed of different regimes. Each regimes can be treated as an independent MTS. Additionally, it is crucial to note that the number of lags $L$ always remains below the minimum length of the regimes $\zeta$.

\begin{figure}[H]
    \centering
    \vspace{-0.8em}
    \includegraphics[scale=0.68]{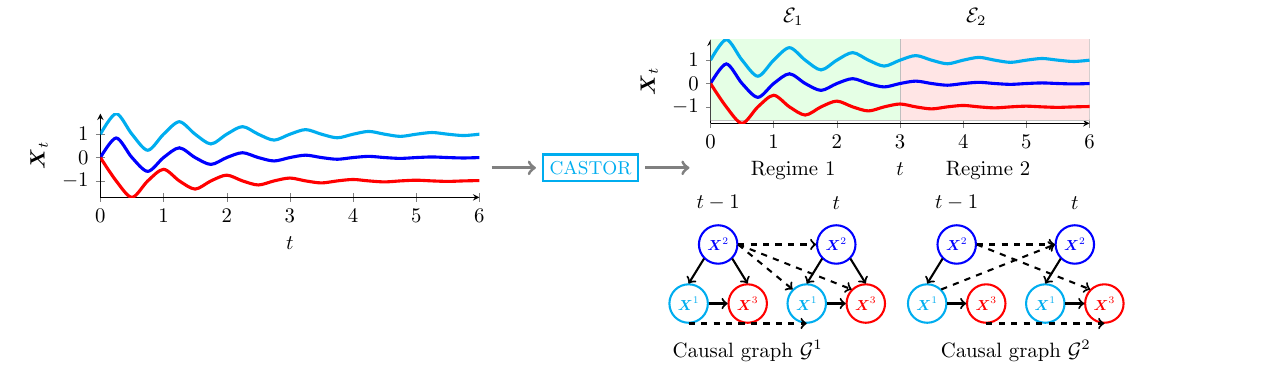}
    \vspace{-1em}
    \caption{Overview of CASTOR: This illustration demonstrates that CASTOR relies on the MTS to infer the number of regimes (equal to 2 in this figure), the regime partition ($\mathcal{E}_1$ for the first regime and $\mathcal{E}_2$ for the second) and  learn the temporal causal graphs ($\mathcal{G}^1$ for the first regime and $\mathcal{G}^2$ for the second). Dashed edges symbolize time-lagged links, while normal arrows represent instantaneous links.}
    \vspace{-0.5em}
    \label{castor_over_new}
\end{figure}

Figure (\ref{castor_over_new}) illustrates a MTS on its left side comprising three variables and two unknown distinct regimes. Each regime possesses its temporal DAG, with one lag attributed to each in this demonstrative scenario.

Upon receiving the MTS as input, CASTOR engages in the process of discerning the number of regimes, determining the indices associated with each regime (indicating their commencement and conclusion), and inferring the temporal DAGs. The resultant DAGs facilitate the straightforward reconstruction of summary graphs encapsulating the entire MTS (CD-NOD output).

\begin{figure}[H]
    \centering
    \includegraphics[scale=0.24]{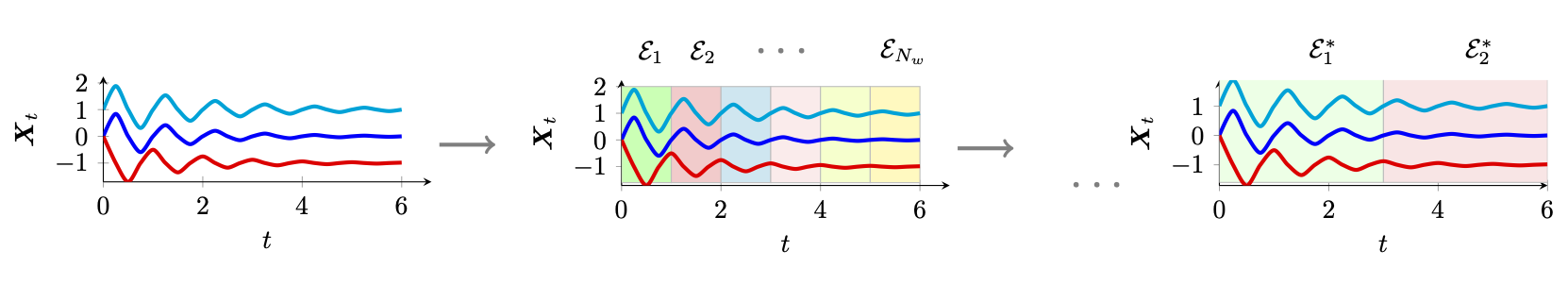}
    \vspace{-1.5em}
    \caption{CASTOR initially starts with $N_w$ equal windows, employing the M-step to learn the graph for the initial regimes. The method alternates between updating regime indices in the E-step and inferring temporal causal graphs in the M-step until maximum iterations. Over iterations, some initial regimes disappear, resulting in $N_w$ gradually approaching $K$ (two in this figure).}
    \vspace{-1em}
    \label{castor_reg_lea}
\end{figure}

To elucidate the regime learning process, Figure (\ref{castor_reg_lea}) delineates the step-by-step procedure followed by CASTOR in determining the number of regimes and their corresponding indices. The process commences with CASTOR partitioning the MTS into equal windows. In the initial iteration, the length of each regime equals the window size, a user-specified hyperparameter.

Subsequently, CASTOR learns a temporal DAG for each regime. This involves solving an optimization problem, as outlined in Eq (\ref{1313}) for the linear case and Eq (\ref{eq1010}) for the non-linear scenario. Following graph acquisition, CASTOR updates the regime indices utilizing Eq (\ref{10}). Notably, CASTOR employs a filtering mechanism to eliminate regimes characterized by an insufficient number of samples. In practical terms, any regime with fewer samples than a defined hyperparameter, denoted as $\zeta$ (representing the minimum regime duration), is discarded.

In the event of regime elimination, samples from the discarded regimes are reallocated to the nearest regime in terms of probability. Specifically, if the discarded regime is denoted as $u$, the sample $\boldsymbol{x}_t$ will be assigned to regime $v$ in the subsequent iteration, where $v$ is the regime with the highest $\gamma_{t,v}$.
\newpage
\section{EXPECTATION-MAXIMIZATION DERIVATION}
\label{appA}
In this section, we shall elucidate the computational details surrounding the resolution of our optimization problem. Specifically, we will provide clarity on the various equations introduced in Section 3, namely, Eq (\ref{Mstep}, \ref{1313}, \ref{10}, \ref{eq1010}).\\
\textbf{E-step. } We model regime participation through a binary latent variable $z_t \in \mathbf{R}^{N_w}$; $\boldsymbol{x}_t$ belongs to regime $u \Rightarrow z_{t,u} = 1$.
\begin{equation}
\begin{aligned}
\gamma_{t, u} & =p\left(z_{t, u}=1 \mid \boldsymbol{x}_t,\boldsymbol{x}_{<t}, \boldsymbol{G}^{u}_{\{0: L\}}\right) \\
& =\frac{p\left(z_{t, u}=1\right) p\left(\boldsymbol{x}_t \mid \boldsymbol{x}_{<t},z_{t, u}=1, \boldsymbol{G}^{u}_{\{0: L\}}\right)}{\sum_{j=1}^{N_w} p\left(z_{t, j}=1\right) p\left(\boldsymbol{x}_t \mid \boldsymbol{x}_{<t}, z_{t, j}=1, \boldsymbol{G}^{j}_{\{0: L\}}\right)} \\
& =\frac{\pi_{t, u}(\alpha) f^{u}\left(\boldsymbol{x}_t\right)}{\sum_{j=1}^{N_w} \pi_{t, j}(\alpha) f^{j}\left(\boldsymbol{x}_t\right)}
\end{aligned}
\end{equation}
\textbf{M-step. }Having estimated probabilities $\gamma_{t, u}$ in the E-step, we can now maximise the expected posterior distribution given the MTS $(\boldsymbol{x}_t)_{t \in \mathcal{T}}$ and we have:
\begin{equation}
\begin{aligned}
&\sup _{\theta,\alpha} \frac{1}{|\mathcal{T}|} \sum_{u=1}^{N_{w}} \sum_{t = 0}^{|\mathcal{T}|} \gamma_{t, u} \log 
 \pi_{t, u}(\alpha) f^{u}\left(\boldsymbol{x}_t\right) -\lambda|\mathcal{G}^u|, \text{  s.t } \boldsymbol{G}^u_0 \text{ is a DAG,}\\
& \iff \left\{\begin{array}{l}
\max _\alpha \frac{1}{|\mathcal{T}|}\sum_{u=1}^{N_{w}} \sum_{t=1}^{|\mathcal{T}|} \gamma_{t, u} \ln \left(\pi_{t, u}(\alpha)\right) \\
\sup _\theta \frac{1}{|\mathcal{T}|} \sum_{u=1}^{N_{w}} \sum_{t \in \mathcal{E}_u} \log f^{u}\left(\boldsymbol{x}_t\right) -\lambda|\mathcal{G}^u|, \text{  s.t } \boldsymbol{G}^u_0 \text{ is a DAG,}
\end{array}\right.
\end{aligned}
\end{equation}
We know $f^{u}\left(\boldsymbol{x}_t\right) = \mathcal{N}\left(\boldsymbol{x}_t \boldsymbol{G}_0^u+\sum_{\tau=1}^L \boldsymbol{x}_{t-\tau} \boldsymbol{G}_\tau^u, I\right)$, hence:
\begin{equation}
\begin{aligned}
&\iff\sup _\theta \frac{1}{|\mathcal{T}|} \sum_{u=1}^{N_{w}} \sum_{t \in \mathcal{E}_u} \log f^{u}\left(\boldsymbol{x}_t\right) -\lambda|\mathcal{G}^u|, \text{  s.t } \boldsymbol{G}^u_0 \text{ is a DAG,}\\
&\iff\min _{\theta} \frac{1}{|\mathcal{T}|} \sum_{u=1}^{N_{w}} \sum_{t \in \mathcal{E}_u}\left\|\boldsymbol{x}_t- \left( \boldsymbol{x}_t\boldsymbol{G}^u_0+\sum_{\tau=1}^L \boldsymbol{x}_{t-\tau}\boldsymbol{G}^u_\tau \right)\right\|_F^2 + \lambda|\mathcal{G}^u|, \text{  s.t } \boldsymbol{G}^u_0 \text{ is a DAG}\\
   &\iff \min _{\theta} \frac{1}{|\mathcal{T}|} \sum_{u=1}^{N_w} \sum_{t=1}^{|\mathcal{T}|} \gamma_{t, u}\left\|\boldsymbol{x}_t- \left( \boldsymbol{x}_t\boldsymbol{G}^u_0+\sum_{\tau=1}^L \boldsymbol{x}_{t-\tau}\boldsymbol{G}^u_\tau \right)\right\|_F^2 + \lambda|\mathcal{G}^u|, \text{  s.t } \boldsymbol{G}^u_0 \text{ is a DAG}\\
   & \iff \min _{\theta} \frac{1}{|\mathcal{T}|} \sum_{u=1}^{N_w} \sum_{t=1}^{|\mathcal{T}|} \gamma_{t, u}\left\|\boldsymbol{x}_t- \left( \boldsymbol{x}_t\boldsymbol{G}^u_0+\sum_{\tau=1}^L \boldsymbol{x}_{t-\tau}\boldsymbol{G}^u_\tau \right)\right\|_F^2 + \lambda|\mathcal{G}^u|+\frac{\rho}{2} h\left(\boldsymbol{G}^u_0\right)^2 
+\alpha h\left(\boldsymbol{G}^u_0\right),
\end{aligned}
\end{equation}
The only difference between the linear and the non linear cases is how we estimate the mean of the normal distribution $f^u$ for every regime $u$. As we mentioned in section 3.2, we estimate these means using NNs and we have $f^{u}_i\left(\boldsymbol{x}_t\right) = \mathcal{N}\left(\psi^u_i\left(\phi^u_i(\boldsymbol{x}_t),\phi^{u,\text{lag}}_i(\boldsymbol{x}_t^{\text{lag}})\right), 1\right)$, Hence, our M-step for non-linear CASTOR:
\begin{equation}
\begin{aligned}
\min _{\theta,\mathcal{G}}  \frac{1}{|\mathcal{T}|} \sum_{u=1}^{N_w} \sum_{t=1}^{|\mathcal{T}|}\sum_{i=1}^d   \gamma_{t, u}\mathcal{L}(x^i_t, \psi^u_i\left(\phi^u_i(\boldsymbol{x}_t),\phi^{u,\text{lag}}_i(\boldsymbol{x}_t^{\text{lag}})\right))+\lambda|\mathcal{G}^u| + \frac{\rho}{2} h\left(\boldsymbol{G}^u_0\right)^2 
+\alpha h\left(\boldsymbol{G}^u_0\right)
\end{aligned}
\end{equation}
\section{COMPLEXITY, CONVERGENCE DISCUSSION AND LIMITATIONS}
\textbf{Convergence discussion. }We provided theoretical results on identifiability in Section~\ref{identi_res}, which 
is a key statistical property to ensure that the causal discovery problem is well-defined, 
as is standard in most causal discovery papers \citep{brouillard2020differentiable, 
gong2022rhino, pamfil2020dynotears,balsellsidentifiability}. Although obtaining convergence 
rates or finite-data bounds is undoubtedly interesting, it is extremely challenging due 
to the non-convexity of the acyclicity constraint in an EM procedure. In fact, we are not aware of any results applicable to the case. However, we 
 empirically demonstrate that CASTOR converges in both linear and non-linear cases, correctly 
identifying the number of regimes and their indices, while also learning the corresponding 
DAGs.\\

\textbf{Complexity. }We know that the time complexity of one regime is $\mathcal{O}\left(d^3\right)$, where $d$ is the number of nodes, because of the computation of the acyclicity constraints. The complexity of CASTOR per iteration is $\mathcal{O}\left(|\mathcal{T}| N_w d^3\right)$ where $|\mathcal{T}|$ the number of samples and $N_w$ is the number of regimes at each iteration.\\

\textbf{Limitations. }A fundamental limitation of this work is the assumption that Gaussian additive noise with equal variance. Many real world scenarios have either a non Gaussian noise and different variance for different MTS components. Future research could address this by extending the analysis to non-Gaussian noise models. Another valuable direction would be to conduct a statistical analysis to understand how the number of samples affects the convergence of CASTOR to the ground truth regimes. Additionally, extending the framework to account for the presence of confounders represents an important avenue for future work.
\section{FURTHER EXPERIMENTAL RESULTS}
\label{appexp}
\subsection{Synthetic data}
\label{appsyntheticdata}
We employ the Erdos–Rényi (ER) \citep{newman2018networks} model with mean degrees of 1 or 2 to generate lagged graphs, and the Barabasi–Albert (BA) \citep{barabasi1999emergence} model with mean degrees 4 for instantaneous graphs. The maximum number of lags, $L$, is set at 1 or 2. We experiment with varying numbers of nodes $\{5,10,20,40\}$ and different numbers of regimes $\{2,3,4,5\}$, each representing diverse causal graphs. The length of each regime is randomly sampled from the set $\{300,400,500,600\}$.
\begin{itemize}
    \item \textbf{Linear case. } Data is generated as follows:
    $$
\forall u \in \{1,...,K\}, \forall t \in \mathcal{E}_u: \boldsymbol{x}_t=\boldsymbol{x}_t\boldsymbol{G}^u_0+\sum_{\tau=1}^L \boldsymbol{x}_{t-\tau}\boldsymbol{G}^u_\tau + \boldsymbol{\epsilon}_t,
$$
with $\boldsymbol{G}^u_0$ is adjacency matrix of the generated graph by BA model, $\forall \tau \in \{1,..,L\}: \boldsymbol{G}^u_\tau$ are the adjacency of the time lagged graphs generated by ER and $\epsilon_t \sim \mathcal{N}(0, I)$, follows to a normal distribution.
\item \textbf{Non-linear case. } The formulation used to generated the data is:$$
\forall u \in \{1,...,K\}, \forall t \in \mathcal{E}_u:  x_t^i= g^u_i\left(\mathbf{P a}_{\mathcal{G}^u}^i(<t), \mathbf{P a}_{\mathcal{G}^u}^i(t)\right)+\epsilon_t^i,
$$
where $g^u_i$ is a general differentiable linear/non-linear function and $\epsilon_t^i \sim \mathcal{N}(0, 1)$, follows a normal distribution. The function $g^u_i$ is a random combination between a linear transformation and a randomly chosen function from the set: $\{\texttt{Tanh, LeakyReLU, ReLU}\}$.
\end{itemize}
\subsection{Web activity data}
\label{webdata}
The web activity dataset \url{https://github.com/ckassaad/causal_discovery_for_time_series} has the following variables:  NetIn that represents the data received by the network interface card in Kbytes/second; NPH represents the number of HTTP processes; NPP represents the number of PHP processes; NCM represents the number of open MySql connections which are started by PHP processes, CpuH represents the percentage of CPU used by all HTTP processes; CpuP represents the percentage of CPU used by all PHP processes; CpuG represents the percentage of global CPU usage.\\

This dataset is challenging because it is collected from multiple sources resulting in a misaligned time series. Also the presence of partially sleeping time series (NetIn and NPP) due to inactivity of certain servers. The low sampling rate and also the presence of missing values (CpuG) can complicate the inference of causal relationships. Also, the experts in the field of IT systems are uncertain whether IT data, such as web activity data, satisfy the causal sufficiency assumption.\\

\citep{ait2023case} presented a study about the challenges presented by IT monitoring time series, their results shows that the causal discovery method suffers in these scenarios due to aforementioned challenges.

\subsection{Baselines}
All used benchmarks for the synthetic experiments are run by using publicly available libraries:
VARLINGAM \citep{hyvarinen2010estimation} is implemenented in the \texttt{lingam}\footnote{\url{https://lingam.readthedocs.io/en/latest/}} python package.
PCMCI+ \citep{runge2020discovering} and RPCMCI \citep{saggioro2020reconstructing} are implemented in \texttt{Tigramite}\footnote{\url{https://jakobrunge.github.io/tigramite/}} 
and DYNOTEARS \citep{pamfil2020dynotears} on \texttt{causalnex}\footnote{\url{https://causalnex.readthedocs.io/en/latest/}} package. For Rhino we use the publicly available GitHub shared by the authors\footnote{\url{https://github.com/microsoft/causica/tree/v0.0.0}}. We fine tuned the parameters to achieve the optimal graph for each model.\\
For CASTOR, an edge threshold of 0.4 is selected. In the linear scenario, we establish $\zeta = 100$ as the minimum regime duration, while in the non-linear context, $\zeta$ is set at 200. To demonstrate the model's robustness to the choice of the window size, we train CASTOR using diverse window sizes, specifically $w = 200$ or $w=300$. For the sparsity coefficient, we use $\lambda = 0.05$. In order to optimise our M-step, we use L-BFGS-B algorithm \citep{zhu1997algorithm}.
\subsection{Ablation studies}
\label{ablation}
\begin{table}[h]
\caption{CASTOR ablation study (Linear case) with fixed window size $w = 300$, fixed number of nodes $d = 10$ and regime durations randomly sampled from \{400, 500, 600\}. We report running time and iterations per regime and per minimum regime duration $\zeta$. \textcolor{black}{CASTOR consistently achieves a 100\% F1 score in graph learning and 100\% regime accuracy.}}
\centering
\begin{tabular}{@{}ccc|cc|cc|@{}}
                             & \multicolumn{2}{|c|}{$K=2$}      & \multicolumn{2}{c|}{$K=3$}      & \multicolumn{2}{c|}{$K=4$}      \\ \midrule
\multicolumn{1}{c|}{$\zeta$} & Running time (R.T) & Iter. & R.T & Iter. & R.T & Iter. \\ \midrule
\multicolumn{1}{c|}{100}     & 0 min 42s    & 3                & 1 min 55s    & 4                & 2 min 32s    & 3                \\
\multicolumn{1}{c|}{200}     & 0 min 38s    & 3                & 1 min 50s    & 4                & 2 min 22s    & 3                \\
\multicolumn{1}{c|}{300}     & 0 min 37s    & 3                & 2 min 10s    & 4                & 2 min 27s    & 3                \\ \bottomrule
\end{tabular}
\label{lintime}
\end{table}

\begin{table}[h]
\caption{CASTOR ablation study (Linear case) with fixed minimum regime duration $\zeta = 90$, fixed number of nodes $d = 10$ and regime durations randomly sampled from \{400, 500, 600\}. We report running time and iterations per regime and per window size $w$. \textcolor{black}{CASTOR consistently achieves a 100\% F1 score in graph learning and 100\% regime accuracy.}}
\centering
\begin{tabular}{@{}ccc|cc|cc|@{}}
                             & \multicolumn{2}{|c|}{$K=2$}      & \multicolumn{2}{c|}{$K=3$}      & \multicolumn{2}{c|}{$K=4$}      \\ \midrule
\multicolumn{1}{c|}{$w$} & R.T & Iter. & Running time & Iter. & R.T & Iter. \\ \midrule
\multicolumn{1}{c|}{100}     & 1 min 31s    & 8                & 6 min 18s    & 8                & 7 min 57s    & 8                \\
\multicolumn{1}{c|}{150}     & 0 min 43s    & 5                & 6 min 04s    & 10               & 5 min 40s    & 5                \\
\multicolumn{1}{c|}{200}     & 0 min 45s    & 5                & 2 min 17s    & 5                & 5 min 35s    & 5                \\
\multicolumn{1}{c|}{250}     & 0 min 40s    & 5                & 1 min 59s    & 4                & 2 min 4s     & 4                \\ \bottomrule
\end{tabular}
\end{table}
\begin{table}[h]
\centering
\caption{CASTOR ablation study (Non-linear case) with fixed window size $w = 300$, fixed number of nodes $d = 10$ and regime durations randomly sampled from \{400, 500, 600\}. We report running time and iterations per regime and per minimum regime duration $\zeta$.}
{\fontsize{10pt}{9pt}\selectfont 
\begin{tabular}{@{}c|ccccc|ccccc|@{}}

                                                   & \multicolumn{5}{c|}{$K=2$}                          & \multicolumn{5}{c|}{$K=3$}                                                   \\ \midrule
\multicolumn{1}{c|}{$\zeta$}& R.T & Iter. & F1 Inst. & F1 Lag & Reg Acc. & R.T & Iter. & F1 Inst. & F1 Lag & Reg Acc.  \\ \midrule
\multicolumn{1}{c|}{100}                                              & 2'45s    & 5     & 92.5     & 73.5   & 95.1     & 4'27s    & 6     & 93.9     & 87.1   & 92.0         \\
\multicolumn{1}{c|}{200}                                               & 1'50s    & 4     & 92.5     & 73.5   & 96.2     & 2'44s    & 5     & 95.2     & 84.2   & 92.3      \\
\multicolumn{1}{c|}{300}                                                & 1'31s    & 4     & 92.5     & 73.5   & 96.2     & 2'48s    & 5     & 93.9     & 87.1   & 91.7        \\ \bottomrule
\end{tabular}}
\end{table}
\begin{table}[h]
\centering
\caption{CASTOR ablation study (Non-linear case) with fixed minimum regime duration $\zeta = 90$, fixed number of nodes $d = 10$ and regime durations randomly sampled from \{400, 500, 600\}. We report running time, iterations, F1 score for instantaneous and time lagged links and regime accuracy per regime and per minimum regime duration $\zeta$.}
{\fontsize{10pt}{9pt}\selectfont 
\begin{tabular}{@{}c|ccccc|ccccc|@{}}

                                                   & \multicolumn{5}{c|}{$K=2$}                          & \multicolumn{5}{c|}{$K=3$}                                                   \\ \midrule
\multicolumn{1}{c|}{$w$}& R.T & Iter. & F1 Inst. & F1 Lag & Reg Acc. & R.T & Iter. & F1 Inst. & F1 Lag & Reg Acc.  \\ \midrule
\multicolumn{1}{c|}{100}                                              & 11'30s    & 10     & 89.7     & 100.  & 97.4     & 11'50s    & 10     & 94.1     & 81.1   & 93.1        \\
\multicolumn{1}{c|}{200}                                               & 8'50s    & 10     & 85.1     & 90.1  & 96.7     & 5'58s    & 6     & 92.6     & 79.1   & 93.2      \\
\multicolumn{1}{c|}{250}                                                & 4'46s    & 7     & 89.7     & 100.   & 97.2     & 3'24s    & 4     & 92.1     & 78.0   & 92.1       \\ \bottomrule
\end{tabular}}
\label{nonlintime}
\end{table}
CASTOR demonstrates robustness in handling both linear, Table \ref{lintime}, and nonlinear causal relationships, Ta ble\ref{nonlintime}, regardless of the choice of minimum regime duration or window size. The primary impact of these parameters is on the number of iterations and the overall running time.
\begin{figure}[H]
    \centering
    \includegraphics[width=0.7\linewidth]{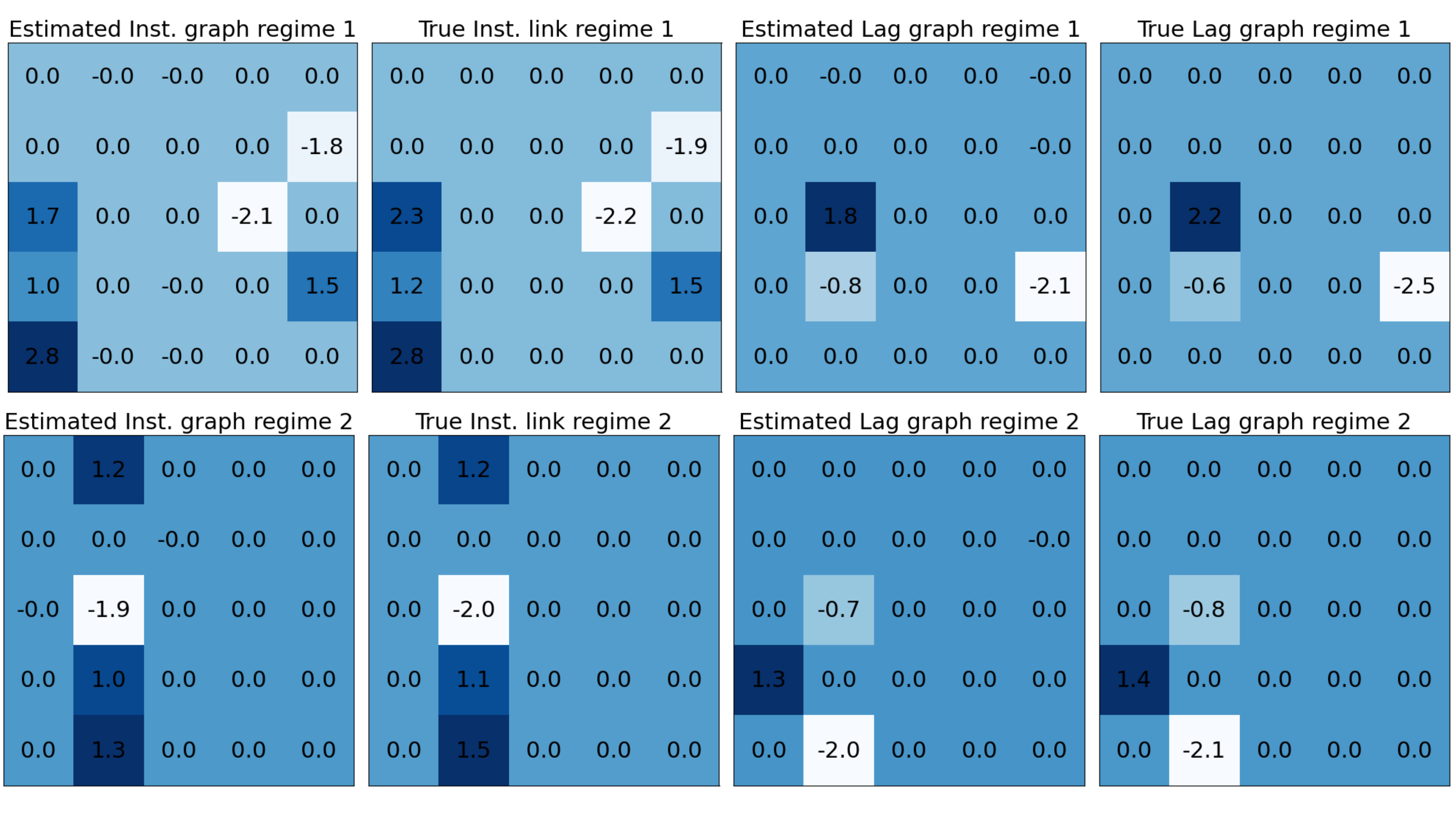}
    \caption{Illustration of CASTOR's estimated graphs compared to ground truth ones, for data generated with weighted DAGs.}
    \label{fig:wei_mat}
\end{figure}
Figure 3 illustrates that CASTOR effectively handles weighted adjacency matrices, accurately estimating both the edge weights and the presence of edges.
\subsection{Table 1 with standard deviation and $K=2$}
\begin{table}[H]
\centering
\caption{Average F1 Scores on regimes for different models and settings for linear causal relationships. Here, $d$ denotes the number of nodes, $K$ the number of regimes, ``Split'' specifies whether the regime split is automatic (A) or manual (M), and ``Type'' categorizes the graph as either a window (W) or summary (S). ``Inst.'' refers to instantaneous links and ``Lag'' to time-lagged edges.}
    \resizebox{\columnwidth}{!}{
    \begin{tabular}{  *{3}{l}*{6}{l} *{6}{l}}
    \toprule
          &\multicolumn{2}{c}{} & \multicolumn{6}{c}{$d=10$} &
       \multicolumn{6}{c}{$d=40$}\\ \midrule
      &  & & \multicolumn{2}{c}{$K=2$}& \multicolumn{2}{c}{$K=3$} & \multicolumn{2}{c}{$K=4$} &  \multicolumn{2}{c}{$K=2$}& \multicolumn{2}{c}{$K=3$} & \multicolumn{2}{c}{$K=4$}  \\ \midrule  
    Model &  Split & Type & Inst.&Lag& Inst.&Lag & Inst.&Lag &  Inst.&Lag& Inst.&Lag & Inst.&Lag  \\ \midrule
     VARLINGAM  &  M&W&\multicolumn{1}{c}{$18.2_{\pm3.8}$} &\multicolumn{1}{c}{$10.4_{\pm7.8}$}       & \multicolumn{1}{c}{$11.0_{\pm6.4}$}  & \multicolumn{1}{c}{$5.01_{\pm1.4}$}                         & \multicolumn{1}{c}{$9.70_{\pm1.8}$}     & \multicolumn{1}{c}{$5.10_{\pm3.1}$}         & \multicolumn{1}{c}{$8.40_{\pm 1.2}$} & \multicolumn{1}{c}{$1.2_{\pm 0.1}$}          & \multicolumn{1}{c}{$9.83_{\pm0.9}$}  & \multicolumn{1}{c}{$1.13_{\pm0.8}$}          & $10.9_{\pm3.1}$                          & $1.43_{\pm 0.6}$  \\
     Rhino     & M & W & \multicolumn{1}{c}{$62.4_{\pm10.}$} & \multicolumn{1}{c}{$40.7_{\pm 4.4}$}                  & \multicolumn{1}{c}{$54.8_{\pm3.9}$}   & \multicolumn{1}{c}{$44.4_{\pm 5.0}$}                                   & \multicolumn{1}{c}{$50.7_{\pm2.5}$} & \multicolumn{1}{c}{$43.0_{\pm 6.6}$}                     & \multicolumn{1}{c}{$0.00_{\pm 0.0}$ }         & \multicolumn{1}{c}{$21.1_{\pm2.1}$}                  & \multicolumn{1}{c}{$0.00_{\pm 0.0}$ }       & \multicolumn{1}{c}{$20.8_{\pm1.3}$}                   & $0.00_{\pm 0.0}$  \\                                                   Rhino w/o hist    & M & W & \multicolumn{1}{c}{$90.2_{\pm3.4}$} & \multicolumn{1}{c}{$58.5_{\pm 2.3}$}                  & \multicolumn{1}{c}{$87.5_{\pm3.6}$}   & \multicolumn{1}{c}{$53.5_{\pm 2.2}$}                                   & \multicolumn{1}{c}{$88.2_{\pm1.7}$} & \multicolumn{1}{c}{$61.8_{\pm 2.9}$}                     & \multicolumn{1}{c}{$0.00_{\pm 0.0}$ }         & \multicolumn{1}{c}{$44.1_{\pm1.2}$}                  & \multicolumn{1}{c}{$0.00_{\pm 0.0}$ }       & \multicolumn{1}{c}{$38.4_{\pm3.6}$}                   & $0.00_{\pm 0.0}$                                                                    & $39.1_{\pm1.3}$ \\                       
     PCMCI+   & M& W & \multicolumn{1}{c}{$81.1_{\pm 7.4}$} & \multicolumn{1}{c}{$89.5_{\pm 3.8}$}      & \multicolumn{1}{c}{$80.4_{\pm 4.5}$} & \multicolumn{1}{c}{$83.6_{\pm3.5}$}                        & \multicolumn{1}{c}{$77.7_{\pm 1.8}$}     & \multicolumn{1}{c}{$77.6_{\pm 3.2}$}     & \multicolumn{1}{c}{$55.2_{\pm 3.6}$}  & \multicolumn{1}{c}{$85.4_{\pm4.1}$}         & \multicolumn{1}{c}{$54.1_{\pm 5.4}$} & \multicolumn{1}{c}{$84.6_{\pm 2.6}$}        & $53.7_{\pm3.2}$                                                                   & $86.1_{\pm 1.8}$         \\
      DYNOTEARS  & M & W & \multicolumn{1}{c}{\underline{100.$_{\pm0.0}$}}& \multicolumn{1}{c}{\underline{100.$_{\pm0.0}$}} & \multicolumn{1}{c}{\underline{96.9$_{\pm2.5}$}}& \multicolumn{1}{c}{\underline{100.$_{\pm0.0}$}} & \multicolumn{1}{c}{\textbf{99.6$_{\pm0.9}$}}& \multicolumn{1}{c}{\textbf{99.8$_{\pm1.4}$}} & \multicolumn{1}{c}{\textbf{100.$_{\pm 0.0}$}}& \multicolumn{1}{c}{\underline{100.$_{\pm0.0}$}}  & \multicolumn{1}{c}{\underline{97.4$_{\pm1.7}$}}& \multicolumn{1}{c}{\underline{98.8$_{\pm0.2}$}} & \underline{97.3$_{\pm0.8}$}  & \underline{97.9$_{\pm0.6}$} \\
     RPCMCI  & A & W & \multicolumn{1}{c}{-}   & \multicolumn{1}{c}{$42.3_{\pm11.}$} & \multicolumn{1}{c}{-}                 & \multicolumn{1}{c}{$18.8_{\pm2.5}$}                          & \multicolumn{1}{c}{-}                        & \multicolumn{1}{c}{-}                         & \multicolumn{1}{c}{-}      &\multicolumn{1}{c}{$42.1_{\pm3.5}$}                   & -     & \multicolumn{1}{c}{$18.4_{\pm14.}$}                             & \multicolumn{1}{c}{-}                                            & -  \\ 
     CASTOR  & A & W & \multicolumn{1}{c}{\textbf{100.$_{\pm0.0}$}}& \multicolumn{1}{c}{\textbf{100.$_{\pm0.0}$}} & \multicolumn{1}{c}{\textbf{97.3$_{\pm2.5}$}}& \multicolumn{1}{c}{\textbf{100.$_{\pm0.0}$}} & \multicolumn{1}{c}{\textbf{99.3$_{\pm0.9}$}}& \multicolumn{1}{c}{\underline{98.0$_{\pm1.4}$}} & \multicolumn{1}{c}{\underline{98.3$_{\pm 1.7}$}}& \multicolumn{1}{c}{\textbf{100.$_{\pm0.0}$}}  & \multicolumn{1}{c}{\textbf{98.2$_{\pm1.2}$}}& \multicolumn{1}{c}{\textbf{99.8$_{\pm0.2}$}} & \textbf{98.3$_{\pm0.4}$}  & \textbf{98.9$_{\pm0.9}$} \\
    \midrule
     CD-NOD  & A& S & \multicolumn{2}{c}{20.2 } & \multicolumn{2}{c}{11.4 }  & \multicolumn{2}{c}{38.8}   &\multicolumn{2}{c}{0}  &\multicolumn{2}{c}{         11.3            }  & \multicolumn{2}{c}{ 5.57   }  \\
     CASTOR  &A&S& \multicolumn{2}{c}{\textbf{100}} & \multicolumn{2}{c}{\textbf{100}}   & \multicolumn{2}{c}{\textbf{97.9}}& \multicolumn{2}{c}{\textbf{100}} & \multicolumn{2}{c}{\textbf{99.8}}   & \multicolumn{2}{c}{\textbf{99.2}}  \\
     \bottomrule
    \end{tabular}
    }
\label{lineartable2}

\end{table}
Table \ref{lineartable2} presents the complete results for linear case. CASTOR and DYNOTEARS outperform
all the baselines, either those designed for MTS with
multiple regimes, such as RPCMCI and CD-NOD or
the other methods that assumes stationarity (Table
1). We emphasize that CASTOR performs similarly to
DYNOTEARS, even though the latter benefits from
prior access to ground truth regime partitions by being
trained separately on each pure regime.
\subsection{Violations}
In the main text, we introduced our method for multivariate time series that exhibit multiple regimes, each characterized by a distinct causal graph. To challenge this assumption, we created two scenarios:
\begin{itemize}
    \item \textbf{Scenario 1,} \textit{Causal Graph fixed, only the nonlinear mixing function changes}: CASTOR showed the same causal graph learning performance and also robustness to such scenarios. However, when the number of regime increases, CASTOR regime accuracy drops in this situation. But we want to clarify that this is a violation of the setting that CASTOR is designed to tackle.

\item \textbf{Scenario 2,} \textit{Causal graph fixed , changes in nonlinear function and the noise distribution (case of soft intervention)}: CASTOR showed the same causal graph learning performance and also robustness to such scenarios. In this case, when the number of regimes increases CASTOR succeeded to identify the different regimes, we want to highlight that in this setting we violate our assumption.
\end{itemize}
\begin{table}[ht]
\centering
\caption{Comparison of performance metrics for different regime configurations.}
\begin{tabular}{l|ccc|ccc}
\toprule
& \multicolumn{3}{c|}{\(\mathbf{K=2}\)} & \multicolumn{3}{c}{\(\mathbf{K=3}\)} \\
\cmidrule(lr){2-4}\cmidrule(lr){5-7}
& \textbf{F1 Inst} & \textbf{F1 lag} & \textbf{Reg Acc}
& \textbf{F1 Inst} & \textbf{F1 lag} & \textbf{Reg Acc} \\
\midrule
Scenario 1
& 80.0 & 81.1 & 97.1 
& 73.4 & 84.6 & 62.0 \\
Scenario 2 
& 81.8 & 88.8 & 98.1 
& 86.5 & 74.9 & 89.1 \\
\bottomrule
\end{tabular}
\label{tab:nonlinear_mixing}
\end{table}

\subsection{Further experiments and evaluation using SHD: Linear case}
\label{applinexp}
\begin{table}[H]
\centering
\caption{We report the average F1 Score on regimes for different models and settings for linear causal relationships. Here, $d$ denotes the number of nodes, $K$ indicates the number of regimes, ``Split" specifies whether the algorithm automatically splits the regimes (``A") or if the split was done manually beforehand (``M"), and ``Type" categorizes the type of returned graph as either window graph (``W") or summary graph (``S"). Inst. refers to instantaneous links and Lag to time-lagged edges.}
\resizebox{\columnwidth}{!}{
\begin{tabular}{*{3}{l}*{6}{l} | *{6}{l}}
\toprule
          & \multicolumn{2}{c}{} & \multicolumn{6}{c}{$d=10$} & \multicolumn{6}{c}{$d=40$} \\ \midrule
Model     & Split & Type & \multicolumn{2}{c}{$K=2$} & \multicolumn{2}{c}{$K=3$} & \multicolumn{2}{c}{$K=4$} & \multicolumn{2}{c}{$K=2$} & \multicolumn{2}{c}{$K=3$} & \multicolumn{2}{c}{$K=4$} \\ \midrule  
          &       &      & Inst. & Lag   & Inst. & Lag   & Inst. & Lag   & Inst. & Lag   & Inst. & Lag   & Inst. & Lag   \\ \midrule
Rhino            & M & W & \multicolumn{1}{c}{22}  & \multicolumn{1}{c}{25}  & \multicolumn{1}{c}{35}  & \multicolumn{1}{c}{43}  & \multicolumn{1}{c}{53}  & \multicolumn{1}{c}{60}  & \multicolumn{1}{c}{136} & \multicolumn{1}{c}{372} & \multicolumn{1}{c}{199} & \multicolumn{1}{c}{539} & 265 & 693 \\
Rhino w/o hist   & M & W & \multicolumn{1}{c}{5}   & \multicolumn{1}{c}{14}  & \multicolumn{1}{c}{9}   & \multicolumn{1}{c}{28}  & \multicolumn{1}{c}{11}  & \multicolumn{1}{c}{32}  & \multicolumn{1}{c}{136} & \multicolumn{1}{c}{189} & \multicolumn{1}{c}{208} & \multicolumn{1}{c}{340} & 273 & 420 \\
PCMCI+           & M & W & \multicolumn{1}{c}{16}  & \multicolumn{1}{c}{9}   & \multicolumn{1}{c}{18}  & \multicolumn{1}{c}{11}  & \multicolumn{1}{c}{24}  & \multicolumn{1}{c}{17}  & \multicolumn{1}{c}{29}  & \multicolumn{1}{c}{8}   & \multicolumn{1}{c}{53}  & \multicolumn{1}{c}{8}   & 72  & 12 \\
DYNOTEARS       & M & W & \multicolumn{1}{c}{\underline{0}} & \multicolumn{1}{c}{\underline{0}} & \multicolumn{1}{c}{\underline{3}} & \multicolumn{1}{c}{\underline{0}} & \multicolumn{1}{c}{\textbf{1}}      & \multicolumn{1}{c}{\textbf{1}}      & \multicolumn{1}{c}{\textbf{0}}      & \multicolumn{1}{c}{\underline{0}} & \multicolumn{1}{c}{\underline{11}} & \multicolumn{1}{c}{\underline{6}} & \underline{9} & \underline{4} \\
RPCMCI          & A & W & \multicolumn{1}{c}{-}   & \multicolumn{1}{c}{$38$}  & \multicolumn{1}{c}{-}   & \multicolumn{1}{c}{$48$}  & \multicolumn{1}{c}{-}   & \multicolumn{1}{c}{-}   & \multicolumn{1}{c}{-}   & \multicolumn{1}{c}{$155$} & -                    & \multicolumn{1}{c}{$451$} & \multicolumn{1}{c}{-}  & - \\
CASTOR          & A & W & \multicolumn{1}{c}{\textbf{$0$}} & \multicolumn{1}{c}{\textbf{$0$}} & \multicolumn{1}{c}{\textbf{1}}  & \multicolumn{1}{c}{\textbf{0}}  & \multicolumn{1}{c}{\textbf{1}}  & \multicolumn{1}{c}{\underline{2}} & \multicolumn{1}{c}{\underline{2}} & \multicolumn{1}{c}{\textbf{0}}  & \multicolumn{1}{c}{\textbf{9}}  & \multicolumn{1}{c}{\textbf{1}}  & \textbf{7} & \textbf{2} \\
\bottomrule
\end{tabular}}
\label{shdlinear}
\end{table}

\textcolor{black}{The evaluation using the SHD metric demonstrates consistent findings: CASTOR outperforms the baselines (Rhino, PCMCI+) even when we place these baselines in more favorable scenarios by providing them with the ground truth regime partition.}\\ 
\textcolor{black}{RPCMCI encounters challenges in our settings due to its assumption of only inferring time-lagged relations. This assumption makes our scenario more complex for this algorithm, impacting both regime learning and graph inference tasks.}\\
\textcolor{black}{Although our settings are identifiable, PCMCI+ infers a Markov equivalent class for the instantaneous links, which explains its performance deterioration, particularly in instantaneous relations.}\\
\textcolor{black}{Rhino, which is a state-of-the-art causal discovery method for nonlinear relationships, faces challenges in the absence of historical dependent noises (as confirmed by Figure 4 on page 24 of the Rhino paper). Moreover, Rhino utilizes ConvertibleGNN with Normalizing flows to learn the causal graphs. To train this model, a minimum of 50 time series of length 200 (10000 samples), all sharing the same causal graph. In contrast, our dataset consists of regimes that do not exceed 600 samples. This difference in dataset characteristics poses challenges for Rhino's performance in our scenario.}
\begin{table}[H]
\centering
\caption{We report the average F1 Score on regimes for different Models and Settings for linear causal relationships: $d$ denotes the number of nodes, $K$ indicates the number of regimes, ``Split" specifies whether the algorithm automatically splits the regimes (``A") or if the split was done manually beforehand (``M") for the models. And ``Type" categorizes the type of returned graph as either window graph (``W") or summary graph (``S"). Inst. refers to instantaneous links and Lag to time-lagged edges.}
\resizebox{\columnwidth}{!}{
    \begin{tabular}{*{3}{l}*{6}{l} | *{6}{l}}
    \toprule
          & \multicolumn{2}{c}{} & \multicolumn{6}{c}{$d=10$} & \multicolumn{6}{c}{$d=40$} \\ \midrule
     Model & Split & Type & \multicolumn{2}{c}{$K=2$} & \multicolumn{2}{c}{$K=3$} & \multicolumn{2}{c}{$K=4$} & \multicolumn{2}{c}{$K=2$} & \multicolumn{2}{c}{$K=3$} & \multicolumn{2}{c}{$K=4$} \\ \midrule  
          &       &      & Inst. & Lag & Inst. & Lag & Inst. & Lag & Inst. & Lag & Inst. & Lag & Inst. & Lag \\ \midrule
     VARLINGAM     & M & W & $22.9_{\pm5.6}$ & $8.01_{\pm12}$ & $18.4_{\pm3.6}$ & $15.9_{\pm4.2}$ & $24.1_{\pm6.5}$ & $8.20_{\pm6.5}$ & $8.83_{\pm1.7}$ & $2.96_{\pm0.2}$ & $10.3_{\pm2.4}$ & $1.66_{\pm1.2}$ & $14.0_{\pm2.4}$ & $2.30_{\pm0.7}$ \\                                                               
     PCMCI+   & M & W & $98.5_{\pm2.1}$ & $86.1_{\pm11.1}$ & $99.0_{\pm1.4}$ & $88.6_{\pm9.4}$ & $96.2_{\pm1.8}$ & $85.9_{\pm6.9}$ & $59.2_{\pm2.9}$ & $79.8_{\pm5.1}$ & $61.5_{\pm1.4}$ & $81.9_{\pm5.1}$ & $60.2_{\pm4.0}$ & $81.4_{\pm3.6}$ \\
     RPCMCI  & A & W & \multicolumn{1}{c}{-} & \multicolumn{1}{c}{-} & \multicolumn{1}{c}{-} & \multicolumn{1}{c}{-} & \multicolumn{1}{c}{-} & \multicolumn{1}{c}{-} & - & $46.6_{\pm9.1}$ & \multicolumn{1}{c}{-} & 14.1$_{\pm2.3}$ & \multicolumn{1}{c}{-} & - \\ 
     CASTOR  & A & W & \textbf{100$_{\pm0.0}$} & \textbf{100$_{\pm0.0}$} & \textbf{100$_{\pm0.0}$} & \textbf{100$_{\pm0.0}$} & \textbf{97.3$_{\pm1.9}$} & \textbf{97.2$_{\pm2.0}$} & \textbf{97.0$_{\pm2.7}$} & \textbf{100$_{\pm0.0}$} & \textbf{88.1$_{\pm6.2}$} & \textbf{89.9$_{\pm5.1}$} & \textbf{98.3$_{\pm1.4}$} & \textbf{99.7$_{\pm0.4}$} \\
     \bottomrule
    \end{tabular}}
\label{table4_lin_5_20}
\vspace{-15pt}
\end{table}

We examine varying numbers of nodes, specifically $\{5, 20\}$,  and generated time series with different regime counts $\{2, 3, 4\}$. Our model's performance is benchmarked against multiple baselines, namely Rhino \citep{gong2022rhino}, PCMCI+ \citep{runge2020discovering}, RPCMCI \citep{saggioro2020reconstructing} and VARLINGAM \citep{hyvarinen2010estimation} and
the results are presented in Table \ref{table4_lin_5_20}. 

RPCMCI represents the sole baseline tailored to address a similar setting. RPCMCI necessitates prior knowledge of the number of regimes and the maximum number of transitions, and with this input, it only infers time-lagged relations. Even with this detailed information, RPCMCI struggles to achieve convergence, particularly in  settings with more than 3 different regimes. 
The absence of the inference of instantaneous relationships in RPCMCI poses a challenge for learning regime indices within our setting, as we assume the presence of instantaneous relationships. In contrast, CASTOR does not only surpass RPCMCI in performance but also converges consistently, correctly identifying the number of regimes and recovering both the regime indices and the underlying causal graphs of each regime. We can notice that CASTOR successively infers the regime indices and learns as well the instantaneous links as well as time lagged relations.

When we compare CASTOR and PCMCI+ with VARLINGAM and Rhino. CASTOR and PCMCI+ also demonstrate markedly superior performance (Specially for $d=5$ for PCMCI+, the model struggles when the number of nodes become greater). To provide context, we manually partition our generated data into $K$ regimes to facilitate the evaluation of VARLINGAM, PCMCI+. We then execute these aforementioned models on each segmented regime, infer the graphs, and compare these composite structures against their true counterparts. 
Even when executing VARLINGAM, PCMCI+ separately on each regime, CASTOR still outperforms these models without access to any prior information, such as the number of regimes or the indices of the regimes. PCMCI+, while excelling in capturing time-lagged relations, faces challenges with instantaneous links, particularly when dealing with a higher number of nodes. It can only identify the graph up to MECs without explicit functional relations.
\subsection{Further experiments and evaluation using SHD: nonlinear case}
\label{appnonlin}
\begin{table}[H]
\centering
\caption{We report the SHD on regimes for different Models and Settings for linear causal relationships: $d$ denotes the number of nodes, $K$ indicates the number of regimes, ``Split" specifies whether the algorithm automatically splits the regimes (``A") or if the split was done manually beforehand (``M") for the models. And ``Type" categorizes the type of returned graph as either window graph (``W") or summary graph (``S"). Inst. refers to instantaneous links and Lag to time-lagged edges.}
\resizebox{\columnwidth}{!}{
\begin{tabular}{*{3}{l}*{4}{l}*{4}{l}}
\toprule
          & \multicolumn{2}{c}{} & \multicolumn{4}{c}{$d=10$} & \multicolumn{4}{c}{$d=40$} \\ \midrule
     Model & Split & Type & \multicolumn{2}{c}{$K=2$} & \multicolumn{2}{c}{$K=4$} & \multicolumn{2}{c}{$K=2$} & \multicolumn{2}{c}{$K=4$} \\ \midrule  
          &       &      & Inst. & Lag & Inst. & Lag & Inst. & Lag & Inst. & Lag \\ \midrule
Rhino         & \multicolumn{1}{c}{M} & \multicolumn{1}{c}{W} & \multicolumn{1}{c}{21}  & \multicolumn{1}{c}{29} & \multicolumn{1}{c}{43}  & \multicolumn{1}{c}{53} & \multicolumn{1}{c}{71}  & \multicolumn{1}{c}{106} & \multicolumn{1}{c}{128}  & \multicolumn{1}{c}{207} \\
Rhino w/o hist& \multicolumn{1}{c}{M} & \multicolumn{1}{c}{W} & \multicolumn{1}{c}{\underline{15}} & \multicolumn{1}{c}{10} & \multicolumn{1}{c}{\underline{32}} & \multicolumn{1}{c}{\textbf{7}} & \multicolumn{1}{c}{57} & \multicolumn{1}{c}{10} & \multicolumn{1}{c}{123} & \multicolumn{1}{c}{\underline{30}} \\
PCMCI+       & \multicolumn{1}{c}{M} & \multicolumn{1}{c}{W} & \multicolumn{1}{c}{19}  & \multicolumn{1}{c}{\textbf{6}} & \multicolumn{1}{c}{36}  & \multicolumn{1}{c}{\underline{17}} & \multicolumn{1}{c}{\underline{49}} & \multicolumn{1}{c}{\textbf{6}} & \multicolumn{1}{c}{\underline{91}} & \multicolumn{1}{c}{\textbf{13}} \\
DYNOTEARS   & \multicolumn{1}{c}{M} & \multicolumn{1}{c}{W} & \multicolumn{1}{c}{16}  & \multicolumn{1}{c}{\textbf{6}} & \multicolumn{1}{c}{42}  & \multicolumn{1}{c}{16} & \multicolumn{1}{c}{57}  & \multicolumn{1}{c}{31} & \multicolumn{1}{c}{99}  & \multicolumn{1}{c}{47} \\
CASTOR      & \multicolumn{1}{c}{A} & \multicolumn{1}{c}{W} & \multicolumn{1}{c}{\textbf{3}}  & \multicolumn{1}{c}{\textbf{6}} & \multicolumn{1}{c}{\textbf{27}} & \multicolumn{1}{c}{22} & \multicolumn{1}{c}{\textbf{18}} & \multicolumn{1}{c}{\underline{7}} & \multicolumn{1}{c}{\textbf{58}} & \multicolumn{1}{c}{33} \\
\bottomrule
\end{tabular}}
\label{shdnonlinear}
\end{table}

\textcolor{black}{
In the SHD evaluation for non-linear settings, it's evident that CASTOR outperforms all the baselines in predicting instantaneous links. However, when it comes to time-lagged relationships, PCMCI+ excels and outperforms CASTOR. It's important to highlight that PCMCI+ operates in a more favorable scenario, as it is provided with ground truth regime partitions. Nevertheless, CASTOR stands out for its ability to learn the graphs, determine the number of regimes, and identify their respective indices.}
\begin{figure}[H]
\centering
\includegraphics[scale=0.28]{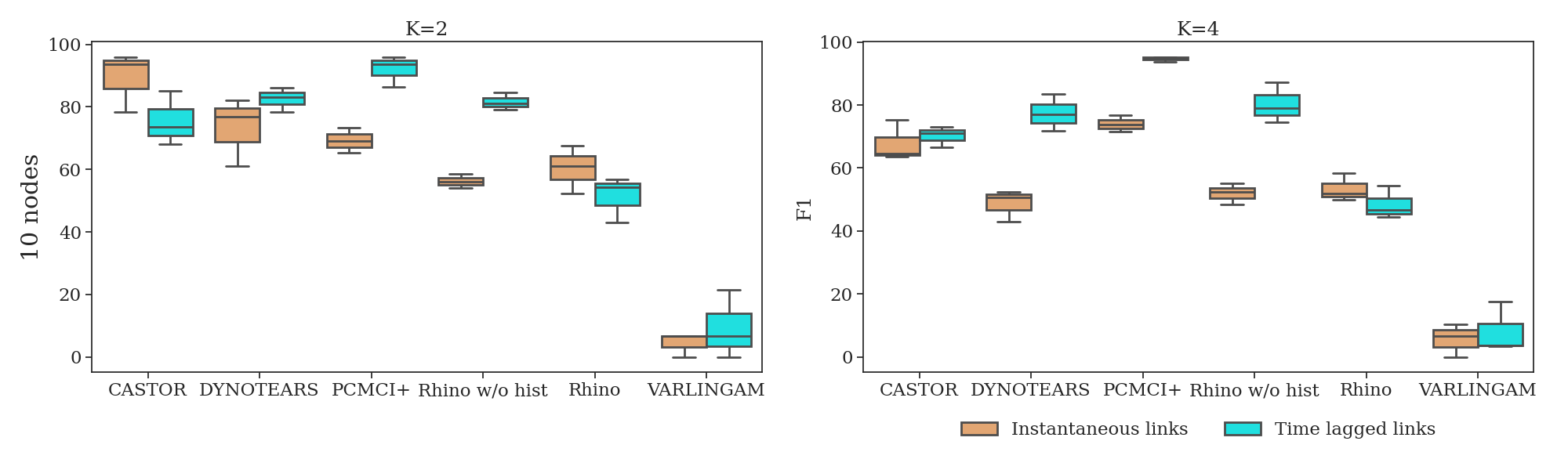}
\vspace{-1.em}
\caption{
F1 scores by Models and Settings: Orange indicates performance on instantaneous links, and sky-blue signifies performance on time-lagged relationships. Notably, \textbf{CASTOR is the only model capable of learning the number of regimes and regime indices; for other baselines, a manual split was performed beforehand.} Number of nodes is 10 nodes}
\label{fig_nonlin_10node_2_4}
\end{figure}
\begin{figure}[H]
\centering
\includegraphics[scale=0.28]{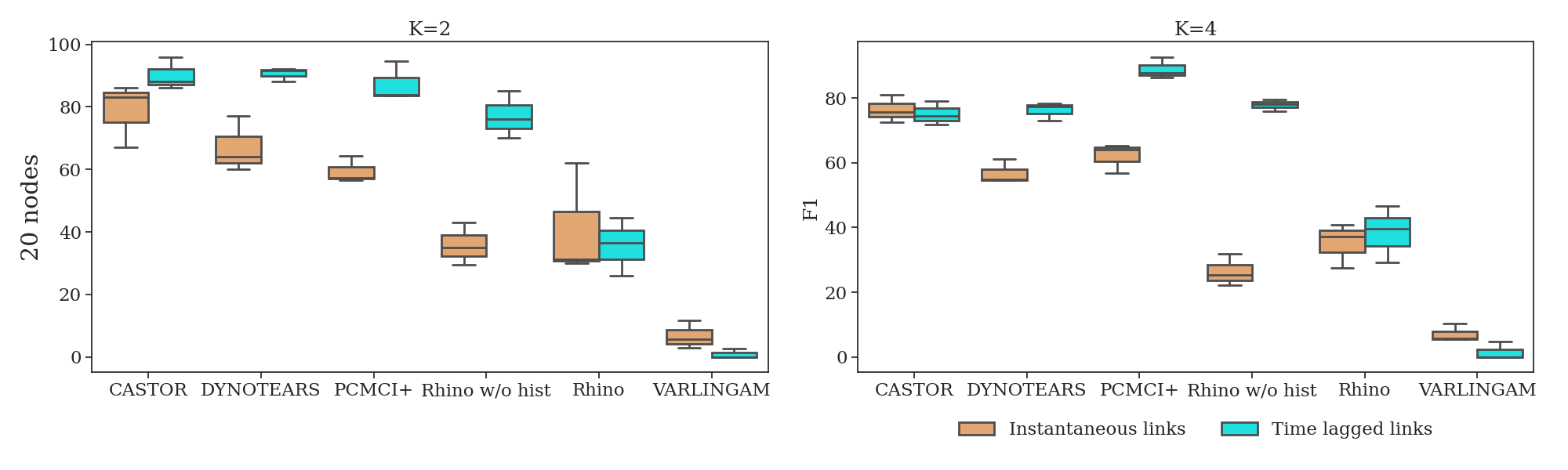}
\vspace{-1.em}
\caption{
F1 scores by Models and Settings: Orange indicates performance on instantaneous links, and sky-blue signifies performance on time-lagged relationships. Notably, \textbf{CASTOR is the only model capable of learning the number of regimes and regime indices; for other baselines, a manual split was performed beforehand.} Number of nodes is 20 nodes}
\label{fig_nonlin_10node_2_4}
\end{figure}
In this section, we present also some additional results using non-linear synthetic data with varying numbers of nodes, specifically $\{10, 20\}$, and diverse numbers of regimes $\{2,3,4\}$. In this case, we compare our model against various baseline models, namely DYNOTEARS \citep{pamfil2020dynotears}, Rhino \citep{gong2022rhino}, PCMCI+ \citep{runge2020discovering} and VARLINGAM \citep{hyvarinen2010estimation}. For setting with 10 nodes, we can see from Figure (\ref{fig_nonlin_10node_2_4}) that CASTOR, Rhino, PCMCI+ and DYNOTEARS exhibit superior performance to VARLINGAM. It is important to outline that Rhino, PCMCI+, DYNOTEARS and VARLINGAM are each applied to individual regimes separately; neither is designed to learn or infer the number or indices of regimes. PCMCI+ outperforms CASTOR, Rhino and DYNOTEARS in modeling time-lagged relations for non-linear scenarios. PCMCI+ demonstrates robustness in capturing non-linear relationships. Benefiting from a manual split performed beforehand, PCMCI+ exhibits comparable performance (for $K=2$) or outperforms (for $K=4$) CASTOR in capturing time-lagged relations. These results are understandable due to the fact that CASTOR has to learn more aspects, including the number of regimes and their indices. For instantaneous links, CASTOR consistently outperforms all models. In the case of $K=3$ regimes, CASTOR outperforms all models in instantaneous links and achieves comparable results in time-lagged relations as DYNOTEARS and PCMCI+. Notably, Rhino, a state-of-the-art causal discovery model, exhibits lower performance than DYNOTEARS and PCMCI+, \textbf{consistent with the findings reported by Rhino authors in their appendix when testing Rhino in settings with non historical dependent noise}. Moreover, Rhino requires a substantial amount of data for training (50 MTS with 200 time steps), while our setting only includes regimes with 600 time steps.

\begin{figure}[H]
\centering
\includegraphics[scale=0.28]{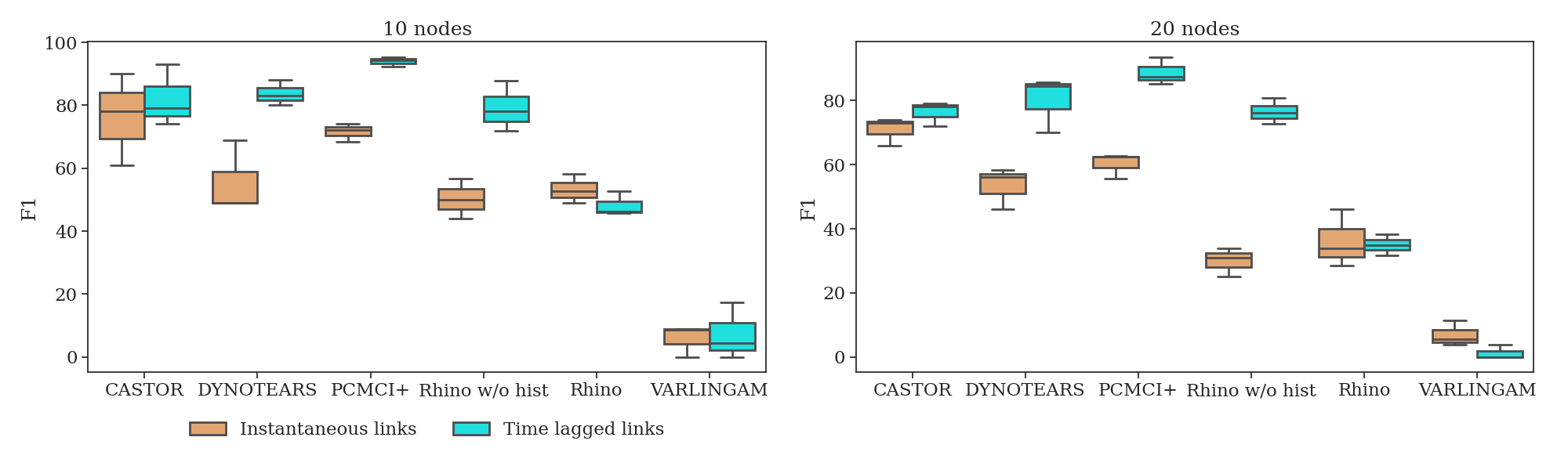}
\vspace{-1.em}
\caption{
F1 by Models for K=3 Setting: Gray indicates performance on instantaneous links, and pink signifies performance on time-lagged relationships. Notably, \textbf{CASTOR is the only model capable of learning the number of regimes and regime indices; for other baselines, a manual split was performed beforehand.}}
\label{fig_nonlin_3reg}
\end{figure}
\subsection{Further experiments: Comparison with CD-NOD}
\label{appcomp_cd_nod}
We conducted a comparative analysis with CD-NOD, a causal discovery model specifically designed for heterogeneous data and non-stationary time series \citep{huang2020causal}. In the context of MTS with multiple regimes, CD-NOD learns a summary causal graph encapsulating the entire MTS.

It's worth noting that in the work by \citet{huang2020causal}, CD-NOD demonstrates the capability to learn both the change points and the summary causal graph. We use the open-source code of CD-NOD implemented in Matlab by the authors\footnote{\url{https://github.com/Biwei-Huang/Causal-Discovery-from-Nonstationary-Heterogeneous-Data}}.

Our experimental setup involves linear causal relations and diverse configurations, including 2, 3, and 4 regimes, each with varying numbers of nodes (10, 20, 40). For independence test of CD-NOD, we chose Fisher’s Z conditional independence test for linear causal relationships and KCI independent test for non linear relations. We systematically compared the performance of CD-NOD against CASTOR, with the evaluation centered on the summary causal graphs as the basis for comparison.
\begin{definition}
(Summary causal graph, \citep{assaad2022survey})
Let $\boldsymbol{(x_t)}_{t \in \mathcal{T}}$ be a MTS and $\mathcal{G}=(V, E)$ the associated summary causal graph. The set of vertices in that graph consists of the set of components $x^1, \ldots, x^d$ at each time $t \in \mathbb{N}$. The edges $E$ of the graph are defined as follows: variables $x^p$ and $x^q$ are connected if and only if there exists some time $t$ and some time lag $\tau$ such that $x^p_{t-\tau}$ causes $x^q_{t}$ at time t 
with a time lag of $0\leq \tau $ for $p\neq q$ and with a time lag of $0 < \tau$ for $p = q$.
\end{definition}
\begin{table}[h]
\centering
\begin{tabular}{@{}lcllll@{}}
 & \multicolumn{1}{l}{} &        &              &               &               \\ \cmidrule(l){3-6} 
 & \multicolumn{1}{l}{$d$} & Method & $K=2$          & $K=3  $         & $K=4$           \\ \cmidrule(l){2-6} 
 & \multirow{2}{*}{10}  & CD-NOD  & 20.2         & 11.4          & 38.8          \\
 &                      & CASTOR & \textbf{100} & \textbf{100}  & \textbf{97.9} \\ \cmidrule(l){2-6} 
 & \multirow{2}{*}{20}  & CD-NOD  & 25.2         & 23.7          & 12.7          \\
 &                      & CASTOR & \textbf{100} & \textbf{97.2} & \textbf{93.4} \\ \cmidrule(l){2-6} 
 & \multirow{2}{*}{40}  & CD-NOD  & 0            & 11.3          & 5.57          \\
 &                      & CASTOR & \textbf{100} & \textbf{99.8} & \textbf{99.2} \\ \cmidrule(l){2-6} 
\end{tabular}
\vspace{0.4em}
\caption{F1 Scores by Models and Settings: $d$ indicates number of nodes and $K$ refers to the number of regimes. The comparison is made for linear relations. \textbf{CASTOR detects the regimes automatically}.}
\label{tab6}
\end{table}
From Table \ref{tab6}, we can notice that CD-NOD does not manage to outperform CASTOR in various settings (with an F1 score that does not exceed 26\%). Additionally, a clear trend emerges where CD-NOD's performance declines when the number of nodes is 40. On the contrary, CASTOR exhibits consistent performance across different settings, achieving a F1 score of over 93\% in all scenarios.
\begin{table}[H]
\centering
\textcolor{black}{\begin{tabular}{@{}c|cc|@{}}
\cmidrule(l){2-3}
                             & \multicolumn{2}{c|}{$d=10$}                        \\ \cmidrule(l){2-3} 
                             & \multicolumn{1}{c|}{$K=2$}         & $K=3$         \\ \midrule
\multicolumn{1}{|c|}{CD-NOD} & \multicolumn{1}{c|}{28.5}          & 25.3          \\ \midrule
\multicolumn{1}{|c|}{CASTOR} & \multicolumn{1}{c|}{\textbf{86.1}} & \textbf{85.2} \\ \bottomrule
\end{tabular}}
\vspace{0.4em}
\caption{\textcolor{black}{F1 Scores by Models and Settings: $d$ indicates number of nodes and $K$ refers to the number of regimes. The comparison is made for non linear relations. \textbf{CASTOR detects the regimes automatically}.}}
\label{tab_cdnod_nonlinear}
\end{table}
\textcolor{black}{From Table \ref{tab_cdnod_nonlinear}, CD-NOD with KCI independent does not manage to outperform CASTOR in non linear causal relationships settings (with an F1 score that does not exceed 28\%). This is understandable, because firstly, CD-NOD learns one summary graph. The model assumes smooth changes in graphs between regimes i.e. it expects only a few variables of the graph to be affected by the regime switch (Assumption may not hold true in scenarios such as epileptic seizures or climate science). However, CASTOR did not have this assumption and also learn one causal graph per regime.} 
\subsection{Further experiments: Regime detection experiment}
\label{app_reg_detect}
We compare CASTOR to CD-NOD \citep{huang2020causal} and KCP \citep{arlot2019kernel} in the task of regime detection. KCP is a multiple change-point detection method designed to handle univariate, multivariate, or complex data. Being non-parametric, KCP does not necessitate knowing the true number of change points in advance. It detects abrupt changes in the complete distribution of the data by employing a characteristic kernel. 

As we described in the previous section, CD-NOD is a causal discovery model specifically designed for heterogeneous data and non-stationary time series. In phase III of CD-NOD, a method is proposed to learn change points (that occur when causal relationships changes) for MTS with multiple regimes. CD-NOD estimates a non-stationary driving force for each component (node, considering its parents), where this driving force is a function of the time index. If the driving force changes with the time index, it indicates a change in the regime for that component; otherwise, it signifies that the component is within the same regime.

As previously explained, CD-NOD learns a driving force for components that change behavior. To detect the regimes, one approach is to learn the change points for all components exhibiting changing behavior and then form the union of these change points, yielding the regime partition.

In contrast to CD-NOD, CASTOR directly learns the regime indices. Consequently, for every sample, CASTOR assigns it to a specific regime, directly yielding the regime partition. Additionally, CASTOR conducts regime detection at the graph level rather than the node level which makes it faster than CD-NOD in the task of regime detection.

We opted to perform regime detection in two settings: one with 10 nodes and four different regimes, and another with 20 nodes and four distinct regimes, the causal relationships are non linear in this scenario. For a fair comparison, we chose four regimes without re-occurrence, as KCP and CD-NOD only detect change points and cannot identify the re-occurrence of a specific regime. CASTOR excels in this regard since it detects regime indices. For instance, if regime $u$ occurs from $t=1$ to $t=400$ and then reoccurs from $t=1000$ to $t=1300$, CASTOR encompasses all these indices in the previously defined regime partition $\mathcal{E}_u$. It utilizes all these indices collectively to learn a more accurate graph.

Regarding the models employed, we use the open-source code of CD-NOD implemented in Matlab by the authors\footnote{\url{https://github.com/Biwei-Huang/Causal-Discovery-from-Nonstationary-Heterogeneous-Data}}. For KCP, we employ the \texttt{Rupture} package\footnote{\url{https://centre-borelli.github.io/ruptures-docs/}}.
\begin{table}[H]
\centering
\begin{tabular}{@{}c|c|c|@{}}
\cmidrule(l){2-3}
                             & \begin{tabular}[c]{@{}c@{}}Regime accuracy\\ $K=4$ and $d=10$\end{tabular} & \begin{tabular}[c]{@{}c@{}}Regime accuracy\\ $K=4$ and $d=20$\end{tabular} \\ \midrule
\multicolumn{1}{|c|}{KCP}    & 33.4                                                                       & 66.8                                                                       \\ \midrule
\multicolumn{1}{|c|}{CD-NOD} & 70.1                                                                       & 88.2                                                                       \\ \midrule
\multicolumn{1}{|c|}{CASTOR} & \textbf{95.9}                                                              & \textbf{98.7}                                                              \\ \bottomrule
\end{tabular}
\vspace{0.6em}
\caption{Regime detection accuracy by models: CASTOR and CD-NOD outperform the state-of-the-art change-point detection method KCP. CASTOR achieves maximum accuracy, surpassing 95\% for both settings.}
\label{tab_reg_detect}
\end{table}
From the table, it is evident that causal models (CASTOR and CD-NOD) outperform the change-point detection method KCP. This outcome can be attributed to the limitation of KCP in detecting changing points within causal mechanisms that are represented by conditional distributions. CASTOR outperforms CD-NOD in detecting regime indices. This result can be explained by the fact that CASTOR learns regime indices based on graph-level change points, while CD-NOD detects change points at the node level. The node-level approach in CD-NOD may not effectively detect simultaneous changes in behavior for components that actually change behavior simultaneously.\\
From this analysis, we can conclude that in scenarios involving MTS with multiple regimes and unknown regime indices, CASTOR offers a robust solution. Additionally, employing other methods to split the regimes and learn the causal graph through traditional causal discovery methods may not be an optimal solution:
\begin{itemize}
    \item We demonstrate that regime indices are not well recoverable by CD-NOD and other state-of-the-art change point detection method KCP. Therefore, employing CD-NOD or KCP to learn the regimes and subsequently using methods like DYNOTEARS, PCMCI, or Rhino to learn the graph may not constitute an optimal solution.
    \item In cases of regime recurrence, the aforementioned methods are unable to accurately detect the exact number of regimes. Therefore, if a user employs CD-NOD and subsequently uses the regime partitions revealed by CD-NOD as an input to a causal discovery method (such as PCMCI+, DYNOTEARS, Rhino, etc.), the running time will be significantly high.\\
    Example: To elaborate further, let's consider the epilepsy setting. Imagine we have a recording from an epileptic patient where the sequence involves a non-seizure phase, followed by a seizure phase, and then a reappearance of the non-seizure phase. Employing CD-NOD, particularly with a KCI independence test, to detect regimes in such a scenario can be computationally expensive (\textcolor{red}{For the table above scenario with 20 nodes CD-NOD takes more than 24h compared to CASTOR that learns the regimes and the graph in less than 1h}). Subsequently, applying algorithms like Rhino, DYNOTEARS, or Lingam to learn temporal causal graphs would also be resource-intensive. This is primarily because the user would need to run the chosen algorithm at least three times (twice for non-seizure and once for seizure) since CD-NOD does not clarify that a the non-seizure regime is reappearing.
   
\end{itemize}
\subsection{Models running time}
\label{runtime}
\begin{figure}[H]
\centering
\includegraphics[scale=0.23]{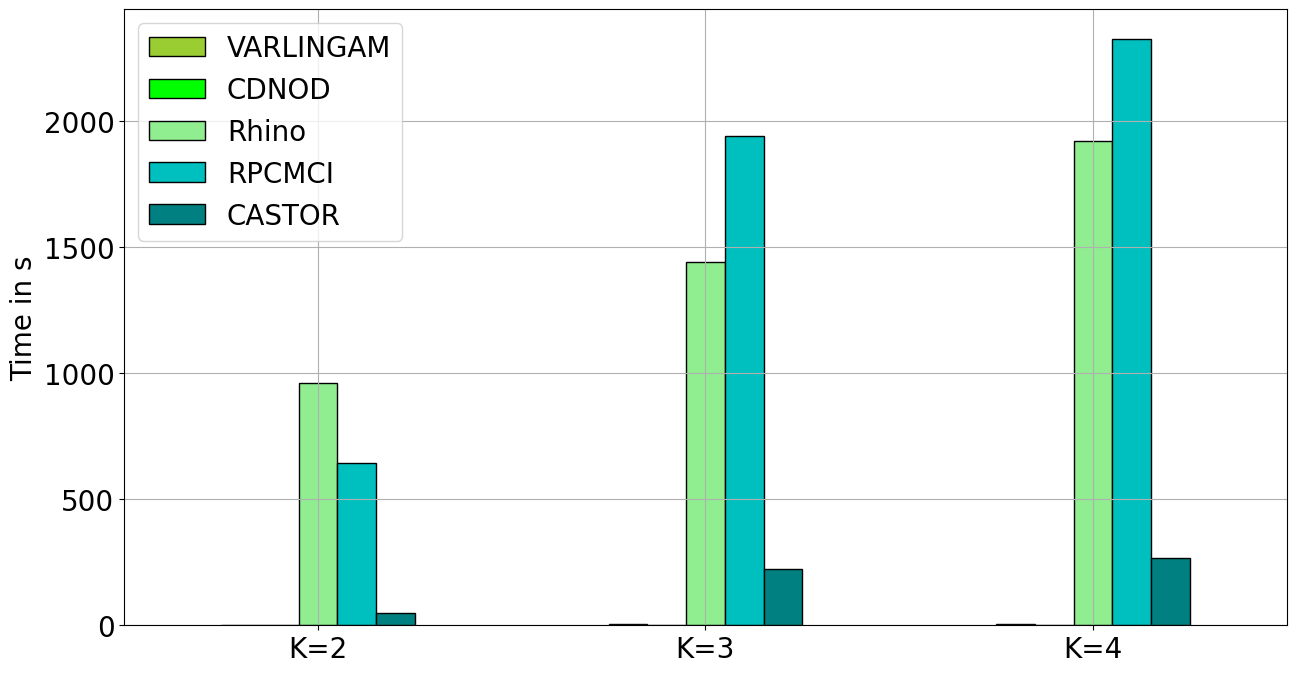}
\caption{
Running time per model, the y axis represents the running time in s and the x axis the number of regime}
\vspace{-0.5em}
\label{timecomp}
\end{figure}
\vspace{-1em}
We compute the running time of every model in different settings, that includes 10 nodes and 2,3 or 4 different regimes, Figure \ref{timecomp} summarizes the results.

VARLINGAM and CD-NOD (employing Fisher’s Z conditional independence test for faster runs; note that KCI CD-NOD takes over 2000 seconds for 2 regimes, with F1 scores in a similar range) exhibit remarkable speed compared to other methods. However, in terms of scores, both models encounter challenges in effectively learning causal graphs. Notably, CASTOR runs faster than Rhino, even though CASTOR learns both temporal causal graphs and regime indices.

A fair comparison arises when comparing RPCMCI and CASTOR, as both models learn regime indices and temporal causal graphs. It's essential to mention that RPCMCI necessitates specifying the number of regimes and the maximum number of transitions, producing only time-lagged relations. From Figure \ref{timecomp}, it is apparent that CASTOR converges more rapidly than RPCMCI. This difference in convergence time becomes more pronounced as the settings become more complex.
\subsection{Biosphere–Atmosphere data}
\label{app_biosphere}
\begin{figure}[H]
\centering
\includegraphics[scale=0.6]{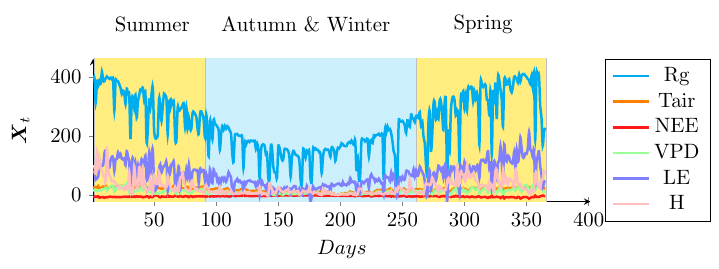}
\includegraphics[scale=0.5]{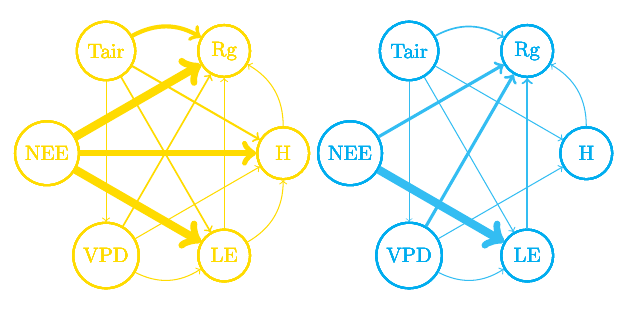}

\caption{Applying CASTOR to Biosphere atmosphere data, which lacks a ground-truth regime partition or causal graph, aims to emphasize the practical utility of CASTOR. CASTOR identifies two regimes with distinct graphs. Here, only instantaneous links are shown, with the blue graph corresponding to the blue regime and the yellow one to the yellow regime.}
\end{figure}
We apply CASTOR to Biosphere atmosphere data\footnote{FLUXNET Dataset San Luis site, Argentina published by \citep{sanluis}}. The objective is to demonstrate the practical utility of CASTOR in a real-world scenario. \\

\citet{krich2021functional} employ this type of data to learn causal relations between six variables (global radiation ($R_g$), air temperature ($T_{air}$), net ecosystem exchange (NEE), vapor pressure deficit (VPD), sensible heat (H), latent heat flux (LE)) under climate change conditions. They utilize the causal discovery model PCMCI+ for windows of three months (with one overlapping month) to learn causal relations. The objective of our experiments is to automate the learning of causal graphs and regime partitions. In other words, instead of assuming that the causal graph changes every three months and applying a causal discovery method at that interval, we provide MTS of one year's length to CASTOR  and we let the model (CASTOR) decide when the causal graph changes and which months are more similar. Although we initially set a window length of three months, CASTOR stabilizes after a few iterations in a state with two regimes. \\ It is essential to note that assuming causal sufficiency in this setting is a difficult assumption, as the biosphere-atmosphere relationship is considerably more complex in reality. Consequently, validating the causal sufficiency assumption is challenging, which explains the appearance of some suspicious relationships. Nevertheless, we will interpret some of the relations and explain why certain causal relations appear and why others are inverted.\\

In our application, we use CASTOR on this MTS data of six variables, aiming to automatically learn the regime partition and the causal graphs. Initiating with non-overlapping windows of three months (4 initialed different regimes) and a minimum regime duration of two months, CASTOR splits the data into two regimes: the first corresponds to the cold regime grouping Autumn and Winter (From April to September), while the second encompasses Summer and Spring. The accuracy of this partition is 85.4\%, with some days in the second regime occasionally misclassified as the first one.\\

Regarding the causal links, an initial observation is that the strength (represented by thickness) of certain causal relations is greater in the hot regime (yellow) compared to the cold regime (blue). Furthermore, all variables in both regimes are identified as parents of global radiation, which is intuitively incorrect. However, this observation can be explained by our assumption of causal sufficiency, where we assume observation of all causal variables, which is, in reality, not accurate. To explain this result, our model interprets global radiation as the net radiation. Mathematically, we have:
$$
R_n=R_g-S W_{\uparrow}+L W_{\downarrow}-L W_{\uparrow},
$$
where $R_n$ is the net radiation, $R_g$ is the global radiation which is also global shortwave radiation, part of the the incoming radiation is reflected at the surface (shortwave upward radiation $S W_{\uparrow} = \alpha R_g$, $L W_{\uparrow}$ is the long-wave upward radiation which is the amount of long-wave radiation emitted at the surface and $L W_{\uparrow}$ is the long-wave downward radiation. Also we have that the net radiation $R_n$:
$$
R_n=H+LE+G,
$$
where $H$ is the sensible heat, $LE$ is the latent heat flux and $G$ is the ground heat flux.\\
In our setting, where we assume causal sufficiency (which is not accurate), CASTOR learns incorrect link directions. Every variable appears to act as a parent of $R_g$ (global radiation), a behavior explained by the two equations above, where CASTOR interprets $R_g$ as the net radiation $R_n$. Additionally, we observe that temperature influences vapor pressure deficit (VPD), which is understandable. For instance, in hot weather, evaporation can be sustained for a longer period, leading to higher VPD (VPD is the difference between the maximum moisture that could be hold in the air could hold and the actual moisture in the air). Furthermore, we can notice that the temperature $T_{air}$ is a common cause of sensible heat $H$ and latent heat flux $LE$ which is understandable as we know that these two variable can mathematically be described as function temperature. 
The relationship between $NEE$ and $LE$ is complex and influenced by various factors including plant physiology, climate, soil moisture, and atmospheric conditions. There's no direct relation linking NEE and LE, both variables are influenced by the conductance of stomata which regulates both $CO_2$ uptake for photosynthesis (affecting $NEE$) and water vapor release (affecting $LE$). Thus, having this common confounder variable causes the appearance of suspicious link ($NEE \rightarrow LE$). As we mentioned, $LE$ is affected by water vapor release, which mathematically could be written as follows:  $LE \propto g_s \cdot VPD$ where (VDP) Vapor pressure deficit and $g_s$ is the conductance of stomata. Hence, this equation explains the link $VPD \rightarrow LE$.

The key difference between the two regimes lies in the strength of the causal relationships. This is because the variables take on different values in each regime. This finding demonstrates CASTOR's ability to distinguish between graphs with identical structures but varying edge weights.
\\
\newpage
\section{REGIME AND CAUSAL GRAPHS IDENTIFIABILITY}
\label{appB}

In this section, we concentrate on establishing the identifiability of regimes and causal graphs within the CASTOR framework. Before diving into the details, let us set and clarify the required assumptions.\\
\begin{definition}
(Causal Stationarity \cite{runge2018causal}). The time series (that has one regime) process $(\boldsymbol{x}_t)_{t \in \mathcal{T}}$ with a graph $\mathcal{G}$ is called causally stationary over a time index set $\mathcal{T}$ if and only if for all links $x_{t-\tau}^i \rightarrow x_t^j$ in the graph
$$
x_{t-\tau}^i \not\!\perp\!\!\!\perp  x_t^j \mid \boldsymbol{x}_{<t} \backslash\left\{x_{t-\tau}^i\right\} \text { holds for all } t \in \mathcal{T}
$$
\label{causalstat}
\end{definition}
This elucidates the inherent characteristics of the time-series data generation mechanism, thereby validating the choice of the auto-regressive model. In our setting, we generalize Causal Stationarity as follows:
\begin{assumption}
(Causal Stationarity for time series with multiple regimes). The time series  process $(\boldsymbol{x}_t)_{t \in \mathcal{T}}$ comprise
multiple regimes $K$, where $K$ is the number of regime, we note $\mathcal{E}_u = \{t| \gamma_{t,u} = 1\}$ the set of time indices where the regime $u$ is active, and $\mathcal{T} = \cup_u \mathcal{E}_u$. $(\boldsymbol{x}_t)_{t \in \mathcal{T}}$  with a graph $\{\mathcal{G}^u\}_{u \in \{1,...,K\}}$ is called causally stationary over a time index set $\mathcal{T}$ if and only if for all $u \in \{1,...,K\}$, $(\boldsymbol{x}_t)_{t \in \mathcal{E}_u}$ is causal stationary with graph $\mathcal{G}^u$ for time index set $\mathcal{E}_u$.
\label{ass1}
\end{assumption}
\begin{definition} (Causal Markov Property, \cite{peters2017elements}). Given a DAG $\mathcal{G}$ and a joint distribution $p$, this distribution is said to satisfy causal Markov property w.r.t. the DAG $\mathcal{G}$ if each variable is independent of its non-descendants given its parents. 
\label{markovprop}
\end{definition}

This is a common assumptions for the distribution induced by an SEM. With this assumption, one
can deduce conditional independence between variables from the graph.
\begin{assumption}
    (Causal Markov Property for  multiple regimes). Given a set of DAGs $(\mathcal{G}^u)_{u \in \{1,...,K\}}$ and a set of joint distribution $(p(\cdot|\mathcal{G}^u))_{u \in \{1,...,K\}}$, we say that this set of distributions satisfies causal Markov property w.r.t. the set of DAGs $(\mathcal{G}^u)_{u \in \{1,...,K\}}$  if for every $u$: $p(\cdot|\mathcal{G}^u)$ satisfy causal Markov property w.r.t the DAG $\mathcal{G}^u$.
    \label{ass2}
\end{assumption}
\begin{definition}
   (Causal Minimality, \cite{gong2022rhino}). Consider a distribution $p$ and a DAG $\mathcal{G}$, we say this distribution satisfies causal minimality w.r.t. $\mathcal{G}$ if it is Markovian w.r.t. $\mathcal{G}$ but not to any proper subgraph of $\mathcal{G}$.
   \label{causalmini}
\end{definition}

\begin{assumption}
(Causal Minimality for multiple regimes). Given a set of DAGs $(\mathcal{G}^u)_{u \in \{1,...,K\}}$ and a set of joint distribution $(p(\cdot|\mathcal{G}^u))_{u \in \{1,...,K\}}$, we say that this set of distributions satisfies causal minimality w.r.t. the set of DAGs $(\mathcal{G}^u)_{u \in \{1,...,K\}}$  if for every $u$: $p(\cdot|\mathcal{G}^u)$ satisfy causal minimality w.r.t the DAG $\mathcal{G}^u$.
\label{ass3}
\end{assumption}
\begin{assumption}
(Causal Sufficiency). A set of observed variables $V$ is causally sufficient for a process $\boldsymbol{x}_t$ if and only if in the process every common cause of any two or more variables in $\boldsymbol{V}$ is in $\boldsymbol{V}$ or has the same value for all units in the population.  
\label{ass4}
\end{assumption}

This assumption implies there are no latent confounders present in the time-series data.
\begin{table}[h]
\begin{adjustbox}{width=\columnwidth,center}

\begin{tabular}{c|cccccc|}
\cline{2-7}
                                & \begin{tabular}[c]{@{}c@{}}Causal \\
                                  graph\end{tabular} & \begin{tabular}[c]{@{}c@{}}Causal \\
                                  Markov\end{tabular} & \begin{tabular}[c]{@{}c@{}}Causal \\
                                  sufficiency\end{tabular} & \begin{tabular}[c]{@{}c@{}}Faithfulness \\
                                  / Minimality\end{tabular} & \begin{tabular}[c]{@{}c@{}}Linear\\
                                  model\end{tabular} & \begin{tabular}[c]{@{}l@{}}Stationarity\\
                                  per regime\end{tabular} \\ \hline
\multicolumn{1}{|c|}{DYNOTEARS} & W                                                       & \checkmark                                               & \checkmark                                                    &                                                                      & \checkmark                                             &                                                            $\boldsymbol{\times}$       \\
\multicolumn{1}{|c|}{PCMCI+}    & W                                                       & \checkmark                                               & \checkmark                                                    & F                                                                    & $\boldsymbol{\times}$                                  &                                                              $\boldsymbol{\times}$     \\
\multicolumn{1}{|c|}{RPCMCI}    & W                                                       & \checkmark                                               & \checkmark                                                    & F                                                                    & $\boldsymbol{\times}$                                  & \checkmark                                                        \\
\multicolumn{1}{|c|}{Rhino}     & W                                                       & \checkmark                                               & \checkmark                                                    & M                                                                    & $\boldsymbol{\times}$                                  &  $\boldsymbol{\times}$                                                                 \\
\multicolumn{1}{|c|}{VARLINGAM} & W                                                       & \checkmark                                               & \checkmark                                                    &                                                                      & \checkmark                                             &      $\boldsymbol{\times}$                                                             \\
\multicolumn{1}{|c|}{CD-NOD}    & S                                                       & \checkmark                                               & \checkmark                                                    & F                                                                    & $\boldsymbol{\times}$                                  & \checkmark                                                        \\
\multicolumn{1}{|c|}{CASTOR}    & W                                                       & \checkmark                                               & \checkmark                                                    & M                                                                    & \checkmark                                 & \checkmark                                                        \\ \hline
\end{tabular}
\end{adjustbox}
\vspace{0.5em}
\label{tab_assumption}
\vspace{0.5em}
\caption{Summary of the main assumptions of algorithms considered in the paper. For causal graphs, S means that the algorithm provides a summary causal graph and W means that the algorithm provides
a window causal graph; F corresponds to faithfulness and M to minimality. An empty cell mean that the information given in the corresponding column was not discussed by the authors of the corresponding algorithm.}
\label{tab_assumption}
\end{table}
\textcolor{black}{The table \ref{tab_assumption} illustrates that most assumptions (causal sufficiency, causal Markov, faithfulness/minimality) are commonly shared among various state-of-the-art models in causal discovery.}\\

\textcolor{black}{However, CASTOR, RPCMCI, and CD-NOD relax the assumption of stationarity and instead assume that the MTS (Multivariate Time Series) are composed of different regimes. While CD-NOD predicts only a summary causal graph, CASTOR and RPCMCI predict a window causal graph, which can subsequently be used to reconstruct a summary graph.}\\

\textcolor{black}{One notable difference in assumptions between CASTOR and RPCMCI is that RPCMCI assumes only time-lagged relations, whereas CASTOR incorporates the presence of instantaneous links.}
\subsection{Proof of theorem 1}
Assuming the aforementioned assumptions we want to prove the theorem \ref{theo:1}.\\
\label{theoproof}
We consider $\mathcal{G} = (\mathcal{G}^u)_{u \in \{1,..,N_w\}}$, $ \mathcal{E} = \cup_{u=1}^{N_w} \mathcal{E}_u$ where $N_w$ is the number of window, $\mathcal{G}^* = (\mathcal{G}^{*,u})_{u \in \{1,..,K\}}$, $K$ is the exact number of true regimes and $ \mathcal{E}^* = \cup_{u=1}^{K} \mathcal{E}^*_u$. We denote $\mathcal{E}_c \mathcal{E}_\ell^*$ the set of time indices that is shared between regime $c$ of our model estimation and the true regime $\ell$ and  $q_{\ell}^*:=\frac{\left|\mathcal{E}_{\ell}^*\right|}{T}, q_c:=\frac{\left|\mathcal{E}_c\right|}{T}$, $q_{c \ell}:=\frac{\left|\mathcal{E}_c \mathcal{E}_{\ell}^*\right|}{T}$. Our objective is to prove that for any estimation $(\hat{\mathcal{G}}, \hat{\mathcal{E}})$ : if $ \exists u \in \{1,...,K\} \text{  s.t.   } \hat{\mathcal{G}^u}  \text{ disagree with}$ $\mathcal{G}^{*,u}$ on instantaneous or/and time lagged link,  or any
regime $\hat{\mathcal{E}}_u \in \hat{\mathcal{E}}$ is close to none of the true regimes in the sense of Kullback–Leibler divergence: $
\mathcal{S}\left(\mathcal{G}^*, \mathcal{E}^*\right)>\mathcal{S}(\hat{\mathcal{G}},  \hat{\mathcal{E}}).
$ We have by Eq (\ref{Mstep}):
\begin{equation*} 
    \mathcal{S}(\mathcal{G}, \mathcal{E}):=  \sup _{\theta,\mathcal{G}} \frac{1}{T} \sum_{u=1}^{N_{\mathrm{u}}} \sum_{t \in \mathcal{E}_u} \log f^{u}\left(\boldsymbol{x}_t\right)  -\lambda|\mathcal{G}^u|,
\end{equation*}
where $\lambda$ is the sparsity penalty coefficient and $f^{u}\left(\boldsymbol{x}_t\right):=\prod_{j=1}^d f^u_i\left(\mathbf{P a}_{\mathcal{G}^u}^i(<t), \mathbf{P a}_{\mathcal{G}^u}^i(t)\right)$ with $f^u_i\left(\mathbf{P a}_{\mathcal{G}^u}^i(<t), \mathbf{P a}_{\mathcal{G}^u}^i(t)\right)$ the function used to describe the distribution family in Eq (\ref{6f}).\\
We will structure the proof as follows:
\begin{itemize}
    \item Prove that if the score is optimized, then all the estimated regimes will be pure (have only elements of the same true regime).
    \item Prove that, when the regimes are pure and $N_w = K$, we have $
\mathcal{S}\left(\mathcal{G}^*, \mathcal{E}^*\right)>\mathcal{S}(\hat{\mathcal{G}},  \hat{\mathcal{E}})
$ for any estimation $\hat{\mathcal{G}}$ where $ \exists u \in \{1,...,K\} \text{  s.t  } \hat{\mathcal{G}^u}  \text{ disagrees with}$ $\mathcal{G}^{*,u}$ on instantaneous or/and time lagged link.

\end{itemize}
\subsubsection{Optimizing the score will lead to pure regimes}
We denote by $p^{(u)}$ the distribution $p(.|\mathcal{G}^{u})$, ignoring penalty terms, we have:
\begin{equation}
\begin{aligned}
     -\mathcal{S}(\mathcal{G}, \mathcal{E})&=-\sup _\theta \sum_{c=1}^{N_w} \sum_{\ell=1}^K q_{c \ell} \frac{1}{\left|\mathcal{E}_e \mathcal{E}_{\ell}^*\right|} \sum_{t \in \mathcal{E}_e \mathcal{E}_{\ell}^*}\left[\log f^{c}(\boldsymbol{x}_t)\right] \\
& \rightarrow-\sup _\phi \sum_{c=1}^{N_w} \sum_{\ell=1}^K q_{c \ell} \underset{\boldsymbol{x}_t \sim p^{(\ell)}}{\mathbb{E}}\left[\log f^{c}\right] \\
& =-\sup _\theta \sum_{c=1}^{N_{w}} \sum_{\ell=1}^K q_{c \ell} \underset{\boldsymbol{x}_t  \sim p^{(\ell)}}{\mathbb{E}}\left[\sum_{i=1}^d \log f^c_i\left(\mathbf{P a}_{\mathcal{G}^c}^i(<t), \mathbf{P a}_{\mathcal{G}^c}^i(t)\right)\right] \\
& =-\sup _\theta \sum_{c=1}^{N_{w}} \sum_{\ell=1}^K \sum_{i=1}^d q_{c \ell} \\&\underset{\boldsymbol{x}_t \sim p^{(\ell)}}{\mathbb{E}}\left[-\log \frac{p\left(x_t^{i} \mid \left(\mathbf{P a}_{\mathcal{G}^{*,\ell}}^i(<t), \mathbf{P a}_{\mathcal{G}^{*,\ell}}^i(t)\right)\right)}{f^c_i\left(\mathbf{P a}_{\mathcal{G}^c}^i(<t), \mathbf{P a}_{\mathcal{G}^c}^i(t)\right)}+\log p\left(x_t^{i} \mid \left(\mathbf{P a}_{\mathcal{G}^{*,\ell}}^i(<t), \mathbf{P a}_{\mathcal{G}^{*,\ell}}^i(t)\right)\right)\right] \\
& =-\sup _\theta \sum_{c=1}^{N_{w}} \sum_{\ell=1}^K \sum_{i=1}^d q_{c \ell}\\& \underset{\boldsymbol{x}_t \sim p^{(\ell)}}{\mathbb{E}}\Big[- \mathrm{D}_{\mathrm{KL}}\left(p\left(x_t^{i} \mid \left(\mathbf{P a}_{\mathcal{G}^{*,\ell}}^i(<t), \mathbf{P a}_{\mathcal{G}^{*,\ell}}^i(t)\right)\right) \| f^c_i\left(\mathbf{P a}_{\mathcal{G}^c}^i(<t), \mathbf{P a}_{\mathcal{G}^c}^i(t)\right)\right)\\&-\mathrm{H}\left(p\left(x_t^{i} \mid \left(\mathbf{P a}_{\mathcal{G}^{*,\ell}}^i(<t), \mathbf{P a}_{\mathcal{G}^{*,\ell}}^i(t)\right)\right)\right)\Big] \\
&=-\sup _\theta \sum_{c=1}^{N_{w}} \sum_{\ell=1}^K \sum_{i=1}^d q_{c \ell} \underset{\boldsymbol{x}_t \sim p^{(\ell)}}{\mathbb{E}}\left[- \mathrm{D}_{\mathrm{KL}}\left(p\left(x_t^{i} \mid \left(\mathbf{P a}_{\mathcal{G}^{*,\ell}}^i(<t), \mathbf{P a}_{\mathcal{G}^{*,\ell}}^i(t)\right)\right) \| f^c_i\left(\mathbf{P a}_{\mathcal{G}^c}^i(<t), \mathbf{P a}_{\mathcal{G}^c}^i(t)\right)\right)\right]\\&-\sup _\theta \sum_{c=1}^{N_{\mathrm{c}}} \sum_{\ell=1}^K \sum_{i=1}^d q_{c \ell} \underset{\boldsymbol{x}_t \sim p^{(\ell)}}{\mathbb{E}}\left[-\mathrm{H}\left(p\left(x_t^{i} \mid \left(\mathbf{P a}_{\mathcal{G}^{*,\ell}}^i(<t), \mathbf{P a}_{\mathcal{G}^{*,\ell}}^i(t)\right)\right)\right)\right] \\
& =\inf _\theta \sum_{c=1}^{N_w} \sum_{\ell=1}^K \sum_{j=1}^d q_{c \ell} \underset{\boldsymbol{x}_t \sim p^{(\ell)}}{\mathbb{E}}\left[\mathrm{D}_{\mathrm{KL}}\left(p\left(x_t^{i} \mid \left(\mathbf{P a}_{\mathcal{G}^{*,\ell}}^i(<t), \mathbf{P a}_{\mathcal{G}^{*,\ell}}^i(t)\right)\right) \| f^c_i\left(\mathbf{P a}_{\mathcal{G}^c}^i(<t), \mathbf{P a}_{\mathcal{G}^c}^i(t)\right)\right)\right]\\&+\sum_{c=1}^{N_w} \sum_{\ell=1}^K \sum_{j=1}^d q_{c \ell} \underset{\boldsymbol{x}_t \sim p^{(\ell)}}{\mathbb{E}}\left[\mathrm{H}\left(p\left(x_t^{i} \mid \left(\mathbf{P a}_{\mathcal{G}^{*,\ell}}^i(<t), \mathbf{P a}_{\mathcal{G}^{*,\ell}}^i(t)\right)\right)\right)\right] \\
& =\inf _\theta \sum_{c=1}^{N_{w}} \sum_{\ell=1}^K \sum_{i=1}^d q_{c \ell} \underset{\boldsymbol{x}_t \sim p^{(\ell)}}{\mathbb{E}}\left[\mathrm{D}_{\mathrm{KL}}\left(p\left(x_t^{i} \mid \left(\mathbf{P a}_{\mathcal{G}^{*,\ell}}^i(<t), \mathbf{P a}_{\mathcal{G}^{*,\ell}}^i(t)\right)\right) \| f^c_i\left(\mathbf{P a}_{\mathcal{G}^c}^i(<t), \mathbf{P a}_{\mathcal{G}^c}^i(t)\right)\right)\right]\\&+\sum_{\ell=1}^K \sum_{i=1}^d\left(\sum_{c=1}^{N_{\mathrm{c}}} q_{c \ell}\right) \underset{\boldsymbol{x}_t \sim p^{(\ell)}}{\mathbb{E}}\left[\mathrm{H}\left(p\left(x_t^{i} \mid \left(\mathbf{P a}_{\mathcal{G}^{*,\ell}}^i(<t), \mathbf{P a}_{\mathcal{G}^{*,\ell}}^i(t)\right)\right)\right)\right] \\
& =\inf _\theta \sum_{c=1}^{N_{w}} \sum_{\ell=1}^K \sum_{i=1}^d q_{c \ell} \underset{\boldsymbol{x}_t \sim p^{(\ell)}}{\mathbb{E}}\left[\mathrm{D}_{\mathrm{KL}}\left(p\left(x_t^{i} \mid \left(\mathbf{P a}_{\mathcal{G}^{*,\ell}}^i(<t), \mathbf{P a}_{\mathcal{G}^{*,\ell}}^i(t)\right)\right) \| f^c_i\left(\mathbf{P a}_{\mathcal{G}^c}^i(<t), \mathbf{P a}_{\mathcal{G}^c}^i(t)\right)\right)\right]\\&+\sum_{\ell=1}^K \sum_{i=1}^d q_{\ell}^* \underset{\boldsymbol{x}_t \sim p^{(\ell)}}{\mathbb{E}}\left[\mathrm{H}\left(p\left(x_t^{i} \mid \left(\mathbf{P a}_{\mathcal{G}^{*,\ell}}^i(<t), \mathbf{P a}_{\mathcal{G}^{*,\ell}}^i(t)\right)\right)\right)\right] \\
\end{aligned}
\label{18}
\end{equation}
Note that $\theta$ could be the parameters of the neural networks used in Eq (\ref{eq1010}) for non linear causal relationship or $\theta = (\mathcal{G}^u)_{u \in \{1,..,N_w\}}$ for linear case Eq \ref{1313}.

For the score of ground truth (ignoring penalty terms):
\begin{equation}
    \begin{aligned}
-\mathcal{S}\left(\mathcal{G}^*, \mathcal{E}^*\right) 
& \rightarrow 0+\sum_{\ell=1}^K \sum_{i=1}^d q_{\ell}^* \underset{\boldsymbol{x}_t \sim p^{(\ell)}}{\mathbb{E}}\left[\mathrm{H}\left(p\left(x_t^{i} \mid \left(\mathbf{P a}_{\mathcal{G}^{*,\ell}}^i(<t), \mathbf{P a}_{\mathcal{G}^{*,\ell}}^i(t)\right)\right)\right)\right]    
\end{aligned}
\label{19}
\end{equation}

Combining Equation (\ref{18}) and Equation (\ref{19}) , we have (considering penalty terms): 
\begin{equation}
\begin{aligned}
\mathcal{S}\left(\mathcal{G}^*, \mathcal{E}^*\right)-\mathcal{S}(\mathcal{G}, \mathcal{E}) 
= & \inf _\theta \sum_{c=1}^{N_{w}} \sum_{\ell=1}^K \sum_{i=1}^d q_{c \ell} \\&\underset{\boldsymbol{x}_t \sim p^{(\ell)}}{\mathbb{E}}\left[\mathrm{D}_{\mathrm{KL}}\left(p\left(x_t^{i} \mid \left(\mathbf{P a}_{\mathcal{G}^{*,\ell}}^i(<t), \mathbf{P a}_{\mathcal{G}^{*,\ell}}^i(t)\right)\right) \| f^c_i\left(\mathbf{P a}_{\mathcal{G}^c}^i(<t), \mathbf{P a}_{\mathcal{G}^c}^i(t)\right)\right)\right] \\&+\lambda \left(\sum_{c=1}^{N_{w}}|\mathcal{G}^c|-\sum_{\ell=1}^K\left|\mathcal{G}^{*,\ell}\right|\right)
\end{aligned}
\label{20}
\end{equation}
The first term in Equation (\ref{20}) is the score term, others are penalty term.\\ 

In the following lines, our goal is to demonstrate that optimizing the score term ensures that all identified regimes will accurately match the real regimes. In other words, each estimated regime will be a true representation of an actual one. Additionally, by shifting samples from less significant regimes (regimes with few samples) to the most similar significant regimes, our variable $N_w$ will eventually stabilize at the value of K. To do this, we will proceed by contradiction:\\

Suppose the score term in Eq (\ref{20}) is optimized and there exists a regime $e$ that is \textbf{not pure}, i.e., there exist $a, b \in[K]$ with $a \neq b$ but $q_{e a}>0$ and $q_{e b}>0$. Since they are different distributions for two different regimes with two different causal graphs, there exists $i \in \{1,...,d\}$ such that $p\left(x_t^{i} \mid \left(\mathbf{P a}_{\mathcal{G}^{*,a}}^i(<t), \mathbf{P a}_{\mathcal{G}^{*,a}}^i(t)\right)\right)  \neq p\left(x_t^{i} \mid \left(\mathbf{P a}_{\mathcal{G}^{*,b}}^i(<t), \mathbf{P a}_{\mathcal{G}^{*,b}}^i(t)\right)\right) $. Then the score term in Equation (\ref{20}) has the following lower bound:

\begin{equation}
\begin{aligned}
& \inf _\theta \sum_{\ell=1}^K \sum_{i=1}^d q_{e \ell} \underset{\boldsymbol{x}_t \sim p^{(\ell)}}{\mathbb{E}}\left[\mathrm{D}_{\mathrm{KL}}\left(p\left(x_t^{i} \mid \left(\mathbf{P a}_{\mathcal{G}^{*,\ell}}^i(<t), \mathbf{P a}_{\mathcal{G}^{*,\ell}}^i(t)\right)\right) \| f^e_i\left(\mathbf{P a}_{\mathcal{G}^e}^i(<t), \mathbf{P a}_{\mathcal{G}^e}^i(t)\right)\right)\right] \\
\geq & \inf _\theta\Big\{q_{e a} \underset{\boldsymbol{x}_t \sim p^{(a)}}{\mathbb{E}}\left[\mathrm{D}_{\mathrm{KL}}\left(p\left(x_t^{i} \mid \left(\mathbf{P a}_{\mathcal{G}^{*,a}}^i(<t), \mathbf{P a}_{\mathcal{G}^{*,a}}^i(t)\right)\right) \| f^e_i\left(\mathbf{P a}_{\mathcal{G}^e}^i(<t), \mathbf{P a}_{\mathcal{G}^e}^i(t)\right)\right)\right]\\&+q_{e b} \underset{\boldsymbol{x}_t \sim p^{(b)}}{\mathbb{E}}\left[\mathrm{D}_{\mathrm{KL}}\left(p\left(x_t^{i} \mid \left(\mathbf{P a}_{\mathcal{G}^{*,b}}^i(<t), \mathbf{P a}_{\mathcal{G}^{*,b}}^i(t)\right)\right) \| f^e_i\left(\mathbf{P a}_{\mathcal{G}^e}^i(<t), \mathbf{P a}_{\mathcal{G}^e}^i(t)\right)\right)\right]\Big\}
\end{aligned}
\label{21}
\end{equation}
As we assumed that the score term in Eq (\ref{20}) is optimized, it means that: 

\begin{equation}
    \begin{aligned}
       &0= \inf _\theta \sum_{c=1}^{N_{w}} \sum_{\ell=1}^K \sum_{i=1}^d q_{c \ell} \underset{\boldsymbol{x}_t \sim p^{(\ell)}}{\mathbb{E}}\left[\mathrm{D}_{\mathrm{KL}}\left(p\left(x_t^{i} \mid \left(\mathbf{P a}_{\mathcal{G}^{*,\ell}}^i(<t), \mathbf{P a}_{\mathcal{G}^{*,\ell}}^i(t)\right)\right) \| f^c_i\left(\mathbf{P a}_{\mathcal{G}^c}^i(<t), \mathbf{P a}_{\mathcal{G}^c}^i(t)\right)\right)\right]\\
       &\Rightarrow 0=\inf _\theta \sum_{\ell=1}^K \sum_{i=1}^d q_{e \ell} \underset{\boldsymbol{x}_t \sim p^{(\ell)}}{\mathbb{E}}\left[\mathrm{D}_{\mathrm{KL}}\left(p\left(x_t^{i} \mid \left(\mathbf{P a}_{\mathcal{G}^{*,\ell}}^i(<t), \mathbf{P a}_{\mathcal{G}^{*,\ell}}^i(t)\right)\right) \| f^e_i\left(\mathbf{P a}_{\mathcal{G}^e}^i(<t), \mathbf{P a}_{\mathcal{G}^e}^i(t)\right)\right)\right]\\
       & \Rightarrow \left\{\begin{array}{l}
\mathrm{D}_{\mathrm{KL}}\left(p\left(x_t^{i} \mid \left(\mathbf{P a}_{\mathcal{G}^{*,a}}^i(<t), \mathbf{P a}_{\mathcal{G}^{*,a}}^i(t)\right)\right) \| f^e_i\left(\mathbf{P a}_{\mathcal{G}^e}^i(<t), \mathbf{P a}_{\mathcal{G}^e}^i(t)\right)\right) = 0\\
\mathrm{D}_{\mathrm{KL}}\left(p\left(x_t^{i} \mid \left(\mathbf{P a}_{\mathcal{G}^{*,b}}^i(<t), \mathbf{P a}_{\mathcal{G}^{*,b}}^i(t)\right)\right) \| f^e_i\left(\mathbf{P a}_{\mathcal{G}^e}^i(<t), \mathbf{P a}_{\mathcal{G}^e}^i(t)\right)\right) = 0

\end{array}\right.\\
& \Rightarrow \forall i \in \{1,...,d\}: p\left(x_t^{i} \mid \left(\mathbf{P a}_{\mathcal{G}^{*,a}}^i(<t), \mathbf{P a}_{\mathcal{G}^{*,a}}^i(t)\right)\right) = p\left(x_t^{i} \mid \left(\mathbf{P a}_{\mathcal{G}^{*,b}}^i(<t), \mathbf{P a}_{\mathcal{G}^{*,b}}^i(t)\right)\right)
    \end{aligned}
    \label{22}
\end{equation}
and the last line, Eq (\ref{22}), is a contradiction because the two distributions represent two different regimes with two different graphs. Hence, if the score term of Eq (\ref{20}) is optimized all the estimated regimes will be pure. \\

\textbf{First case:} If we matched the samples of less significant regimes to the wrong regimes, the regime is not pure and then the score term is not optimized (contradiction).\\

\textbf{Second case:} If we eliminate a lot of regimes such that $N_w \leq K-1$, at least one of our estimated regimes will not be pure and this contradicts the assumption of optimized score term (same reasoning).\\
Based on this reasoning, optimizing the score term of Equation (\ref{20}) will ensure convergence to the true number of regimes and also every regime will be pure.\\

\subsubsection{In case of edge disagreement  $
\mathcal{S}\left(\mathcal{G}^*, \mathcal{E}^*\right)>\mathcal{S}(\hat{\mathcal{G}},  \hat{\mathcal{E}})
$}
Now we will show that Eq (\ref{20} )is positive, if $ \exists u \in \{1,...,K\} \text{  s.t  } \hat{\mathcal{G}^u}  \text{ disagrees with}$ $\mathcal{G}^{*,u}$ on instantaneous or/and time lagged link.\\
To simplify the notation, \textcolor{cyan}{we denote by $p^{(u)}$ the distribution $p(.|\mathcal{G}^{u})$} the optimal distribution that describes the CGM of regime $u$. We assume that each estimated regime $\hat{\mathcal{E}}_c\left(c \in\left\{1, \ldots, N_{w}\right\}, N_{w} \geq K\right)$ contains samples from same true regime. Then Equation (\ref{20}) has lower bound:
\begin{equation}
\begin{array}{r}
\inf _\theta \sum_{\ell=1}^{K} q_{\ell}^* \underset{\boldsymbol{x}_t \sim p^{(\ell)}}{\mathbb{E}} \mathrm{D}_{\mathrm{KL}}\left(p^{(\ell)} \| f^{\ell}\right) \\
\geq\left(\min _{\ell} q_{\ell}^*\right) \inf _\theta \sum_{\ell=1}^K \mathrm{D}_{\mathrm{KL}}\left(p^{(\ell)} \| f^{\ell}\right)
\end{array}
\label{23}
\end{equation}
Equation (\ref{23}) is positive if and only if $\eta(\mathcal{G})$ is positive.
\begin{equation}
\eta(\mathcal{G}):=\inf _\theta \sum_{\ell=1}^K \mathrm{D}_{\mathrm{KL}}\left(p^{(\ell)} \| f^{\ell}\right)
\label{24}
\end{equation}
Let assume that $\exists r \in \in \{1,...,K\} \text{  s.t  } \hat{\mathcal{G}^r}  \text{ disagrees with}$ $\mathcal{G}^{*,r}$ on instantaneous or/and time lagged link. We follow the same intuition as \cite{gong2022rhino, peters2013causal, peters2017elements}, we will show that $\mathrm{D}_{\mathrm{KL}}\left(p^{(r)} \| f^{r}\right)$ is positive in two cases:
\begin{itemize}
    \item \textbf{Disagreement on lagged parents only. } This means that for all $t \in[\mathcal{S}+1, T]$, the instantaneous connections at $t$ for $\hat{\mathcal{G}^r}$ and $\mathcal{G}^{*,r}$ are the same, and $\exists t \in[\mathcal{S}+1, T] \text{ and } i \in \{1,...,d\}$ such that $\mathbf{P a}_{\mathcal{G}^{*,r}}^{x_t^i}(<t) \neq \overline{\mathbf{P a}}_{\hat{\mathcal{G}^r}}^{x_t^i}(<t)$. We can use a similar argument as the theorem 1 in \cite{peters2013causal}. Without loss of generality, we assume under $\hat{\mathcal{G}^r}$, we have $x^j_{t-\tau} \rightarrow x^i_t$ and there is no connections between them under $\mathcal{G}^{*,r}$. Thus, from Markov conditions, we have
$$
x^{i}_{t} \perp\!\!\!\perp x^{j}_{t-\tau} \mid \mathbf{P a}_{\mathcal{G}^{*,r}}^{x_t^i}(<t) \cup \mathrm{ND}_{t}^{x^i} \backslash\left\{x^{i}_t, x^j_{t-\tau}\right\}
$$
under $\mathcal{G}^{*,r}$, where $\mathrm{ND}_{t}^{x^i}$ are the non-descendants of node $x^{i}_{t}$ at some time $t$. However, from the causal minimality and Proposition 6.16 in \cite{peters2017elements}, we have
$$
x^{i}_{t} \not\!\perp\!\!\!\perp x^{j}_{t-\tau} \mid \overline{\mathbf{P a}}_{\hat{\mathcal{G}^r}}^{x_t^i}(<t)  \cup \mathrm{ND}_{t}^{x^i} \backslash\left\{x^{i}_t, x^j_{t-\tau}\right\}
$$
under $\hat{\mathcal{G}^r}$, and we have $\mathrm{D}_{\mathrm{KL}}\left(p^{(r)} \| f^{r}\right) \neq 0$
\item \textbf{Disagreement on instantaneous parents. } In this Section we will use two different results one for the linear and the other one for the non linear case. 
\begin{itemize}
    \item \textit{Linear case. } For this case, we will use Theorem 1 in \cite{peters2014identifiability}. In this theorem, the author confirms that the graph is identifiable for linear models with Gaussian additive noise, if for each $j \in\{1, \ldots, d\}$, the weights of the causal relations $\beta_{j k} \neq 0$ for all $k \in \mathbf{P A}_j^{\mathcal{G}_0}$. For our instantaneous links, we have all the weights of the parents are non null. Hence, the instantaneous links are identifiable. Otherwise if $\mathrm{D}_{\mathrm{KL}}\left(p^{(r)} \| f^{r}\right) \neq 0$
    \item \textit{Non linear case. } Using Corollary 30 from \cite{peters2014causal}, in which they state that in the case of Gaussian independent noise and non linear mixing functions, the graphs are identifiable. Hence, our instantaneous links are identifiable, otherwise, $\mathrm{D}_{\mathrm{KL}}\left(p^{(r)} \| f^{r}\right) \neq 0$. 
\end{itemize}
\end{itemize}
Based on the above reasoning, we can show that if $\exists r \in \in \{1,...,K\} \text{  s.t.,  } \hat{\mathcal{G}^r}  \text{ disagree with}$ $\mathcal{G}^{*,r}$ on instantaneous or/and time lagged links, $\mathrm{D}_{\mathrm{KL}}\left(p^{(r)} \| f^{r}\right) \neq 0$. 
\\Thus, $\eta(\mathcal{G})>0$. Then as we assume in Theorem \ref{theo:1} that $\lambda$ is sufficient small would implies Equation (\ref{22}) is positive.\\
If $|\hat{\mathcal{G}^r}| \geq\left|\mathcal{G}^{*,r}\right|$ then clearly Eq \ref{22} is positive. Let $\mathbb{G}^{+}:=\left\{\hat{\mathcal{G}^r} \in \mathbb{G}|| \hat{\mathcal{G}^r}|<| \mathcal{G}^{*,r} \mid\right\}$. To make sure that we have $\mathcal{S}\left(\mathcal{G}^*, \mathcal{E}^*\right)-\mathcal{S}(\mathcal{G}, \mathcal{E}) >0$ for all $\mathcal{G} \in \mathbb{G}^{+}$, we need to pick $\lambda$ sufficiently small. Choosing $0<\lambda<\min _{\mathcal{G} \in \mathbb{G}^{+}} \frac{\eta(\mathcal{G})}{\left(\sum_{c=1}^{N_{w}}|\mathcal{G}^c|-\sum_{\ell=1}^K\left|\mathcal{G}^{*,\ell}\right|\right)}$ is sufficient.
\subsection{Proof of theorem \ref{theo:2}}
\label{app_proof2}
Our objective is to prove that for any estimations $(\mathcal{G}, \mathcal{E})$ and $(\mathcal{G'}, \mathcal{E'})$ such that $(\mathcal{G}, \mathcal{E})$ is closer in terms of Kullback–Leibler to the optimal solution $\left(\mathcal{G}^*, \mathcal{E}^*\right)$ than $(\mathcal{G'}, \mathcal{E'})$: $\mathcal{S}(\mathcal{G}, \mathcal{E})>\mathcal{S}(\mathcal{G'}, \mathcal{E'}).$ 
Our goal is to demonstrate that, for any estimation $(\mathcal{G}, \mathcal{E})$, one that is closer in terms of Kullback-Leibler (KL) divergence to the optimal solution $\left(\mathcal{G}^*, \mathcal{E}^*\right)$ will have a higher score estimated by CASTOR compared to another estimation $(\mathcal{G'}, \mathcal{E'})$. To clarify, when we mention "closer to $\left(\mathcal{G}^*, \mathcal{E}^*\right)$ in terms of KL," we are referring to the degree of similarity to the optimal solution $\left(\mathcal{G}^*, \mathcal{E}^*\right)$. In other terms for every regime $\ell$:\\

\begin{equation}
\begin{aligned}
& \mathrm{D}_{\mathrm{KL}}\left(p\left(x_t^{i} \mid \left(\mathbf{P a}_{\mathcal{G}^{*,\ell}}^i(<t), \mathbf{P a}_{\mathcal{G}^{*,\ell}}^i(t)\right)\right) \| f^{\ell}_i\left(\mathbf{P a}_{\textcolor{red}{\mathcal{G}^\ell}}^i(<t), \mathbf{P a}_{\textcolor{red}{\mathcal{G}^\ell}}^i(t)\right)\right)\\
\leq & \mathrm{D}_{\mathrm{KL}}\left(p\left(x_t^{i} \mid \left(\mathbf{P a}_{\mathcal{G}^{*,\ell}}^i(<t), \mathbf{P a}_{\mathcal{G}^{*,\ell}}^i(t)\right)\right) \| f^{\ell}_i\left(\mathbf{P a}_{\textcolor{red}{\mathcal{G'}^\ell}}^i(<t), \mathbf{P a}_{\textcolor{red}{\mathcal{G'}^\ell}}^i(t)\right)\right)
\end{aligned}
\end{equation}
\textbf{First case: } Let's assume, in this first scenario, that both suboptimal estimations ($(\mathcal{G'}, \mathcal{E}^*)$ and $(\mathcal{G}, \mathcal{E}^*)$) accurately detect the regimes, with the only distinction lying in the estimation of the graph.\\

We know that the score function of each estimation after the optimization procedure could be written as the following:
\begin{equation}
\begin{aligned}
    -\mathcal{S}(\mathcal{G}, \mathcal{E}^*) &= \sum_{\ell=1}^K \sum_{i=1}^d q^{*}_{\ell} \underset{\boldsymbol{x}_t \sim p^{(\ell)}}{\mathbb{E}}\left[\mathrm{D}_{\mathrm{KL}}\left(p\left(x_t^{i} \mid \left(\mathbf{P a}_{\mathcal{G}^{*,\ell}}^i(<t), \mathbf{P a}_{\mathcal{G}^{*,\ell}}^i(t)\right)\right) \| f^\ell_i\left(\mathbf{P a}_{\textcolor{red}{\mathcal{G}^\ell}}^i(<t), \mathbf{P a}_{\textcolor{red}{\mathcal{G}^\ell}}^i(t)\right)\right)\right]\\&+\sum_{\ell=1}^K \sum_{i=1}^d q_{\ell}^* \underset{\boldsymbol{x}_t \sim p^{(\ell)}}{\mathbb{E}}\left[\mathrm{H}\left(p\left(x_t^{i} \mid \left(\mathbf{P a}_{\mathcal{G}^{*,\ell}}^i(<t), \mathbf{P a}_{\mathcal{G}^{*,\ell}}^i(t)\right)\right)\right)\right]\\ 
    -\mathcal{S}(\mathcal{G'}, \mathcal{E}^*) &=  \sum_{\ell=1}^K \sum_{i=1}^d q^{*}_{\ell} \underset{\boldsymbol{x}_t \sim p^{(\ell)}}{\mathbb{E}}\left[\mathrm{D}_{\mathrm{KL}}\left(p\left(x_t^{i} \mid \left(\mathbf{P a}_{\mathcal{G}^{*,\ell}}^i(<t), \mathbf{P a}_{\mathcal{G}^{*,\ell}}^i(t)\right)\right) \| f^\ell_i\left(\mathbf{P a}_{\textcolor{red}{\mathcal{G'}^\ell}}^i(<t), \mathbf{P a}_{\textcolor{red}{\mathcal{G'}^\ell}}^i(t)\right)\right)\right]\\&+\sum_{\ell=1}^K \sum_{i=1}^d q_{\ell}^* \underset{\boldsymbol{x}_t \sim p^{(\ell)}}{\mathbb{E}}\left[\mathrm{H}\left(p\left(x_t^{i} \mid \left(\mathbf{P a}_{\mathcal{G}^{*,\ell}}^i(<t), \mathbf{P a}_{\mathcal{G}^{*,\ell}}^i(t)\right)\right)\right)\right]\\ 
    \mathcal{S}(\mathcal{G}, \mathcal{E}^*)-\mathcal{S}(\mathcal{G'}, \mathcal{E}^*) &=  \sum_{\ell=1}^K \sum_{i=1}^d q^{*}_{\ell} \underset{\boldsymbol{x}_t \sim p^{(\ell)}}{\mathbb{E}}[\mathrm{D}_{\mathrm{KL}}\left(p\left(x_t^{i} \mid \left(\mathbf{P a}_{\mathcal{G}^{*,\ell}}^i(<t), \mathbf{P a}_{\mathcal{G}^{*,\ell}}^i(t)\right)\right) \| f^\ell_i\left(\mathbf{P a}_{\textcolor{red}{\mathcal{G'}^\ell}}^i(<t), \mathbf{P a}_{\textcolor{red}{\mathcal{G'}^\ell}}^i(t)\right)\right)\\&-\mathrm{D}_{\mathrm{KL}}\left(p\left(x_t^{i} \mid \left(\mathbf{P a}_{\mathcal{G}^{*,\ell}}^i(<t), \mathbf{P a}_{\mathcal{G}^{*,\ell}}^i(t)\right)\right) \| f^\ell_i\left(\mathbf{P a}_{\textcolor{red}{\mathcal{G}^\ell}}^i(<t), \mathbf{P a}_{\textcolor{red}{\mathcal{G}^\ell}}^i(t)\right)\right)]\\ 
    \mathcal{S}(\mathcal{G}, \mathcal{E}^*)-\mathcal{S}(\mathcal{G'}, \mathcal{E}^*) &\geq 0\\
    \end{aligned}
\end{equation}

In the first case, we demonstrated that if two estimations differ in the graph learning component, the one closer to the optimal solution will have a higher score. The last inequality is correct even when we add the sparsity term because, we can pick $\lambda$ sufficiently small, to ensure that we have $\mathcal{S}\left(\mathcal{G}, \mathcal{E}^*\right)-\mathcal{S}(\mathcal{G'}, \mathcal{E}^*) >0$.\\

\textbf{Second case: } Let's assume that both suboptimal estimations ($(\mathcal{G}, \mathcal{E'})$ and $(\mathcal{G}, \mathcal{E})$) learn identical causal graphs for all regimes. However, for the estimation $\mathcal{E'}$, there exists at least one regime $a$ that misclassifies more $S$ samples and incorrectly assigns them to regime $b$.\\

We know that the score function of each estimation after the optimization procedure could be written as the following:
$$
\begin{aligned}
-\mathcal{S}(\mathcal{G}, \mathcal{E}) &= \sum_{c=1}^{N_{w}} \sum_{\ell=1}^K \sum_{i=1}^d q_{c \ell} \underset{\boldsymbol{x}_t \sim p^{(\ell)}}{\mathbb{E}}\left[\mathrm{D}_{\mathrm{KL}}\left(p\left(x_t^{i} \mid \left(\mathbf{P a}_{\mathcal{G}^{*,\ell}}^i(<t), \mathbf{P a}_{\mathcal{G}^{*,\ell}}^i(t)\right)\right) \| f^c_i\left(\mathbf{P a}_{\mathcal{G}^c}^i(<t), \mathbf{P a}_{\mathcal{G}^c}^i(t)\right)\right)\right]\\&+\sum_{\ell=1}^K \sum_{i=1}^d q_{\ell}^* \underset{\boldsymbol{x}_t \sim p^{(\ell)}}{\mathbb{E}}\left[\mathrm{H}\left(p\left(x_t^{i} \mid \left(\mathbf{P a}_{\mathcal{G}^{*,\ell}}^i(<t), \mathbf{P a}_{\mathcal{G}^{*,\ell}}^i(t)\right)\right)\right)\right] \\
-\mathcal{S}(\mathcal{G}, \mathcal{E'}) &= \sum_{c=1}^{N_{w}} \sum_{\ell=1}^K \sum_{i=1}^d q'_{c \ell} \underset{\boldsymbol{x}_t \sim p^{(\ell)}}{\mathbb{E}}\left[\mathrm{D}_{\mathrm{KL}}\left(p\left(x_t^{i} \mid \left(\mathbf{P a}_{\mathcal{G}^{*,\ell}}^i(<t), \mathbf{P a}_{\mathcal{G}^{*,\ell}}^i(t)\right)\right) \| f^c_i\left(\mathbf{P a}_{\mathcal{G}^c}^i(<t), \mathbf{P a}_{\mathcal{G}^c}^i(t)\right)\right)\right]\\&+\sum_{\ell=1}^K \sum_{i=1}^d q_{\ell}^* \underset{\boldsymbol{x}_t \sim p^{(\ell)}}{\mathbb{E}}\left[\mathrm{H}\left(p\left(x_t^{i} \mid \left(\mathbf{P a}_{\mathcal{G}^{*,\ell}}^i(<t), \mathbf{P a}_{\mathcal{G}^{*,\ell}}^i(t)\right)\right)\right)\right] \\
\end{aligned}
$$
As previously explained, for the estimation $\mathcal{E'}$, S samples from regime $a$ are misclassified as belonging to regime $b$. This implies: $q'_{a a} = q_{a a} - \frac{S}{|\mathcal{T}|}$ and $q'_{b a} = q_{b a} + \frac{S}{|\mathcal{T}|}$, and we have the difference between the 2 scores is written as follows:
\begin{equation}
\begin{aligned}
\mathcal{S}(\mathcal{G}, \mathcal{E})-\mathcal{S}(\mathcal{G}, \mathcal{E'}) &= \sum_{i=1}^d \frac{S}{|\mathcal{T}|} \underset{\boldsymbol{x}_t \sim p^{(a)}}{\mathbb{E}}[ D_{\mathrm{KL}}\left(p\left(x_t^{i} \mid \left(\mathbf{P a}_{\mathcal{G}^{*,a}}^i(<t), \mathbf{P a}_{\mathcal{G}^{*,a}}^i(t)\right)\right) \| f^b_i\left(\mathbf{P a}_{\mathcal{G}^b}^i(<t), \mathbf{P a}_{\mathcal{G}^b}^i(t)\right)\right)\\ &- D_{\mathrm{KL}}\left(p\left(x_t^{i} \mid \left(\mathbf{P a}_{\mathcal{G}^{*,a}}^i(<t), \mathbf{P a}_{\mathcal{G}^{*,a}}^i(t)\right)\right) \| f^a_i\left(\mathbf{P a}_{\mathcal{G}^a}^i(<t), \mathbf{P a}_{\mathcal{G}^a}^i(t)\right)\right)]\\ 
\mathcal{S}(\mathcal{G}, \mathcal{E}^*)-\mathcal{S}(\mathcal{G'}, \mathcal{E}^*) &\geq 0\\
\end{aligned}
\end{equation}
The aforementioned equation is positive because, by definition, the distribution estimated by CASTOR for regime $a$ is closer to the true distribution of regime $b$ than the estimation of the distribution for regime $b$.

\textbf{Third case: } Let's assume that there exists at least one regime $a$ that misclassifies more samples $S$ and incorrectly assigns them to regime $b$. In these two regimes, the two suboptimal estimations yield different causal graphs. However, as previously described, the regime indices learned by $\mathcal{E}$ are closer to $\mathcal{E}^{*}$. In other words, $\mathcal{E'}$ misclassifies more samples $S$ from regime $a$ and incorrectly assigns them to regime $b$. Additionally, the graph inferred by $\mathcal{G}$ is closer to the optimal solution than the graph $\mathcal{G'}$ estimated by the second suboptimal method.\\

We assume that $q_{ba} \ll q_{aa}$, signifying that the number of well-classified regime samples is greater than the number of misclassified samples. This assumption is reasonable because altering graphs between regimes will cause changes in the mean of our mixtures $f^u$. Furthermore, given the higher dimension, making slight modifications such as deleting a few edges and adding new ones will result in distinct mixtures. Consequently, the number of samples that could be misclassified will be low.\\

We know that the score function of each estimation after the optimization procedure could be written as the following:
\begin{equation}
\begin{aligned}
-\mathcal{S}(\mathcal{G}, \mathcal{E}) &= \sum_{c=1}^{N_{w}} \sum_{\ell=1}^K \sum_{i=1}^d q_{c \ell} \underset{\boldsymbol{x}_t \sim p^{(\ell)}}{\mathbb{E}}\left[\mathrm{D}_{\mathrm{KL}}\left(p\left(x_t^{i} \mid \left(\mathbf{P a}_{\mathcal{G}^{*,\ell}}^i(<t), \mathbf{P a}_{\mathcal{G}^{*,\ell}}^i(t)\right)\right) \| f^c_i\left(\mathbf{P a}_{\mathcal{G}^c}^i(<t), \mathbf{P a}_{\mathcal{G}^c}^i(t)\right)\right)\right]\\&+\sum_{\ell=1}^K \sum_{i=1}^d q_{\ell}^* \underset{\boldsymbol{x}_t \sim p^{(\ell)}}{\mathbb{E}}\left[\mathrm{H}\left(p\left(x_t^{i} \mid \left(\mathbf{P a}_{\mathcal{G}^{*,\ell}}^i(<t), \mathbf{P a}_{\mathcal{G}^{*,\ell}}^i(t)\right)\right)\right)\right] \\
\end{aligned}
\end{equation}
\begin{equation}
\begin{aligned}
-\mathcal{S}(\mathcal{G'}, \mathcal{E'}) &= \sum_{c=1}^{N_{w}} \sum_{\ell=1}^K \sum_{i=1}^d q'_{c \ell} \underset{\boldsymbol{x}_t \sim p^{(\ell)}}{\mathbb{E}}\left[\mathrm{D}_{\mathrm{KL}}\left(p\left(x_t^{i} \mid \left(\mathbf{P a}_{\mathcal{G}^{*,\ell}}^i(<t), \mathbf{P a}_{\mathcal{G}^{*,\ell}}^i(t)\right)\right) \| f^c_i\left(\mathbf{P a}_{\mathcal{G'}^c}^i(<t), \mathbf{P a}_{\mathcal{G'}^c}^i(t)\right)\right)\right]\\&+\sum_{\ell=1}^K \sum_{i=1}^d q_{\ell}^* \underset{\boldsymbol{x}_t \sim p^{(\ell)}}{\mathbb{E}}\left[\mathrm{H}\left(p\left(x_t^{i} \mid \left(\mathbf{P a}_{\mathcal{G}^{*,\ell}}^i(<t), \mathbf{P a}_{\mathcal{G}^{*,\ell}}^i(t)\right)\right)\right)\right] \\
\mathcal{S}(\mathcal{G}, \mathcal{E})-\mathcal{S}(\mathcal{G'}, \mathcal{E'}) &= \sum_{i=1}^d \frac{S}{|\mathcal{T}|} \underset{\boldsymbol{x}_t \sim p^{(a)}}{\mathbb{E}}[ D_{\mathrm{KL}}\left(p\left(x_t^{i} \mid \left(\mathbf{P a}_{\mathcal{G}^{*,a}}^i(<t), \mathbf{P a}_{\mathcal{G}^{*,a}}^i(t)\right)\right) \| f^b_i\left(\mathbf{P a}_{\mathcal{G'}^b}^i(<t), \mathbf{P a}_{\mathcal{G'}^b}^i(t)\right)\right)\\ &- D_{\mathrm{KL}}\left(p\left(x_t^{i} \mid \left(\mathbf{P a}_{\mathcal{G}^{*,a}}^i(<t), \mathbf{P a}_{\mathcal{G}^{*,a}}^i(t)\right)\right) \| f^a_i\left(\mathbf{P a}_{\mathcal{G'}^a}^i(<t), \mathbf{P a}_{\mathcal{G'}^a}^i(t)\right)\right)]\\ 
&+\sum_{i=1}^d q_{aa} \underset{\boldsymbol{x}_t \sim p^{(a)}}{\mathbb{E}}[ D_{\mathrm{KL}}\left(p\left(x_t^{i} \mid \left(\mathbf{P a}_{\mathcal{G}^{*,a}}^i(<t), \mathbf{P a}_{\mathcal{G}^{*,a}}^i(t)\right)\right) \| f^a_i\left(\mathbf{P a}_{\mathcal{G'}^a}^i(<t), \mathbf{P a}_{\mathcal{G'}^a}^i(t)\right)\right)\\ &- D_{\mathrm{KL}}\left(p\left(x_t^{i} \mid \left(\mathbf{P a}_{\mathcal{G}^{*,a}}^i(<t), \mathbf{P a}_{\mathcal{G}^{*,a}}^i(t)\right)\right) \| f^a_i\left(\mathbf{P a}_{\mathcal{G}^a}^i(<t), \mathbf{P a}_{\mathcal{G}^a}^i(t)\right)\right)]\\ 
&+\sum_{i=1}^d q_{ba} \underset{\boldsymbol{x}_t \sim p^{(a)}}{\mathbb{E}}[ D_{\mathrm{KL}}\left(p\left(x_t^{i} \mid \left(\mathbf{P a}_{\mathcal{G}^{*,a}}^i(<t), \mathbf{P a}_{\mathcal{G}^{*,a}}^i(t)\right)\right) \| f^b_i\left(\mathbf{P a}_{\mathcal{G'}^b}^i(<t), \mathbf{P a}_{\mathcal{G'}^b}^i(t)\right)\right)\\ &- D_{\mathrm{KL}}\left(p\left(x_t^{i} \mid \left(\mathbf{P a}_{\mathcal{G}^{*,a}}^i(<t), \mathbf{P a}_{\mathcal{G}^{*,a}}^i(t)\right)\right) \| f^b_i\left(\mathbf{P a}_{\mathcal{G}^b}^i(<t), \mathbf{P a}_{\mathcal{G}^b}^i(t)\right)\right)]\\ 
\mathcal{S}(\mathcal{G}, \mathcal{E})-\mathcal{S}(\mathcal{G'}, \mathcal{E'}) & \approx \sum_{i=1}^d \frac{S}{|\mathcal{T}|} \underset{\boldsymbol{x}_t \sim p^{(a)}}{\mathbb{E}}[ D_{\mathrm{KL}}\left(p\left(x_t^{i} \mid \left(\mathbf{P a}_{\mathcal{G}^{*,a}}^i(<t), \mathbf{P a}_{\mathcal{G}^{*,a}}^i(t)\right)\right) \| f^b_i\left(\mathbf{P a}_{\mathcal{G'}^b}^i(<t), \mathbf{P a}_{\mathcal{G'}^b}^i(t)\right)\right)\\ &- D_{\mathrm{KL}}\left(p\left(x_t^{i} \mid \left(\mathbf{P a}_{\mathcal{G}^{*,a}}^i(<t), \mathbf{P a}_{\mathcal{G}^{*,a}}^i(t)\right)\right) \| f^a_i\left(\mathbf{P a}_{\mathcal{G'}^a}^i(<t), \mathbf{P a}_{\mathcal{G'}^a}^i(t)\right)\right)]\\ 
&+\sum_{i=1}^d q_{aa} \underset{\boldsymbol{x}_t \sim p^{(a)}}{\mathbb{E}}[ D_{\mathrm{KL}}\left(p\left(x_t^{i} \mid \left(\mathbf{P a}_{\mathcal{G}^{*,a}}^i(<t), \mathbf{P a}_{\mathcal{G}^{*,a}}^i(t)\right)\right) \| f^a_i\left(\mathbf{P a}_{\mathcal{G'}^a}^i(<t), \mathbf{P a}_{\mathcal{G'}^a}^i(t)\right)\right)\\ &- D_{\mathrm{KL}}\left(p\left(x_t^{i} \mid \left(\mathbf{P a}_{\mathcal{G}^{*,a}}^i(<t), \mathbf{P a}_{\mathcal{G}^{*,a}}^i(t)\right)\right) \| f^a_i\left(\mathbf{P a}_{\mathcal{G}^a}^i(<t), \mathbf{P a}_{\mathcal{G}^a}^i(t)\right)\right)]\\
\mathcal{S}(\mathcal{G}, \mathcal{E}^*)-\mathcal{S}(\mathcal{G'}, \mathcal{E}^*) &\geq 0\\
\end{aligned}
\end{equation}
The last inequality is correct for these two last cases, even when we add the sparsity term because, we can pick $\lambda$ sufficiently small, to ensure that we have $\mathcal{S}\left(\mathcal{G}, \mathcal{E}\right)-\mathcal{S}(\mathcal{G'}, \mathcal{E'}) >0$.
\newpage
\section{Illustration of CASTOR's estimated graphs}
\subsection{Illustration of the estimated graphs by CASTOR: Linear case, 5 regimes with $L=1$}
\begin{figure}[h]
\centering
\includegraphics[scale=0.4]{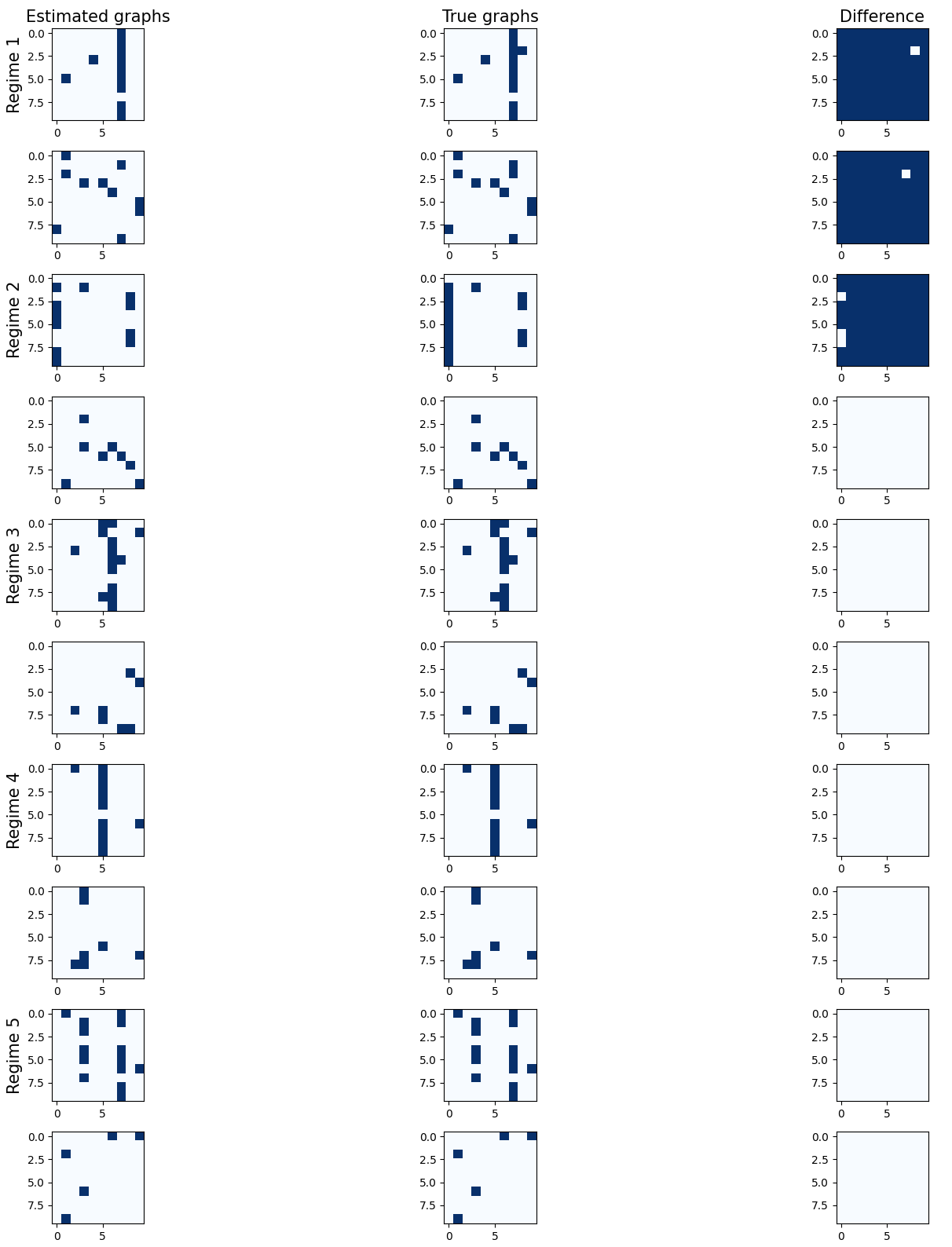}
\vspace{-1.5em}
\caption{The estimated temporal causal graphs for five regimes (\textbf{Linear case}) consist of one matrix of 10 rows and 10 columns representing instantaneous links and another of 10 rows and 10 columns delineating time-lagged relations (with a maximum lag $L=1$ in this case). Dark blue indicates a value of one (presence of an edge), while sky blue symbolizes a value of 0 (absence of an edge). The second column displays the groundtruth causal graphs, and the final column highlights the difference between the estimated and true graphs.}
\vspace{-1.5em}
\label{fig4}
\end{figure}
\newpage
\subsection{Illustration of the estimated graphs by CASTOR: Linear case, 2 regimes with $L=2$}
\begin{figure}[h]
\centering
\includegraphics[scale=0.55]{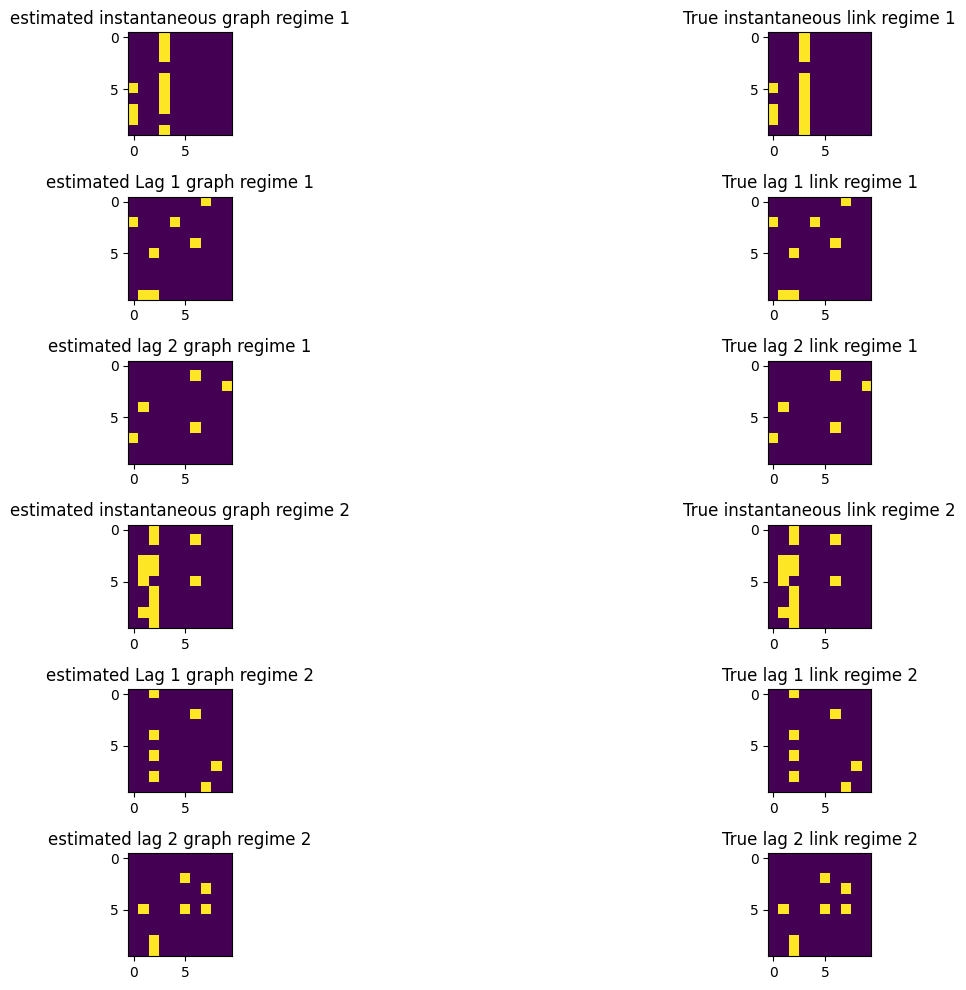}
\vspace{-0.5em}
\caption{
The estimated temporal causal graphs for two regimes \textbf{(non Linear case)}, with one matrix representing instantaneous links and another delineating time-lagged relations. The second column showcases the actual causal graphs, while the final column highlights the discrepancies between the estimated and true graphs. Yellow indicates a value of one (presence of an edge), while black symbolizes a value of 0 (absence of an edge).}
\vspace{-1.5em}
\label{fig5}
\end{figure}
\newpage
\subsection{Illustration of the estimated graphs by CASTOR: Non-linear case, 3 regimes with $L=1$}
\begin{figure}[h]
\centering
\includegraphics[scale=0.4]{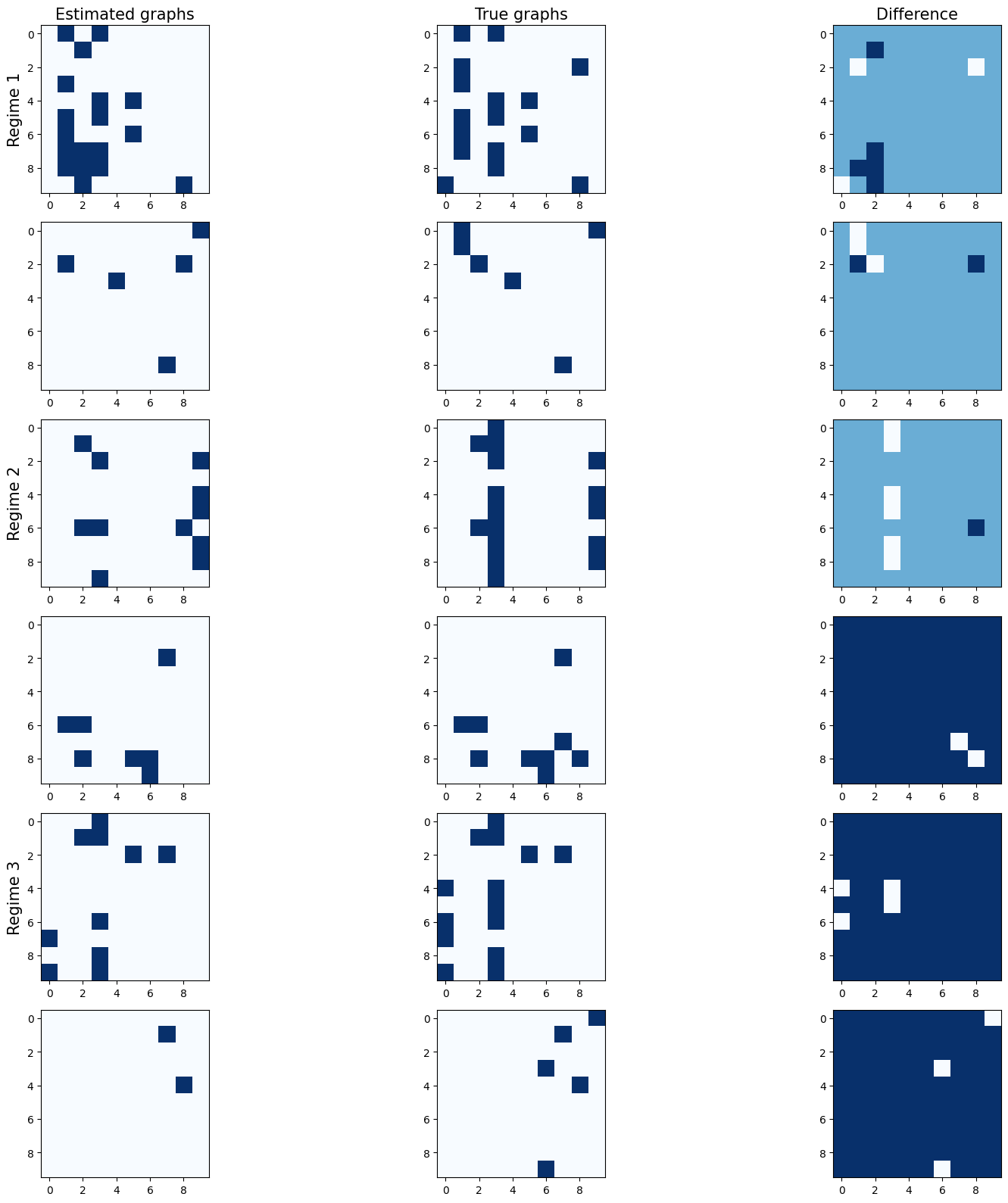}
\vspace{-0.5em}
\caption{
The estimated temporal causal graphs for three regimes (\textbf{Non-Linear case}) consist of one matrix of 10 rows and 10 columns representing instantaneous links and another of 10 rows and 10 columns delineating time-lagged relations (with a maximum lag $L=1$ in this case). Dark blue indicates a value of one (presence of an edge), while sky blue symbolizes a value of 0 (absence of an edge). The second column displays the groundtruth causal graphs, and the final column highlights the difference between the estimated and true graphs.}
\vspace{-1.5em}
\label{fig7}
\end{figure}

\newpage
\subsection{Illustration of the estimated graphs by CASTOR: Non-linear case, 5 regimes with $L=1$}
\begin{figure}[h]
\centering
\includegraphics[scale=0.4]{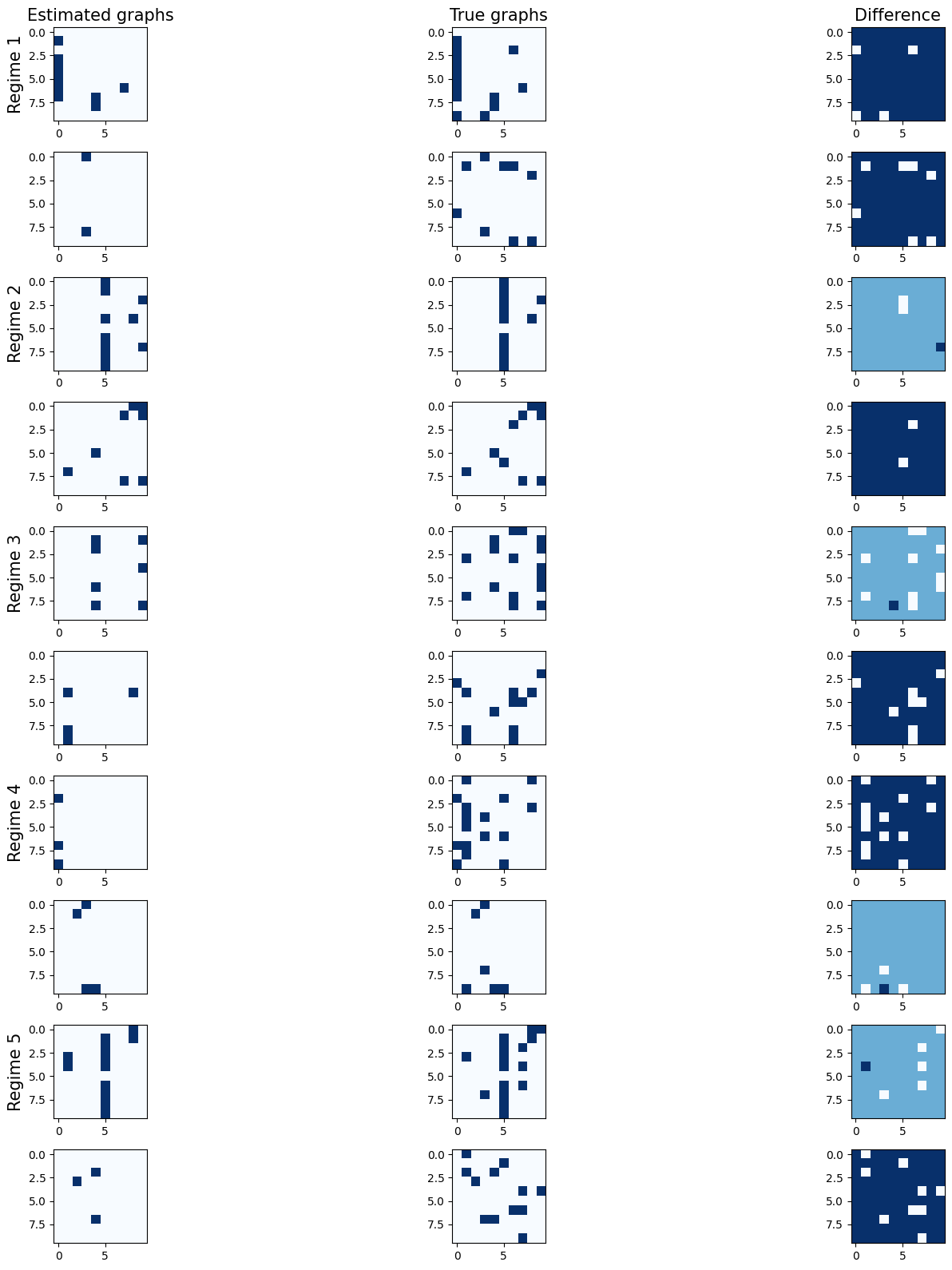}
\vspace{-0.5em}
\caption{
The estimated temporal causal graphs for five regimes (\textbf{Non-Linear case}) consist of one matrix of 10 rows and 10 columns representing instantaneous links and another of 10 rows and 10 columns delineating time-lagged relations (with a maximum lag $L=1$ in this case). Dark blue indicates a value of one (presence of an edge), while sky blue symbolizes a value of 0 (absence of an edge). The second column displays the groundtruth causal graphs, and the final column highlights the difference between the estimated and true graphs.}
\vspace{-1.5em}
\label{fig8}
\end{figure}